\pdfoutput=1 
\documentclass[]{article} 

\usepackage{arxiv} 
\usepackage[T1]{fontenc}
\usepackage[utf8]{inputenc}
\usepackage{microtype}
\usepackage{inconsolata}
\usepackage{hyperref}       
\usepackage{url}            
\usepackage{booktabs}       
\usepackage{amsfonts}  
\usepackage{nicefrac}   
\usepackage[x11names]{xcolor}
\usepackage{fontawesome5}   
\usepackage{natbib}
\usepackage{doi} 
\usepackage{times}
\usepackage{latexsym}
\usepackage[notextcomp]{stix}
\usepackage{graphicx}%
\usepackage{multirow}%
\usepackage{multicol}%
\usepackage{setspace}
\usepackage{amsmath,amssymb,amsfonts}%
\usepackage{amsthm}%
\usepackage{mathrsfs}%
\usepackage[title]{appendix}%
\usepackage{textcomp}%
\usepackage{manyfoot}%
\usepackage{tikz}
\usepackage{utfsym}
\usepackage[most]{tcolorbox}
\usepackage{listings}%
\usepackage{tabularx}
\usepackage{csquotes}
\usepackage[export]{adjustbox}
\usepackage{wrapfig}
\usepackage{geometry}
\usepackage{caption}
\usepackage{subcaption}
\usepackage{lscape}
\usepackage{float}
\usepackage{longtable}
\usepackage{array}
\usepackage{colortbl}
\usepackage{pdflscape}
\usepackage{tabu}
\usepackage{threeparttable}
\usepackage{threeparttablex}
\usepackage[normalem]{ulem}
\usepackage{makecell}
\usepackage[frozencache,cachedir=.]{minted}
\newenvironment{longlisting}{\captionsetup{type=listing}}{}
\newcommand{\cmtt}[1]{{\fontfamily{cmtt}\selectfont #1\normalfont}}

\title{"All that Glitters": Approaches to Evaluations with Unreliable Model and Human Annotations}

\author{\fnm{Michael} \sur{Hardy}
  \affil{Stanford University}  
  \email{hardym@stanford.edu} 
}

\author{ \href{https://orcid.org/0000-0001-8988-1518} {\includegraphics[scale=0.06]{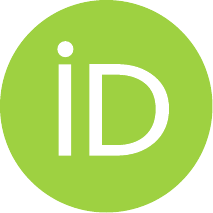} {\hspace{1mm}Michael ~Hardy}}\thanks{Please see \ref{sec:auth} for additional information about the author} \\
	Stanford University University\\
	\texttt{hardym@stanford.edu}
}

\renewcommand{\headeright}{ }
\renewcommand{\undertitle}{A Case Study in Measuring the Quality of Classroom Teaching Using LLMs}

\hypersetup{
pdftitle={All that glitters-Approaches to Evaluations with Unreliable Model and Human Annotations},
pdfauthor={Michael ~Hardy},
pdfkeywords={NLP, LLM, AI, rater bias, teacher development, Generalizability Theory, IRT, hierarchical rater models, reliability, classroom observation},
}

\begin{document}

\maketitle

\begin{abstract}
"Gold" and "ground truth" human-mediated labels have error. The effects of this error can escape commonly reported metrics of label quality or obscure questions of accuracy, bias, fairness, and usefulness during  model evaluation. This study demonstrates methods for answering such questions even in the context of very low reliabilities from expert humans. We analyze human labels, GPT model ratings, and transformer encoder model annotations describing the quality of classroom teaching, an important, expensive, and currently only human task. We answer the question of whether such a task can be automated using two Large Language Model (LLM) architecture families--encoders and GPT decoders, using novel approaches to evaluating label quality across six dimensions: Concordance, Confidence, Validity, Bias, Fairness, and Helpfulness. First, we demonstrate that using standard metrics in the presence of poor labels can mask both label and model quality: the encoder family of models achieve state-of-the-art, even "super-human", results across all classroom annotation tasks. But not all these positive results remain after using more rigorous evaluation measures which reveal spurious correlations and nonrandom racial biases across models and humans. This study then expands these methods to estimate how model use would change to human label quality if models were used in a human-in-the-loop context, finding that the variance captured in GPT model labels would worsen reliabilities for humans influenced by these models. We identify areas where some LLMs, within the generalizability of the current data, could improve the quality of human ratings of classroom instruction. 

\end{abstract}


\section{Introduction}
Human mediated labels always have an unknown amount of error. In machine learning practice, this error is often quantified using inter-rater reliability metrics and correlations. However, this annotation uncertainty is often ignored during standard supervised learning and model evaluation, leading to poorer models \cite[]{belz_non-repeatable_2023}. Thus, imperfect labels are treated as "gold" or "ground truth" \citep{belz_disentangling_2020,hosking_human_2024}. This may be due in part to measures of accuracy being the most preferred methods of assessing and benchmarking model performance \cite[]{birhane_values_2022,ribeiro_beyond_2020,kiela_dynabench_2021}, but common practice might also arise from not using tools expressive enough to interpret labels in low reliability. To that end, this work demonstrates methods for working with low/unknown reliability annotations, often found in tasks requiring complex expert judgment.  

The field of education has many complex tasks that often yield low reliabilities in labels \citep{jurenka_towards_2024, kane_gathering_2012} which make edtech NLP models and research particularly vulnerable to the effects of inexpert annotations  \cite{belz_disentangling_2020,van_der_lee_best_2019,zhou_how_2023}. The case study used to illustrate more expressive methods for working with unreliable labels will be from K12 education. Specifically, this study examines a use case where expert annotations are highly \textit{unreliable} and yet \textit{used in high-stakes decisions}: automated rating of the \textbf{quality of classroom teaching}. Methods used in this paper answer the call from others to evaluate the psychometric properties of models that perform this task \citep{casabianca_effect_2013,liu_measuring_2021}, and do so by comparing metrics across six dimensions of interest: Concordance, Confidence, Validity, Bias, Fairness, and Helpfulness (full results across these metrics against human baselines are in Table \ref{tab:allmqisummary}).
Novel contributions of this work to NLP include: 
\begin{enumerate}
    \item  measurements of the generalizability and dependability of labels used with NLP tasks (Section \ref{sec:gtheorystudy}),
    \item methods for detection of spurious correlations in model outputs via disattenuating low human-model correlations (Section \ref{sec:spurious}),
    \item methods for measuring model biases by disentangling human rater-specific contributions to unknown bias for unknown data sets (Section \ref{sec:biases}),
    \item measurement of model fairness and racial bias in the presence of low label reliabilities (Section \ref{sec:fairness}), and
    \item application of Design Studies (d-studies) from Generalizability Theory (g-theory) for estimating impacts of human-in-the-loop (HIL) model use on human label quality (Section \ref{sec:dstudy}).
\end{enumerate}

This work strengthens the argument that only using simple inter-rater reliability metrics to understand the quality of labels may be masking the limitations of the labeling criteria \citep{hill_when_2012,hosking_human_2024,belz_disentangling_2020}.  It also illustrates how more robust evaluation techniques can yield information in the presence of noisy labels and seemingly inconclusive results. The analyses presented in this study are motivated by issues of model interpretability, fairness, and usefulness. Brief introductions to various techniques will be provided and illustrated via the study task, with interpretation of limitations and recommendations for future research. 

\begin{figure}
    \centering
    \includegraphics[width=0.85\linewidth]{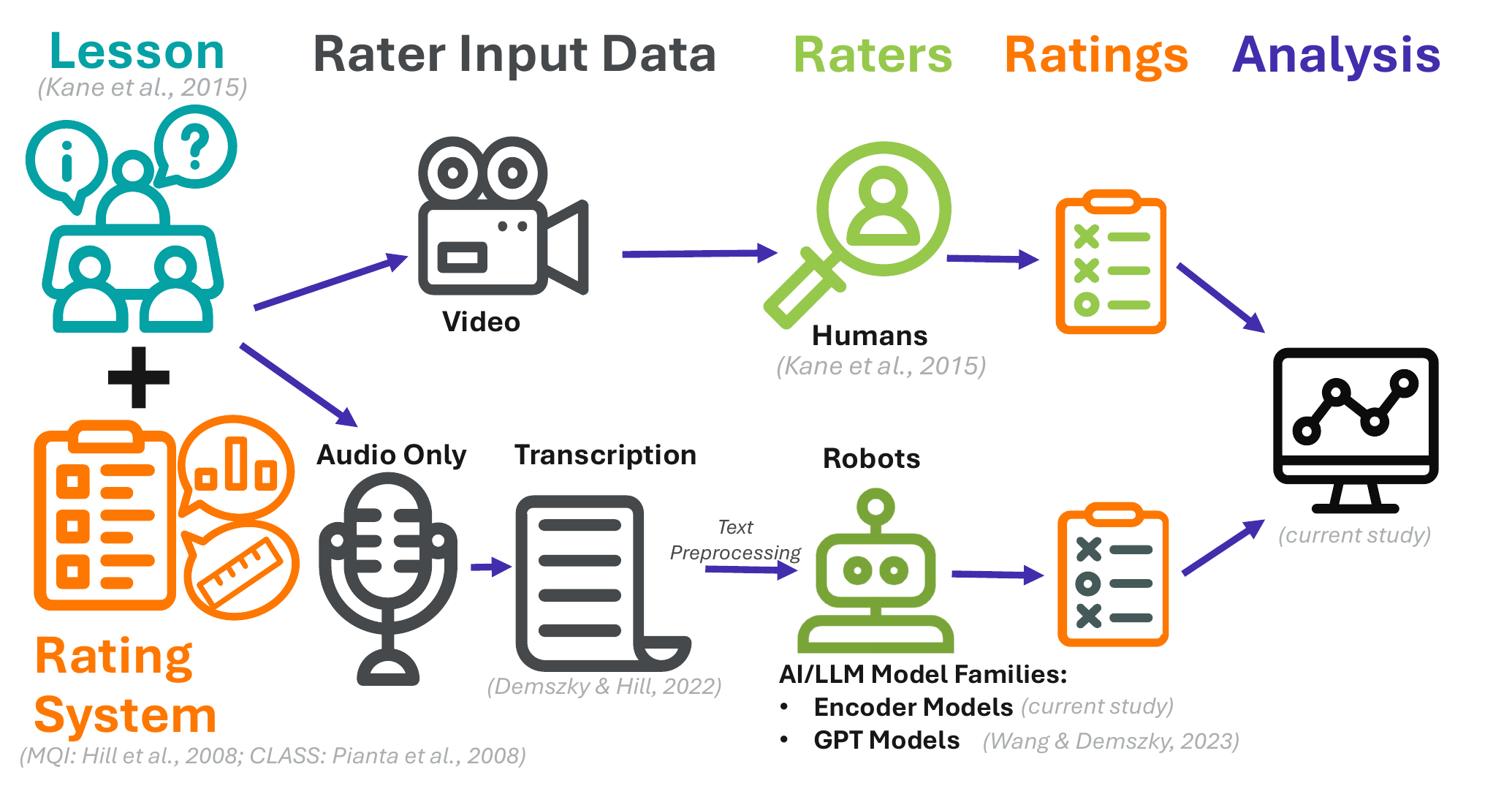}
    \caption{Data Processes and Sources for Studying Teaching and Annotation Quality}
    \label{fig:observation_data}
\end{figure}

\subsection{Study Task: Annotating Teaching Quality}
The classification task of rating teaching may seem deceptively simple: using a rubric, provide a rating for the quality of instruction of an elementary school math classroom. Such ratings are given to all US K12 public education teachers for both formative educator development feedback and as high-stakes teacher evaluations. Despite their ubiquity, these ratings, even when conducted by experts, are unreliable \citep{ho_reliability_2013,kane_national_2015,kane_gathering_2012,glaese_improving_2022,whitehill_automated_2024}, similar to the poor reliability of other K12 education labels \citep{jurenka_towards_2024, tack_bea_2023} that have limited the rigor of education research \citep{slavin_evidence-based_2002,klahr_what_2013, jurenka_towards_2024}.  Studies about ratings of instruction are also extremely expensive to conduct relative to other annotation tasks \citep{grissom_effective_2013,liu_measuring_2021,jurenka_towards_2024}, with only two major studies across hundreds of public school teachers that use authentic instructional metrics to support development: the MET study \citep{kane_have_2013,kane_gathering_2012} and the NCTE Main Study \citep{kane_national_2015}, the latter of which is the source of data for this study.

From the first study, \citeauthor{ho_reliability_2013} estimated that increasing the number of human classroom observers can improve the reliability of ratings assigned. In their major work on the topic, they use methods similar to those in this paper to measure conditions under which the use of additional human raters can increase the reliability of this resource- and time-intensive task \citep{kane_gathering_2012,whitehurst_evaluating_2014}. Considering the expense, importance, complexity, and lack of reliability in ratings of classroom teaching and also the advances in natural language processing, automated ratings based on classroom discourse offer one potential solution. 
\paragraph{Study Research Question:} How can we know when the behaviors of models are good enough to be used lieu of humans as estimated by \citeauthor{ho_reliability_2013}? 

Answering whether automated ratings can similarly improve human annotations is understanding the extent to which models' added contributions would result in similar benefits as expected from humans. Thus, this study illustrates methods for working with unreliable labels in NLP tasks by investigating and disentangling the variation found in human and model raters from the variation found within the observations and the instrument used for the annotation task. The model raters are comprised of two families: the "GPT" family of autoregressive in-context learners from \cite{wang_is_2023} (using ChatGPT) with three models whose siblings differ by prompt engineering strategies and an "Encoder" family built for this study whose five siblings differ in embeddings and a few adjustments to training hyperparameters. Quality of ratings will be examined between and within families and individual raters.

\section{Related Work}
\subsection{Annotation Quality and Bias}
Better understanding human label behaviors is key to training and evaluating models \citep{webson_are_2023, webson_prompt-based_2022,gordon_jury_2022}. 
Accuracy, based on "gold" or "ground truth" labels, is the primary and most valued performance metric by which LLMs are evaluated  \cite[]{birhane_values_2022,ribeiro_beyond_2020,kiela_dynabench_2021}. For expediency of development, data scientists often choose to assume data labels are reliable, accurate, and end-task aligned for intended real-world use cases, \cite[]{hosking_human_2024,bejar_human_2006,messick_test_1998}, even in scenarios where these assumptions could be detrimental (e.g., performing complex high-stakes tasks, reducing discriminatory biases found in data \citep{field_survey_2021} that are immutably historical by definition of their creation, etc.), which is especially true of autoregressive models, whose labels are Internet text and which contain harmful biases \citep{hofmann_ai_2024,hofmann_dialect_2024}. Assessing the accuracy and reliability of idiosyncratically  human annotated "ground truth" can be difficult \cite[]{eckes_detecting_nodate,wind_exploring_2019, wind_nonparametric_2019,abercrombie_consistency_2023,baan_interpreting_2024,baan_stop_2022,waseem_are_2016, kazai_analysis_2013, hosseiny_marani_one_2022,tack_bea_2023,hosking_human_2024}, a challenge that is exacerbated when label uncertainty is underexamined or underreported. Limited transparency around label quality makes it more challenging to measure biases, interpret model findings, assess individual fairness, and establish real-world validity \citep{hill_when_2012,jurenka_towards_2024}.

Powerful and provocative research has begun to address the limitations of accuracy-only evaluations and propose more fair and responsible solutions under assumptions of uncertainty \citep{hardt_equality_2016, dwork_fairness_2012,kasy_fairness_2021,song_learning_2020, zhao_right_2021, corbett-davies_measure_2023, pleiss_fairness_2017,zemel_learning_2013},  including techniques for addressing when labels lead to undesirable model behaviors \cite[]{ding_retiring_2022,hebert-johnson_multicalibration_2018,qi_fine-tuning_2023}. This paper offers several ways to quantify these issues and improve interpretability and explainability \cite[]{adebayo_sanity_2020, lundberg_unified_2017,rudin_stop_2019,kim_interpretability_2018}. 

\subsection{Teacher Development and Evaluation}
School leaders working with teachers to improve the quality of instruction typically evaluate the teacher's proficiency in a range of competencies (typically measured during in-class observation and evaluation on a teaching rubric; \cite{aguilar_developing_2013, bambrick-santoyo_get_2016,bambrick-santoyo_leverage_2018}), then determine which competencies are most important to improve first (i.e., which change will have the biggest impact on student learning), and then provide supportive feedback and coaching. This paper focuses on the first step of evaluating teacher proficiency, which is often time-consuming and produces ratings (labels) that are unreliable \cite[]{kane_gathering_2012,blazar_validating_2018,kane_have_2013,casabianca_effect_2013}. Without accurate classifications, it is challenging for practitioners to prioritize instructional needs and aligned practices from among the many elements of good teaching \citep{saphier_skillful_2008,darling-hammond_what_2014,hammond_culturally_2015,lemov_teach_2015,lemov_teach_2021,liljedahl_building_2021,darling-hammond_implications_2020,schwartz_abcs_2016} and for researchers to empirically quantify the impact of good teaching practices \cite{pianta_conceptualization_2009,charalambous_13_2019,blazar_challenges_2022,jurenka_towards_2024}.  

Thus, this work provides a bridge to research seeking to improve teaching quality by providing feedback to teachers on various instructional techniques \citep{samei_domain_2014, donnelly_words_2017, kelly_automatically_2018, demszky_measuring_2021, suresh-etal-2022-talkmoves, jacobs-etal-2022-masked, alic-etal-2022-computationally,demszky_m-powering_2023, demszky_does_2024, demszky_improving_2023}. These feedback studies identify linguistic features correlated with an aspect of good teaching, but may optimistically overgeneralize the usefulness, efficacy, and universality of identifiable features, providing specific prescriptions without diagnosis. Matching these models with the specific needs of teachers will help provide a more individualized approach to teacher development, one based on understanding instructional needs and then providing corresponding supports.

Only three recent studies have sought to use LLMs to provide ratings of classroom instruction (via classroom transcripts) using authentic rating rubrics. \citet{whitehill_automated_2024} use a mix of zero-shot and bag-of-words model configurations to provide scores to instructional domains for Pre-Kindergarden classrooms using a private dataset, commenting on their highest Pearson $r$ correlation statistic of 72 experiments ($r = 0.48$)  that it "approaches human inter-rater reliability".  \citet{wang_is_2023} and \cite{xu_promises_2024} both use the same publicly available datasets as the present study, and the approach of the former will be discussed further. \citeauthor{xu_promises_2024} use a by-item "best of" modeling approach which included experiments with BERT \citep{devlin_bert_2019}, DistilBERT \citep{sanh_distilbert_2020}, RoBERTa \citep{liu_roberta_2019}, XLNet \citep{yang_xlnet_2020}, Llama 2 \citep{touvron_llama_2023}, and ChatGPT, using models in two-stages where the first stage LLM provides the best text to the second stage which generates the rating. Unfortunately, the LLM-facilitated preprocessing of text and the by-item model training and selection processes limit the generalizability and transferability of their methods. While Xu et al. did not publicly release model ratings or the combinations of ensembles used, they did report Spearman correlation values for each of the best of several item-specific model constructions. In Figure \ref{fig:spearmanfig}, the results from their reported held-out test set are displayed alongside those from the present study for a comprehensive comparison across all studies reporting performance of automated ratings which use the MQI rubric or which use publicly accessible data.

\section{Data}\label{sec:data}
The data used in this study and in \cite{wang_is_2023} are from the National Center for Teacher Effectiveness (NCTE) Main Study \citep{kane_national_2015}, which contains three years of data collection and observations of math instruction in approximately fifty schools and three-hundred (4th and 5th grade) mathematics classrooms across four school districts in the United States, including expert human ratings of individual video-captured classroom lessons across two observation instruments \cite[]{bacher-hicks_evaluation_2017, bacher-hicks_experimental_2019}: the CLASS framework (12 items) \citep{pianta_classroom_2008} for general instructional practice and the content-specific Mathematical Quality of Instruction (MQI; 13 items) \citep{hill_mathematical_2008}, together yielding over 400,000 distinct human rating labels assigned, the distributions of which are in Figure \ref{fig:rating-dist}. Each instrument item is intended to measure a different aspect of teaching quality. 

Like all human mediated labels,\footnote{Label(er), rate(r), annotat(ion/or), and score(r) will be used interchangeably for these classification tasks, as terminology varies multidisciplinarily.} an individual classroom observation rating requires at a minimum three facets: (1) a task with rating criteria (Section \ref{sec:mqiratingcriteria}), (2) raters/labelers (Section \ref{sec:humraters}), and (3) observations to be classified (sections of transcripts of classroom discourse, Section \ref{sec:classroomdata}). As tasks increase in complexity, three facets contribute more error to estimates. This dataset has the additional real-world challenges of having very long and noisy transcripts and having large imbalances (Figure \ref{fig:panels} panel (a), Figure \ref{fig:rating-dist}) in human labels that have hindered previous research \citep{xu_promises_2024,wang_is_2023}, but which provide extra opportunity to demonstrate the importance of robust methods of evaluation.

\subsection{Rating Criteria: MQI Rubric}\label{sec:mqiratingcriteria}
Just as all raters contribute uncertainty to a system, so too do the measurement instruments. Ambiguity uncertainty is introduced when an instrument, instruction, or criteria for a task has language that could lead to two equally-expert raters to different results, ceteris paribus. The 13 MQI items within the dataset have at least two raters per classroom observation. While both humans and Encoders evaluated all items, the this paper will focus on the 4 of the 13 MQI items evaluated in \cite{wang_is_2023} to support comparability across humans and models.\footnote{\citeauthor{xu_promises_2024} provided results for 11 of the 13 MQI items. No explanation is provided for the exclusion of \cmtt{MGEN} and \cmtt{USEPROD}.}  These four ternary items are teacher explanations ( \cmtt{EXPL}),  remediation of student errors (\cmtt{REMED}), student questioning and reasoning (\cmtt{SMQR}), and imprecision in mathematical language (\cmtt{LANGIMP}).\footnote{\cmtt{LANGIMP} is reverse-coded so higher scores are better and has interesting self-referentiality vis-à-vis instrument uncertainty that is worth noting, but out of scope for the current study. See Appendix \ref{apx:neg-mqi} for more on this and other negatively worded items.} Analyses for all other items are in the appendices. Prior studies have explored the reliability of MQI instrument ratings generally \cite[]{kane_gathering_2012, mantzicopoulos_mathematical_2018,hill_when_2012,kane_national_2015,ji_using_2023}; this study confirms previous findings via reproduced reliability metrics in Section \ref{sec:humraters}, which correspond to the NCTE Study, Appendix Section 2).

\subsection{Human Expert Raters}\label{sec:humraters}
Human rater information for both the MQI and CLASS instruments can be found in the Appendix of the \textit{DS0 Study-Level Files} from the NCTE Main study. MQI raters in particular were recruited from a separate pool of applicants based on their background in mathematics and through contacting colleagues in mathematics departments \citep{hill_validating_2012, blazar_attending_2017} and then passed certification exams to score the MQI, and attended biweekly calibration meetings to ensure standardization of scoring procedures. 

\subsection{Classroom Observations}\label{sec:classroomdata}
63 human raters watched videos and provided ratings at regular intervals across all items in the MQI. Transcripts of these same videos \citep{demszky_ncte_2022} are used by LLMs for the same task, where the class discourse is equipartitioned across utterances (GPT family models) or words (Encoder family models) by the total number of classroom segments to align the text to the human labels in the absence of timestamps. Data from the NCTE Main study \citep{kane_national_2015}  \footnote{\url{https://www.icpsr.umich.edu/web/ICPSR/studies/36095/datadocumentation}} and for the associated transcripts \citep{demszky_ncte_2023}\footnote{\url{https://github.com/ddemszky/classroom-transcript-analysis}} are available online.

\section{Model Families and Model Rater Data}\label{sec:modelfams} 
\paragraph{GPT Models} The GPT model family from \citet{wang_is_2023}\footnote{\url{https://github.com/rosewang2008/zero-shot-teacher-feedback/}} have 7,660 ratings for 223 different teachers. The family consists of three models differing in prompt engineering methods (herein called \textit{N}, \textit{NR}, and \textit{ND}), and brief summary of those differences is in Table \ref{tab:gptdiff}. GPT models were evaluated on curated selections of classroom text with the least transcriptorial noise (i.e., minimizing instances of \texttt{[inaudible]}), and were edited to indicate whether the speakers were teachers or students. 

\paragraph{Encoder Models} Encoder family models are custom transformer encoders trained on the NCTE classroom transcripts. The five models (\textit{un1, un2, un3, gte, }and\textit{ e5}) use fixed-parameter pretrained sentence embeddings, differing in these and in training hyperparamters, thereby exploiting LLM sensitivites to pretraining regimes \citep{damour_underspecification_2020,mccoy_embers_2023}. A summary of differences is in Table \ref{tab:encoder} and more training details can be found in Appendix \ref{apx:encoders}. In contrast to the model experiments of \citeauthor{xu_promises_2024} who used different combinations of models by item, each encoder model produces labels for all 13 MQI (and 12 CLASS) items. In contrast to the GPT models, the only text preprocessing used with the Encoders simply replaced all transcription notes with \verb|[inaudible]| to mimic the uncertainty in live audio transcription, and no edits to indicate speakership were included. For the Encoder models, all model outputs\footnote{\url{https://github.com/hardy-education/LLM-Psychometrics}} in this study were conducted with a lesson-level-stratified held-out test set (see Figure \ref{fig:test_set}) that was not used during model development. Encoder models were trained a single GPU in Google Colab with training detailed in Appendix \ref{apdx:training}.

\section{Evaluation Methods}\label{sec:evalmethods}

\begin{figure}[t]
    \centering
    \includegraphics[width=1\linewidth]{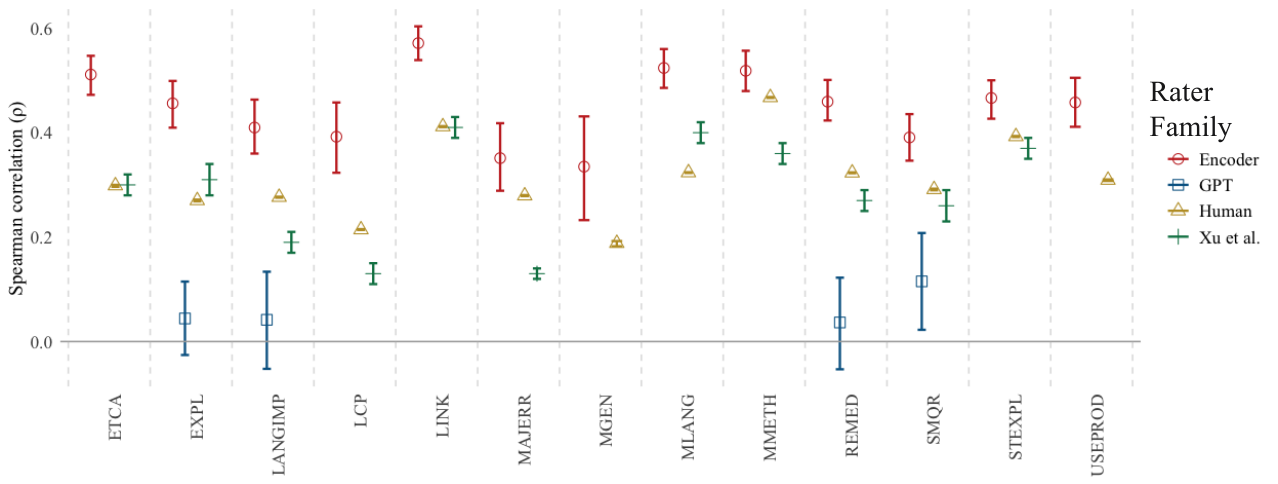}
    
    \renewcommand{\arraystretch}{1.2}
    \setlength{\tabcolsep}{1.5pt} 
    \begin{tabularx}{1\textwidth}{l*{13}{c}}
        \toprule
        Raters & {\small ETCA} & {\small EXPL} & {\small LANGIMP} & {\small LCP} & {\small LINK} & {\small MAJERR} & {\small MGEN} & {\small MLANG} & \small{MMETH} & {\small REMED} & {\small SMQR} & {\small STEXPL} & {\small USEPROD} \\ \hline
        \textcolor{Gold3}{\textbf{Humans}}    & 0.3 & 0.27 & 0.28  & 0.21 & 0.41 & 0.28 & 0.19 & 0.32 & 0.47 & 0.32 & 0.29 & 0.39 & 0.31 \\
        \textcolor{Red3}{\textbf{Encoders}} & \textbf{0.51} & \textbf{0.46} & \textbf{0.41} & \textbf{0.39} & \textbf{0.57} & \textbf{0.35} &\textbf{ 0.33} & \textbf{0.52} & \textbf{0.52 }& \textbf{0.46} & \textbf{0.39} & \textbf{0.47} & \textbf{0.46 }\\
        \textcolor{SteelBlue4}{\textbf{GPTs}}           &      & 0.04 & 0.04 &      &      &      &      &      &      & 0.04 & 0.12 &      &       \\
       \textcolor{Chartreuse4}{\textbf{Xu et al.}} & 0.3 & 0.31 & 0.19  & 0.13 & 0.41 & 0.13 &      & 0.4  & 0.36 & 0.27 & 0.26 & 0.37 &      \\
        \bottomrule
    \end{tabularx}
    
    \caption{Spearman correlation coefficients and confidence intervals by MQI Item for all rater families and studies. \textcolor{Gold3}{\textbf{Human}} \citep{kane_national_2015}, \textcolor{Red3}{\textbf{Encoder}} (current study, Section \ref{sec:modelfams}), and \textcolor{SteelBlue4}{\textbf{GPT}} \citep{wang_is_2023} family correlations are between each rater and one randomly sampled human rater for each observation, following the processes used in the original human study, repeated 1,000 times for bootstrapped confidence intervals. \textcolor{Chartreuse4}{\textbf{\citeauthor{xu_promises_2024}}} coefficients are reported from Tables 5 and 9 of that paper, where each number represents the best of several ensemble models fit for each individual item. \textbf{Bold} in the table indicates highest performing label family.}
    \label{fig:spearmanfig}
\end{figure}

Typical \textbf{reliability} metrics (see Section \ref{sec:reliabilities}) provide a backdrop of descriptives that can flag issues of low quality labels. Measures of statistical \textbf{dependability} can be used for generalizing label conclusions and identifying spurious correlations (see Section \ref{sec:spurious}), a part of improving \textbf{accuracy}. Methods for disentangling human and model label \textbf{biases} (see Section \ref{sec:biases}) are first demonstrated and then extended to estimate \textbf{fairness} across racial lines in Section \ref{sec:fairness}.  \textbf{Usefulness}, as measured by the amount of rating reliability improvement a model can provide to a human rater in human-in-the-loop contexts, including associated cost savings in human time (for encoder models) are in Section \ref{sec:dstudy}.

\subsection{Concordance: Agreement and Reliability Metrics}\label{sec:reliabilities}
\paragraph{RQ 1:} How do automated models perform relative to humans in the presence of low label reliability?  \textbf{\hypertarget{rq1}{RQ 1:} Case Study Reframing:} How well do automated models perform relative to humans when evaluating instruction? 
\subsubsection{Baseline Human Metrics} 
\begin{wrapfigure}[20]{r}{0.35\textwidth}
    \begin{tcolorbox}[colback=Turquoise3!5!white,colframe=Turquoise3!75!black,title={\large \textbf{RQ1: \hypertarget{Concordance}{Concordance}}}]

          \begin{tcolorbox}[colback=Turquoise3!10!white,colframe=Turquoise3!90!black,title=\textbf{Metrics}]
              \textbf{Correlation}:  $r$, $\rho$, $\tau$
              \textbf{Inter-rater Agreement}: \% Agree, \% Agree ± 1, Cohen's $\kappa$, QWK
        \end{tcolorbox}
        \begin{tcolorbox}[colback=Turquoise3!1!white,colframe=Turquoise3!90!black,title=\textbf{Intuition}: \textit{QWK}]
            QWK is the extent to raters agree on ratings, \textit{not by chance}. Bigger \textit{differences} in ratings show less agreement, scaled quadratically.
        \end{tcolorbox}
    \end{tcolorbox}
\end{wrapfigure}

Full reproductions\footnote{Small differences in the reported values here compared to the original study arise from random human rater selection required in the procedure, which were done at the segment level. All families and model evaluations used the same random sample of human raters for comparison.} of all reliability metrics and calculation processes exactly as described in the NCTE Main Study Appendix Section 2 were conducted. \citep{kane_national_2015}. Following their same procedures, replicated calculations were extended to the model families, replacing a human rater score with a \textit{specified} or \textit{random} model for evaluations of \textit{individual} models and \textit{model families}, respectively. Intra-class correlations (ICCs) are with the calculation methods in Appendix \ref{apx:rel_metrix_icc}. Reproduced human results and model results, including additional metrics in this section, are fully reported in Appendix \ref{apx:fullresult} and all item results can be found in the online supplement. 
 
\subsubsection{Commonly Used Metrics}

The results also include three additional correlation and reliability metrics: Quadratic Weighted Kappa (QWK) typically used in ordinal classification tasks to penalize distance quadratically (squared error) while accounting for categorical agreement by chance (e.g., \cite{shermis_state---art_2014,hardy_toward_2021,wang_is_2023}), Pearson correlation $r$, (e.g., \cite{whitehill_automated_2024}) Spearman correlation $\rho$ (e.g., \cite{wang_is_2023,xu_promises_2024}), and Kendall correlation $\tau$ (e.g., \cite{liu_g-eval_2023}). Figure \ref{fig:spearmanfig} shows Spearman correlations ($\rho$) and confidence intervals for all model families and for  models from \cite{xu_promises_2024}. The table in Figure \ref{fig:spearmanfig} contains the $\rho$ estimates. 

\subsubsection{Results}
Using nearly any standardized combination of metrics across all items from Section \ref{sec:reliabilities}, Encoder models perform better than the \textit{single highest performing expert human rater}.  The human labels assigned for the four focus MQI have very low reliabilities, despite the significant training and calibration for human raters described in \ref{sec:humraters}. Overall, the human labels are highly unreliable, but if a researcher were trying to compare the model to human performance, they could be displayed as they are in Table \ref{tab:reliabsummary}. For metrics of agreement and reliability, each encoder model outperformed humans on average, whilst each GPT model underperformed humans on every metric and every item. Table \ref{tab:reliabsummary} has a summary of the full panel of lesson segment-level inter-rater reliability metrics for each MQI item. Specific metrics for the four focus MQI items in this study are in Panel (b) in Figure \ref{fig:panels}, and the full individual model-item comparisons for all MQI items and metrics in this section are in Table \ref{tab:tab:full}. Additionally, the detailed full results for all models and metrics, MQI, and CLASS rubrics can be found in the supplementary materials online. 

Using only these metrics and without further testing, one might assume that the encoder models are therefore ready to help with the task of automated annotations of teaching quality or that GPT models show improvement to ICC measures and could be helpful. \textbf{\textit{Implications}}: Basic statistics in the presence of unreliable labels can mislead interpretations of model performance. Researchers should be wary of studies reporting few metrics in the presence of low reliabilities.

\begin{table}
    \centering

\begin{tabular}{lcccccccccc}
\toprule
Metric & \textbf{Encoders} & \textit{un1} & \textit{un2} & \textit{un3} & \textit{gte} & \textit{e5} & \textbf{GPTs} & \textit{N} & \textit{NR} & \textit{ND} \\ \hline
\%Agr  & \textbf{0.54} & \textbf{0.69} & \textbf{0.77} & \textbf{0.69} & 0.39 & 0.39 & 0.00 & 0.00 & 0.00 & 0.00\\
C's $ \kappa$   & \textbf{0.69} & \textbf{0.85} & \textbf{0.77} & \textbf{0.62} & \textbf{0.62} & \textbf{0.62} & 0.00 & 0.00 & 0.00 & 0.00 \\
QWK  & \textbf{1.00} & \textbf{1.00} & \textbf{1.00} & \textbf{1.00} & \textbf{0.92} & \textbf{0.92} & 0.00 & 0.00 & 0.00 & 0.00 \\
$r$ & \textbf{1.00} & \textbf{1.00} & \textbf{1.00} & \textbf{1.00} & \textbf{1.00} & \textbf{1.00} & 0.00 & 0.00 & 0.00 & 0.00 \\
$\rho$ & \textbf{1.00} & \textbf{1.00} & \textbf{1.00} & \textbf{1.00} & \textbf{0.77} & \textbf{0.77} & 0.00 & 0.00 & 0.00 & 0.00 \\
$\tau$ & \textbf{1.00} & \textbf{1.00} & \textbf{1.00} & \textbf{1.00} & \textbf{0.77} & \textbf{0.77} & 0.00 & 0.00 & 0.00 & 0.00 \\
\bottomrule
\end{tabular} 
    \caption{\textbf{Concordance}: Performance above Human Reliability and Agreement Metrics. Proportion of MQI items where the \textit{model} or \textbf{model family} listed had \textit{better} results than human baselines. \textbf{Bold} indicates where performance was better on more than half of items rated. Inter-rater reliability metrics introduced in Section \ref{sec:reliabilities}. \textbf{C's $ \kappa$}: Cohen's $ \kappa$; \textbf{QWK}: Quadratic Weighted Kappa; \textbf{\%Agr}: percent exact agreement;  \textbf{${r}$}: Pearson's correlation;  $\mathbf{\rho}$: Spearman's rank correlation;  $\mathbf{\tau}$: Kendall's concordance correlation;. Full data can be found in the supplementary material online. }
    \label{tab:reliabsummary}
\end{table}

\subsection{Confidence: Generalizable Reliability}\label{sec:gtheorystudy}
\begin{wrapfigure}[20]{r}{0.35\textwidth}
    \begin{tcolorbox}[colback=Turquoise3!5!white,colframe=Turquoise3!75!black,title={\large \textbf{RQ2: \hypertarget{Confidence}{Confidence}}}]
          \begin{tcolorbox}[colback=Turquoise3!10!white,colframe=Turquoise3!90!black,title=\textbf{Metrics}]
              \textbf{Generalizability}: $\mathbf{E}\rho^2$ \par
              \textbf{Dependability}: $\mathit{\Phi}$
        \end{tcolorbox}
        \begin{tcolorbox}[colback=Turquoise3!1!white,colframe=Turquoise3!90!black,title=\textbf{Intuition}: $\mathbf{E}\rho^2$ and $\mathit{\Phi}$]
            By accounting for the different facets of variation, we can  estimate how much of the relative ($\mathbf{E}\rho^2$) and absolute ($\mathit{\Phi}$) label variation \textit{associated} with the teacher is attributable to the \textit{teacher \textbf{only}}. 
        \end{tcolorbox}
    \end{tcolorbox}
\end{wrapfigure}
\paragraph{RQ 2:}  How generalizable are findings from unreliable labels? \textbf{\hypertarget{rq2}{RQ 2} Case Study Reframing:} To what extent would the ratings of a teacher's instructional quality persist across lessons or contexts?
\subsubsection{Generalizability and Dependability}

Generalizability Study (g-study) \citep{brennan_generalizability_2001, brennan_generalizability_2013,brennan_variability_2001,hill_when_2012} designs utilize random effect estimates across possible configurations of different sources of variance to quantify how generalizable labels. This is done by estimating the extent to which given labels would persist if sources of variation changed (e.g., same teacher, different day; same lesson, different rater; human rater vs model rater; etc.). $\mathbf{E}\rho^2$ is a measure of the relative \textit{generalizability} of a rating (i.e., is rating \textit{order} preserved), and $\mathit{\Phi}$, accounting for absolute error, is a measure of label \textit{dependability}: how likely specific ratings would be numerically the same with different sources of variation. These two reliability-like estimates can help quantify how "golden" labels are. 

The multifaceted g-study design used to estimate the how much variation ($\nu$) in individual teachers' instructional quality, $i$, contributed to a rating label, $X$, annotated for a section of a lesson, $s$, during an observation, $o$, on rubric item $j$ by rater $r$ is known as a Item-by-Rater-by-Segment-within-Observation-within-Individual Teacher design: $J\times R\times (S:O:I)$. Overall estimates across all MQI items for a given rater family, $\mathbb{F}$, are in Table \ref{tab:gen_and_dep}. For item-level reliabilities, we simplify the expression by holding the item fixed, resulting in a $R\times (S:O:I)$ design. Using nested random effects notation, the estimation model is:
\begin{align}\label{eq:gstudy}
  X_{s:o:i r}^{(j)}=\mu+\nu_{i}+\nu_{o:i}+\nu_{s:o:i}+ \nu_{ir}+\nu_{r}+\nu_{s:o:ir} , \forall j \in \textbf{J}
\end{align}
where $j$ indicates the item index.\footnote{For the estimates in Fig. \ref{fig:panels} (c), for dependability metrics of Section \ref{sec:spurious}, and for comparability with human baselines\citep{hill_when_2012,kane_national_2015, ho_reliability_2013,kane_gathering_2012}, a simplified model, an by-item $R\times (O:I)$ design, was conducted for the human expert rater family with results in Appendix \ref{apdx:humgstudy}.  The simplified model is $ X_{o:i r}^{(j)}=\mu+\nu_{i}+\nu_{o:i}+\nu_{ir}+\nu_{r}+\nu_{o:ir}$  The full model structures of Eq. \ref{eq:gstudy}, \ref{eq:Erho} and \ref{eq:phi} are used for Section \ref{sec:dstudy}.} Code for the model specification is in Appendix \ref{apdx:gstudy}. Then, $\mathbf{E}\rho^2$ (Equation \ref{eq:Erho}) and $\mathit{\Phi}$ (Equation \ref{eq:phi}) are easily estimated from the random effects for raters in rater family $\mathbb{F}$:
\begin{align}\label{eq:Erho}
 {\mathbf{E}\mathit{\rho}^2_{\mathbb{F}}}^{(j)} = \frac{\nu_{ij}}{\nu_{ij} +\nu_{o:ij}+ \nu_{s:o:ij} + \nu_{irj} + \nu_{s:o:irj}} \text{, } 
\end{align}

\begin{align}\label{eq:phi}
  \mathit{\Phi}_{\mathbb{F}}^{(j)} = \frac{\nu_{ij}}{\nu_{ij} +\nu_{o:ij}+ \nu_{s:o:ij} + \nu_{irj} + \nu_{rj} + \nu_{s:o:irj}} \text{, } 
\end{align}

$\forall r \in \mathbb{F}$, where the individual item-rating-segment variation, $\nu_{s:o:irj}$, is confounded with error variation. These results are found in Table \ref{tab:gen_and_dep}. A figure comparing the $\mathbf{E}\hat{\rho}^2_j$ item values to item-level reliability estimates related to Guttman's $\lambda_6$ , $\rho^{\lambda_6}_{jj\prime}$,  from Classical Test Theory \citep{zijlmans_item-score_2018, zijlmans_methods_2018}, can be found in Appendix \ref{apx:ep2_g6}. Additionally an illustration of sources of variance including descriptions can be found in Appendix \ref{apx:gstudies}, color-coded to support interpretation of sources of variance with the table of results. 

\subsubsection{Results}

Humans, on average, produce labels that are both more reliable and generalizable. The full results for human rater labels, decomposed into variance components, can be found in \ref{apdx:gstudy}\footnote{Appendix 2.c of \cite{kane_national_2015} provided a g-study, but, surprisingly, not using the data from the study.} and estimates for $\mathbf{E}\rho^2$ and $\mathit{\Phi}$ can also be found in panel (c) of Figure \ref{fig:panels}. The Encoder models outperform humans on nearly every item in terms of inter-rater reliability metrics (Table \ref{tab:reliabsummary}) , but not in generalizable reliability metrics as seen in panel (c) tables of Figure \ref{fig:panels}. Importantly, the large difference between $\mathbf{E}\hat{\rho}^2$ and $\mathit{\hat{\Phi}}$ for Humans and Encoders is due to properties of individual items, which accounted for over 75\% of the variation in those families. GPT models, on the other hand, did not change ratings very much on different items, consistent with literature on these models not understanding such prompts \cite[]{liu_lost_2023,webson_prompt-based_2022,heo_llms_2024}. 

Table \ref{tab:gen_and_dep} shows that Encoder model still performs better than humans on the majority of items, but it is no longer as clear. Interestingly, as mentioned in Section \ref{sec:modelfams}, the encoder models did not receive any annotations outside of the transcript, including speaker. This means that the model would struggle to identify teacher explanations (\cmtt{EXPL}) from student explanations (\cmtt{STEXPL}). This shift in interpreting encoder family performance from superhuman to zero reliability adds validity to the argument that these metrics provide valuable insight, showing that the relationships found in some of the variables could be explained by variance unrelated to the label construct. \textbf{\textit{Implications}}: Measures of generalizability and dependability derived from structured variance decomposition can meaningfully quantify label quality.

\begin{table*}
    \centering

\begin{tabular}{lcccccc}
\toprule
 & & $\mathbf{E}\hat{\rho}^2$ & & & $\mathit{\hat{\Phi}}$ & \\
ITEM & Human & Encoders & GPTs & Human & Encoders & GPTs \\ \hline
ETCA    & 0.17 & \textbf{0.20} &             & 0.15 & \textbf{0.19} &             \\
\underline{EXPL}    & \textbf{0.15} & 0.00 & 0.00 & \textbf{0.12} & 0.00 & 0.00 \\
\underline{LANGIMP} & 0.09 & \textbf{0.15} & 0.08 & 0.08 & \textbf{0.14} & 0.08 \\
LCP     & 0.11 & \textbf{0.27} &             & 0.09 & \textbf{0.26} &             \\
LINK    & 0.13 & \textbf{0.19} &             & 0.12 & \textbf{0.19} &             \\
MAJERR  & \textbf{0.08} & 0.00 &             & \textbf{0.07} & 0.00 &             \\
MGEN    & 0.03 & \textbf{0.08} &             & 0.02 & \textbf{0.08} &             \\
MLANG   & 0.07 & \textbf{0.18} &             & 0.06 & \textbf{0.17} &             \\
MMETH   & 0.13 & \textbf{0.37} &             & 0.13 & \textbf{0.36} &             \\
\underline{REMED}   & \textbf{0.13} & 0.10 & 0.05 & \textbf{0.11} & 0.09 & 0.04 \\
\underline{SMQR}    & \textbf{0.14} & 0.09 & 0.00 &\textbf{ 0.13} & 0.09 & 0.00 \\
STEXPL  & \textbf{0.25} & 0.00 &             & \textbf{0.23} & 0.00 &             \\
USEPROD & 0.19 & \textbf{0.25} &             & 0.17 & \textbf{0.25} &             \\
\hline
All Items & \textbf{0.114} & 0.106 & 0.007 & 0.010 & \textbf{0.014} & 0.004 \\
\bottomrule
\end{tabular} 
\caption{Generalizability and Dependability metrics by model families for each MQI Item. \textbf{Bold} represents the best rater family for each of $\mathbf{E}\rho^2$ and $\mathit{\Phi}$, respectively. \underline{Underlined items} are focus MQI items, because they were evaluated by \cite{wang_is_2023}. For the overall "All Items" calculation, a $J\times R\times (O:I)$ model was used for comparability with other similar research.}
    \label{tab:gen_and_dep}
\end{table*}
\subsection{Validity: Convergent and Spurious Correlations}\label{sec:spurious}
\paragraph{RQ 3:} To what extent can accuracy and validity be estimated with unreliable labels? \textbf{\hypertarget{rq3}{RQ 3}Case Study Reframing:} To what extent do models and humans rate the same underlying construct similarly?

\subsubsection{Disattenuating High Noise Correlations}
Dependability and generalizability do not guarantee accuracy, but even at these very low levels, they can be used in indirect tests of convergent validity to see whether correlations between humans and models are low because of measurement error, such as poor rubric item construction, or because the two sets are really uncorrelated.  If an individual teacher's latent instructional ability $\theta_i$ is about the same from lesson to lesson with the same students, we can correlate $\hat\theta_i$ for human ($\mathbb{h}$) and model ($\mathbb{m}$) family ratings for \textit{different lessons} coming from the \textit{same teacher} and correct for measurement error by disattenuating the correlations by each rater family's $\mathbb{F}$ label generalizability, $\mathbf{E}\hat{\mathit{\rho}}_{\mathbb{F}}^{(j)}$, for a given item $j$. The disattenuated correlation,$\mathbf{\varrho}_{\mathbb{hm}}^{(j)} $, between humans and a family of models for item, $j$, can be estimated:
\begin{align}\label{eq:disatten}
\mathbf{\varrho}_{\mathbb{hm}}^{(j)}=\frac{\operatorname{Corr}[\operatorname{\Tilde{\mathcal{X}}_\mathbb{h}}(i,\mathfrak{L},j, r_{\mathbb{h}}), \operatorname{\Tilde{\mathcal{X}}_\mathbb{m}}(i,\neg  \mathfrak{L},j, r_{\mathbb{m}})]}{\sqrt{{\mathbf{E}\hat{\mathit{\rho}}^2_{\mathbb{h}}}^{(j)} {\mathbf{E}\hat{\mathit{\rho}}^2_{\mathbb{m}}}^{(j)}}}
\end{align}

\begin{wrapfigure}[23]{r}{0.35\textwidth}
    \begin{tcolorbox}[colback=Turquoise3!5!white,colframe=Turquoise3!75!black,title={\large \textbf{RQ3: \hypertarget{Validity}{Validity}}}]
          \begin{tcolorbox}[colback=Turquoise3!10!white,colframe=Turquoise3!90!black,title=\textbf{Metric}]

              \textbf{Disattenuated Convergent Correlation}: $\mathbf{\varrho}_{\mathbb{hm}}^{(j)}$

        \end{tcolorbox}
        \begin{tcolorbox}[colback=Turquoise3!1!white,colframe=Turquoise3!90!black,title=\textbf{Intuition}: $\mathbf{\varrho}_{\mathbb{hm}}^{(j)}$]
            Teaching abilities on item $j$ do not change dramatically each lesson, so if human $r_\mathbb{h}$ and model $r_\mathbb{m}$ observers rate teacher $i$ similarly on different lessons ($\mathfrak{L}$ and $\neg\mathfrak{L}$), they are responding to similar observable indicators of the teacher. 
        \end{tcolorbox}
    \end{tcolorbox}
\end{wrapfigure}
\begin{figure*}
    \centering
        \includegraphics[width=1\linewidth]{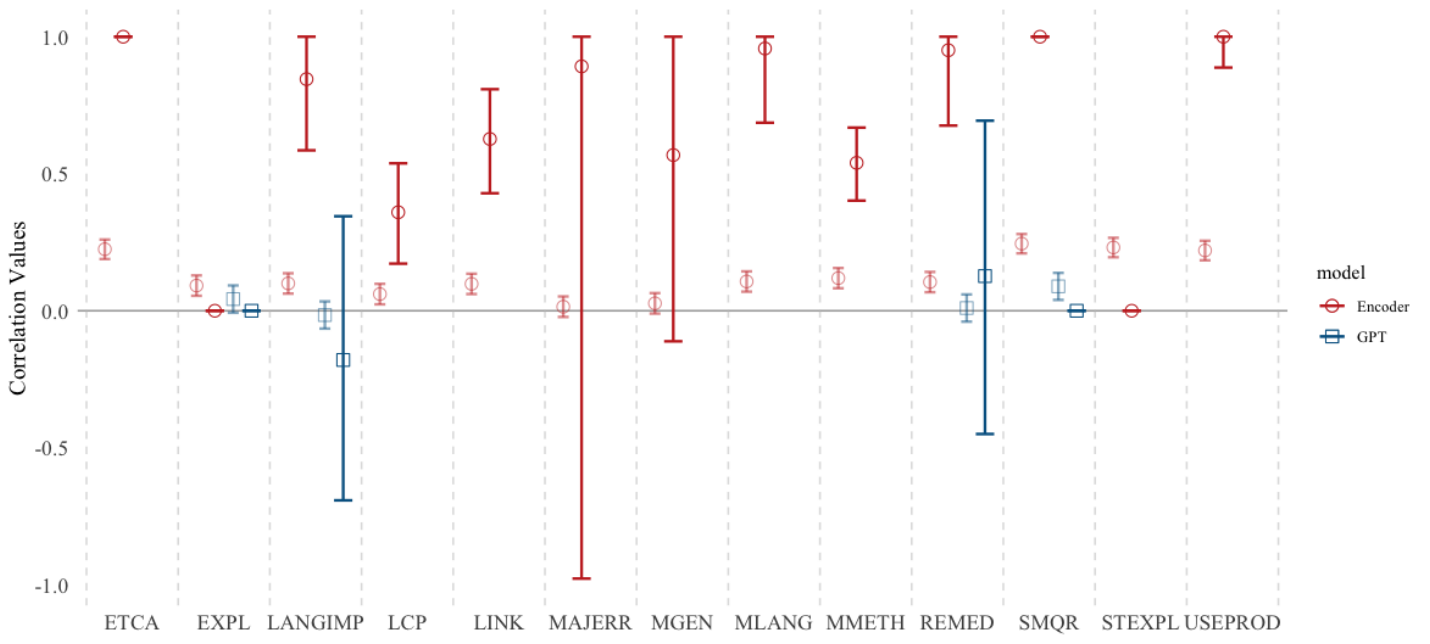}
        \caption{Correlations (\textit{fainter} color hues, numerator of Eq. \ref{eq:disatten}),  disattenuated correlations (\textbf{darker} color hues, Eq. \ref{eq:disatten}), and their respective 95\% confidence intervals between human raters and model raters by MQI item. Item-level rater-label generalizability for both human and model raters, $\mathbf{E}\rho^2$. The attenuated and disattenuated correlations between humans and models $\varrho_{hm}$ are shown. The attenuated correlation confidence intervals were calculated with the standard Fisher Transformation and $\alpha = 0.05$. Disattenuated correlation confidence intervals used the empirical method recommended in \cite{charles_correction_2005}.}
        \label{fig:disatten}
\end{figure*}


where $\Tilde{\mathcal{X}}_\mathbb{F}$ is score retrieval function for individual teacher $i$ on item $j$ by a random member $r$ of rater family $\mathbb{F}$ in relation to some observed lesson $\mathfrak{L}$ with family label generalizability, ${\mathbf{E}\hat{\mathit{\rho}}^2_{\mathbb{F}}}^{(j)}$ defined in Equation \ref{eq:Erho}. In other words, the numerator (represented in red in Figure \ref{fig:disatten}) is the correlation in scores whenever two different lessons from the same teacher were scored by raters from different families (human and model). The denominator then adjusts for based on the reliabilities of raters from each family to account for the known tendency of low reliability to diminish observed correlations.

\begin{figure}
        \centering
        \noindent\makebox[\textwidth]{

        \includegraphics[width=\textwidth]{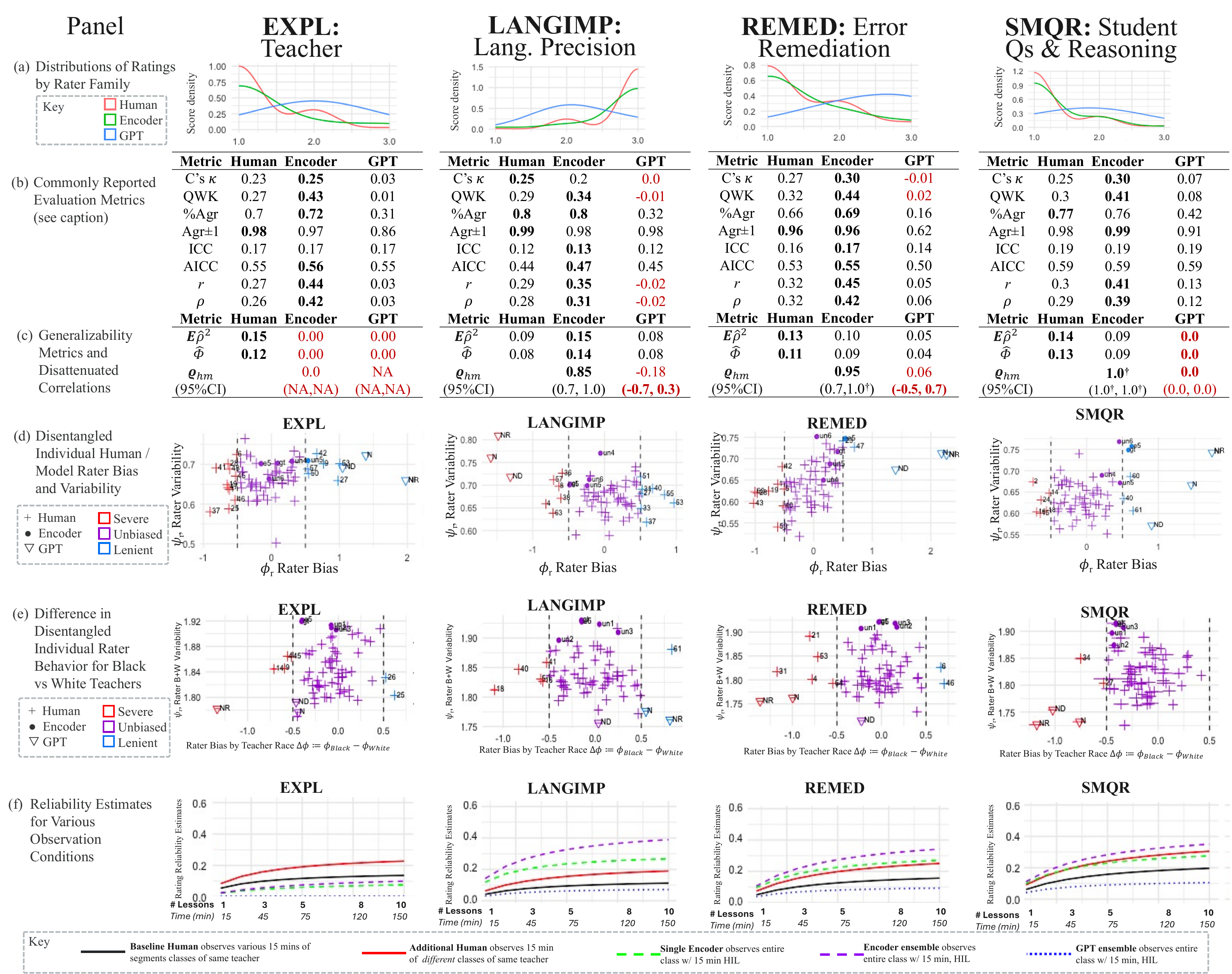} }

        \caption[width=0.7\textwidth]{\textbf{Section \ref{sec:evalmethods} Study Method Results} for four focus MQI Items across Human \citep{kane_national_2015}, Encoder (this study), and GPT \citep{wang_is_2023} rater families.  \textbf{(a)} \textbf{Distributions}. Score distributions by rater type. \textbf{(b)} \textbf{Reliabilities}. Inter-rater reliability metrics introduced in Section \ref{sec:reliabilities}. \textbf{C's $ \kappa$}: Cohen's $ \kappa$; \textbf{QWK}: Quadratic Weighted Kappa; \textbf{\%Agr}: percent exact agreement; \textbf{\%Agr±1}: percent  agreement within 1 category; \textbf{ICC}: intraclass correlation; \textbf{AICC}: adjusted intraclass correlation; \textbf{${r}$}: Pearson's correlation;  $\mathbf{\rho}$: Spearman's rank correlation;  Bold format is highest value for a given metric. \textbf{(c)} \textbf{Generalizability Measures and Spurious Correlation Detection}. Section \ref{sec:gtheorystudy}: generalizability coefficient $\mathbf{E}\rho^2$ and dependability measure $\mathit{\Phi}$. Section \ref{sec:spurious}$: \varrho_{\mathbb{hm}}$ is the disattenuated correlation. Red font indicates correlation was spurious or incalculable due to low reliabilities. \textbf{(d)} \textbf{Disentangled Rater Bias}. Section \ref{sec:biases}: standardized rater bias $\phi_{jr}$ (x axis) and rater variability/consistency, $\psi_{jr}$ (y axis) from Equation \ref{eq:MHRM_SDM}, $\eta_j$-centered. Each point represents an individual human or model rater. More severe raters are left, more lenient right. \textbf{(e)} \textbf{Fairness across Racial Lines}. Section \ref{sec:fairness}: Standardized difference in rater bias $\phi_r$ (x axis) and rater combined variability/consistency, $\psi_r$, (y axis) across Black teachers and White teachers. Leftward values are more severe towards Black teachers, rightward are more lenient. Any horizontal bar present with a marker represents 95\% CI for bias. \textbf{(f)} \textbf{Estimated Improvements to Reliability}. Section \ref{sec:dstudy}: Expected changes to rating reliability are estimated improvements to quality (via reliability) of classroom ratings for various contexts. The single individual human baseline (black) estimates reliability improvements by visiting the same class the x axis represents the number of different 15 min. classroom observations of the same teacher. The red line is estimate of having a \textit{different} human observer conduct observations as described. By contrast, for the model raters--single Encoder (green), Encoder ensemble (average of 3 encoders) (Red), and GPT ensemble (average of 3 GPT prompt engineered models)--the x-axis for models is the number of full classroom observations conducted where the human (black) observes at least 15 minutes (in-the-loop) of the same classroom (models observe the entire class period). A summary of these results can be found in Table \ref{tab:focussummary}.}
        \label{fig:panels}
        
    \end{figure}

Figure \ref{fig:panels} panel (b) has the disattenuated correlations and their respective 95\% confidence intervals, calculated at $\alpha = 0.05$ using empirical confidence scaling methods defined by \cite{charles_correction_2005}, which produces more conservative confidence intervals on this data than traditional Fisher normalization \citep{kromrey_macro_2008}, which is preferable given the low levels of reliability in Section \ref{sec:gtheorystudy} which can lead to overcorrection. Reported disattenuated correlations of 1.0 do not mean perfect correlation: it generally means that \textit{measurement error is not randomly distributed}.  

Disattenuated correlations are not directly comparable\footnote{For example, reported disattenuated correlations of 1.0 do not mean perfect correlation: it generally means that measurement error is not randomly distributed.} to the measures of correlation in Section \ref{sec:reliabilities}  \citep{muchinsky_correction_1996}. However, failure of disattenuation to identify viable human-model correlations for items that previously such showed correlated relationships in Section \ref{sec:reliabilities} suggests the prior correlations may be spurious. Disattenuation does not change the low reliability across items nor the quality of the measurement, but it offers indirect evidence for discerning model predictive validity by quantifying the how changes in the underlying construct result in changes in the same direction for both human and model.

Results for disattenuated correlations described in Section \ref{sec:spurious} and their confidence intervals are in Figure \ref{fig:disatten}. Most items show correlated relationships after disattenuation, and most with confidence intervals above 0.5, suggesting that the encoder models and the humans are likely identifying similar sources of underlying teacher variation for those items. 
\subsubsection{Results}
Disattenuation analyses and Section \ref{sec:gtheorystudy} suggest that the Encoder model family's SOTA-level correlations on the \cmtt{EXPL} and \cmtt{STEXPL} item may have been spurious (likely identifying speech patterns associated with higher teacher performance, and not necessarily specific to explanations), a direct result of low generalizabilities found in Section \ref{sec:gtheorystudy}. Additionally, we see see very large confidence intervals for the encoders for items where item score distributions are most imbalanced (\cmtt{MGEN}, \cmtt{MAJERR}), suggesting that correlations found are not justified in the presence of low reliabilities. Items where the disattenuated correlations are lower (e.g., \cmtt{LCP}, \cmtt{MMETH}) suggests that models and humans interpreted observational features differently. \textbf{\textit{Implications}}: when measurement error is high, disattenuating model and human correlations can help identify whether items with high or similar correlations have spuriousness or are responding to similar features.

This method only minimally provides evidence for investigating accuracy and validity, but, for the Encoder models, evidence can be built upon by comparing how the more continuous ratings of the models and humans change and correlate over the course of a given observation. While not explicitly part of this study, an example of how Encoders' and humans' ratings change from the start to the end of a class for a randomly chosen lesson observation is illustrated in Figure \ref{fig:heartbeats}. Investigating the validity of a construct would require more robust qualitative review of the content.

\newgeometry{margin=0.75cm}

\restoregeometry

\subsection{Bias: Disentangling Individual Rater Behaviors}\label{sec:biases}
\paragraph{RQ 4:} Can bias contributed by individual rater behaviors be identified and disentangled from labels? \textbf{\hypertarget{rq4}{RQ 4:} Case Study Reframe:} How do individual rater effects contribute to ratings bias? 
\subsubsection{Hierarchical Rater Models}
\begin{wrapfigure}[35]{r}{0.4\textwidth}
    \begin{tcolorbox}[colback=Turquoise3!5!white,colframe=Turquoise3!75!black,title={\large \textbf{RQ4: \hypertarget{Bias}{Annotation Bias}}}]
          \begin{tcolorbox}[colback=Turquoise3!10!white,colframe=Turquoise3!90!black,title=\textbf{Method}]

              \textbf{Hierarchical Rater Model}: Three layers of estimation, parameters solved simultaneously (MCMC).

        \end{tcolorbox}
        \begin{tcolorbox}[colback=Turquoise3!1!white,colframe=Turquoise3!90!black,title=\textbf{Top Stage Intuition}]
            The latent teacher abilities $\boldsymbol{\theta}$ are assumed to be normally distributed.
        \end{tcolorbox}
        \begin{tcolorbox}[colback=Turquoise3!1!white,colframe=Turquoise3!90!black,title=\textbf{IRT Stage Intuition}]
            Eq. \ref{eq:MHRM_IRT} estimates the probability of a teacher $i$ receiving an ideal rating $\xi_i$ given the teacher's ability $\boldsymbol{\theta}_i$ and item characteristics ($\boldsymbol{\alpha}$, $\boldsymbol{\gamma}$).
        \end{tcolorbox}
        \begin{tcolorbox}[colback=Turquoise3!1!white,colframe=Turquoise3!90!black,title=\textbf{SDT Stage Intuition}]
            Eq. \ref{eq:MHRM_SDM} estimates the probability that a rater gave teacher $i$ a rating $X_i$ given the ideal rating $\xi_i$ and rater tendencies (bias $\phi$ and variability $\psi^2$).
        \end{tcolorbox}
    \end{tcolorbox}
\end{wrapfigure}
Rater biases in complex tasks are usually not directly measurable, but we can estimate latent constructs that quantify the effects of individual raters' behaviors using methods commonly used to estimate latent attributes of rubric items (e.g., item difficulty) and latent attributes individuals (e.g., ability) throughout Item Response Theory (IRT). If the data had no variation due to raters, various polytomous IRT methods could help estimate "true scores"/"gold" labels ($\xi_{ij}$) during classroom observations, teacher instructional abilities ($\theta_{i}$), and the various individual item effects. 
For tasks with human-mediated labels, human raters introduce additional sources of measurement error for each classification and the data may include multiple measures from multiple raters for a single observation (leading to an accumulation of information at overlap observation points). To address this, hierarchical rater modeling (HRM) \citep{patz_hierarchical_2002,decarlo_using_2003,decarlo_hierarchical_2011} combines an IRT model with a first stage estimation defined by a signal detection theory (SDT) relationship. The latter asks the question, "given the presence of the 'true' score, can a rater detect it?" as the former asks, "given the inputs, can we estimate the 'true' score accounting for differences in the tasks used to measure it?". The hierarchical structure addresses the problem of accumulation of information in the estimates. HRMs consist of three components:
\begin{gather}\label{eq:baseHRM}
\text{HRM}    
\begin{cases}
    \boldsymbol{\theta}_i \sim \text{MVN}(\textbf{0}_{M \times 1},\textbf{I}_{M \times M})
  \text{,}\\
  \xi_{oij} \sim \text{\textbf{IRT model}: Equation \ref{eq:MHRM_IRT}} \\
  X_{soijr} \sim \text{\textbf{SDT model}: Equation \ref{eq:MHRM_SDM}}   
\end{cases}
\end{gather}
where an \textbf{IRT model} estimates the "gold" label score $\xi_{soij}$ for a given item for some time segment $s$ in teacher $i$'s $o$-th observed lesson for item $j$, which arises from $i$'s $M$-dimensionally distributed \textbf{latent instructional ability}/needs ($\boldsymbol{\theta}_i$), and a \textbf{Signal Detection Theory (SDT) model} component disentangles individual rater biases from each recorded score, $X_{soijr}$, by quantifying the latent attributes that mediate whether rater $r$ correctly detects the true score, i.e., $p_{\xi kr}=\ P\left[X_{soijr}=k\ |\xi_{oij}=\xi\ \right]$. 

The IRT component of Equation \ref{eq:baseHRM} estimating the the true scores based on rubric item- and teacher-specific parameters is a  $K_j$-category multidimensional generalized partial credit model (MGPCM) \citep{muraki_generalized_1992,adams_multidimensional_1997, cui_variational_2024,casabianca_digital_2021}. Distributional challenges of negatively worded items can be addressed through a multidimensional parameterization of the underlying latent teacher instructional abilities, with between-item dimensionality confirmatorily defined by the factors in  \cite{blazar_attending_2017}. The MGPCM item discrimination parameters, $\boldsymbol{\alpha}_j  = \alpha_{jm}$, a vector of dimension-specific traits $\boldsymbol{\theta}_i = \theta_{im}$ are separated for $m \in M$ latent dimensions, and parameters for item difficulties $\gamma_{jk}$ exist for each possible score category $k$ in item $j$: 
\begin{align}\label{eq:MHRM_IRT}
 P\left[\xi_{oij}=\xi\ |\boldsymbol{\theta'}_i,\ \boldsymbol{\alpha}_{j\ },\ \gamma_{j\xi}, o\right]=\frac{\exp\left\{(k-1)\boldsymbol{\alpha}_j\boldsymbol{\theta'}_i - \sum_{k=1}^{k} \gamma_{jk}\right\}}{\sum_{h=1}^{K_j}\exp\left\{(k-1)\boldsymbol{\alpha}_j\boldsymbol{\theta'}_i - \sum_{k=1}^{h} \gamma_{jk}\right\}}, \forall s \in o 
\end{align}
where $oi = 1,...,N$ lessons observed for teacher $i$, $j = 1,...,J$ items, $r = 1,...,R$ raters, and $k=1,...,K$ possible scores.  

As parameterized by \cite{patz_hierarchical_2002}, the base-level SDT model of the HRM represents the measurement error induced by rater $r$ whose ability to "detect" the true score changes according to an individual rater's item-specific biases, $\phi_{jr}$ and variabilities, $\psi_{jr}$ , on the x and y axes of Figure \ref{fig:panels}:  
\begin{align}\label{eq:MHRM_SDM}
p_{\xi kr}\propto\exp\left\{-\ \frac{1}{2\psi_{jr}^{2\ }}\left[k-\left(\xi\ +\ \phi_{jr}\right)\right]^2\right\}\ 
\end{align}

where $\boldsymbol{\phi}_{jr} = \textbf{Y}_{jr}\eta$ is a linear model for rating bias for items and with design matrix $\textbf{Y}_{jr}$ of dimensions $(RJ)\times(R+J)$ and $\eta = (\phi_1,...,\phi_R,\eta_1,...\eta_J)^T$  for $R$ raters and $J$ items, as parameterized in \cite{mariano_covariates_2007}. Correspondingly, we update $\ln{\psi_{jr}^2} = \textbf{Y}_{jr} (\ln{\tau^2})$ where $\ln{\mathbf{\tau}^2} = (\ln{\psi_1^2},...,\ln{\psi_R^2},\ln{\tau_1^2},...,\ln{\tau_J^2})^T$ .  The complete rater estimates from these models are displayed in Figure \ref{fig:item_rater_bias}.  The Bayesian estimates were calculated via Markov-chain Monte Carlo (MCMC) simulation using Gibbs sampling across four chains using \texttt{JAGS} \citep{plummer_jags_2003} in \texttt{R} using very weakly-informative priors, converging with $\Hat{R} < 1.1$ for each parameter. A structural plate diagram and \texttt{JAGS} code for the full extended model can be found in Appendix \ref{apx:JAGS}. 

\subsubsection{Results}
Individual annotator tendencies and behaviors can be measured and indiciate significant differences. The vertical dashed lines on the graphs in panels (d) and (e) in Figure \ref{fig:panels} represent 0.5 standard deviations of difference for individual raters from the mean. GPT models show significantly different rater behavior. \textbf{\textit{Implications}}: even tasks where there is minimal overlap of observations to individual raters, behaviors can still be modeled and removed. This allows for improved curation of datasets and model selection.

\subsection{Fairness: Estimation of Ratings Racial Lines}\label{sec:fairness}
\paragraph{RQ 5:}  With unreliable labels and complex tasks, can rater contributions to biased labeling across groups be estimated? \textbf{\hypertarget{rq5}{RQ 5} Case Study Reframe:} Can issues of racial fairness in ratings be disentangled from individual rater behaviors?

\subsubsection{Measuring Racial Discrimination as Rater Covariates}
Disentangling individual rater biases further, across sensitive attributes, can provide a measure of fairness for labels and identify raters (human or model) that display discriminatory biases. Variables representing a sensitive attribute, $\varsigma$ (e.g., race/ethnicity, gender, age, etc.) should be independent of observed score $X_{soijr}$ given the true score $\xi_{soij}$ if ratings are fair: $X \perp \varsigma \Rightarrow P_{\varsigma=a}(X_{jr}|\xi_{j}) = P_{\varsigma=b}(X_{jr}|\xi_{j}),  \forall a,b$ . In the notation used for disentangling rater effects, there should be no difference in variation in scoring from rater $r$ on item $j$ is fair with respect to attribute $\varsigma$ given $\varsigma \perp \xi$:
\begin{align}\label{eq:fair-attribute}
    P[X_{soijr}|\xi_{soij},r, j, \varsigma_i] = P[X_{soijr}|\xi_{soij},r, j]
\end{align}

To measure a rater's item-level fairness with respect to some sensitive teacher attribute, $\varsigma$, the rater parameter vectors are easily updated where $\phi_{jr\varsigma} = \textbf{Y}_{jr\varsigma}\eta$ is now a linear model for rating bias for items and with $\textbf{Y}_{jr\varsigma}$ is a design matrix of dimensions $(RJ\Sigma)\times(R+J + \Sigma)$ and $\Sigma = \{B,W\}$ for Black and White self-identified teachers respectively. In this case, where $\varsigma_i \in \{B,W\}$, we can update the vector explicitly to illustrate those values $\eta = (\phi_{1_B},\dots,\phi_{R_B},\phi_{1_W},\dots,\phi_{R_W},\eta_{1_B},...\eta_{J_B}, \eta_{1_W},...\eta_{J_W})^T$ for $R$ raters, $J$ items, , and , $\ln{\psi_{jr\varsigma}^2} = \textbf{Y}_{jr\varsigma} (\ln{\tau^2})$ is similarly updated such that $\ln{\mathbf{\tau}^2} = (\ln{\psi_1^2},\dots,\ln{\psi_R^2},\ln{\tau_1^2},...,\ln{\tau_J^2, \tau_B^2, \tau_W^2})^T$.  

\begin{wrapfigure}[20]{r}{0.4\textwidth}
    \begin{tcolorbox}[colback=Turquoise3!5!white,colframe=Turquoise3!75!black,title={\large \textbf{RQ5: \hypertarget{Fairness}{Fairness}}}]
          \begin{tcolorbox}[colback=Turquoise3!10!white,colframe=Turquoise3!90!black,title=\textbf{Metric}]
              \textbf{Group $\varsigma$ Independence}: \par 
              \centering $X \perp \varsigma \measeq \phi_B - \phi_W$
        \end{tcolorbox}
        \begin{tcolorbox}[colback=Turquoise3!1!white,colframe=Turquoise3!90!black,title=\textbf{Intuition:} $X \perp \varsigma$]
            Holding a teacher's ideal rating $\xi$ constant for a given rater $r$, a teacher's race ($\varsigma_i \in \{B,W\}$) should be independent of the assigned score $X$. Estimating rater biases directly, \par \centering $\phi_{r,\varsigma = B} - \phi_{r,\varsigma = W} \approxeq 0$. 
        \end{tcolorbox}
    \end{tcolorbox}
\end{wrapfigure}

By approaching the estimation this way, where $\phi_{jr\varsigma}$ is estimated as a parameter, we disentangle contributions to rater scores based on teacher race. This simplifies the task of evaluating for fairness using the metric of group independence, $X \perp \varsigma$, where we can directly calculate $P[X_{soijr}|\xi_{oij},\phi_{jr\varsigma}, \varsigma_i] = P[X_{soijr}|\xi_{oij},\phi_{jr\varsigma}]$.  Thus, $X \perp \varsigma \measeq \phi_B - \phi_W \approxeq 0$.

When estimated, less than 1\% of parameter estimates had $\Hat{R} \ge 1.1$, whose differences in posterior distributions have no material effect on results or discussion; all rater-item-specific 95\% credible intervals for biases are represented as horizontal lines in Figure \ref{fig:panels}, in panel (e). Appendix \ref{apx:JAGS} has full \texttt{JAGS} code used for the formula specification for all items and dimensions, including initial value parameters. Additionally, a plate diagram for MCMC modeling can be found in Figure \ref{fig:plate}.

\subsubsection{Results}
Racial bias at the individual rater level is significiantly measurable. The GPT model families show a negative bias trend against Black teachers relative to White teachers on most items, as seen in the comparison of those models across panels (d) and (e) in Figure \ref{fig:panels}. Potentially more precisely, GPT models' rating centrality seemed to diminish when rating Black teachers, especially with the "reasoning" model, adding evidence that these foundation models may be sensitive to linguistic differences found in African-American English (AAE) \citep{hofmann_dialect_2024,fleisig_linguistic_2024}, possibly due to historical data or models' relative unfamiliarity with AAE \cite{rickford_language_2016}. These results alone should give pause to edtech developers relying on prompt-engineering of foundation LLMs, as subtleties in biases exist in very complex tasks. Additionally, it is not just GPT models showing biases. For some types of items, such as negatively worded items, individual human rater effects could be detected where abnormal rater biases, either positive or negative, towards teachers with some sensitive attribute.  

Overall, encoders displayed much less bias than humans. However, while not as severe as the GPT or human biases, the encoder models did not avoid issues of racial bias. On the worst performing item for both human and encoder models, \cmtt{MGEN}, all of the encoder models found spurious relationships in some language feature while overfitting with a negative bias against Black teachers. The reasons are likely to do with label sparcity and underrepresentativeness across label categories: with so few examples of ratings in the higher categories in the training dataset, overfit on a biased sample was not adequately controlled for, showing a microcosm of alignment to poor data that GPT exhibits in macrocosm. Fortunately for the encoders, many earlier data had already suggested that neither the models nor humans (see Appendix \ref{apx:fullresult} and \cite{hill_when_2012}) could sufficiently distinguish between the item's categories. 

\textbf{\textit{Implications}}: even tasks where there is minimal overlap of observations to individual raters, bias can still be modeled and removed. This allows for improved curation of datasets and model selection. The techniques can be used for evaluation of biases from given populations.

\subsection{Helpfulness: Estimating Real-world of Effects}\label{sec:dstudy}
\paragraph{RQ 6:}  Can we estimate the effects on rating quality and changes in real-world cost if a model were to be used with a human-in-the-loop?  \textbf{\hypertarget{rq6}{RQ 6} Case Study Reframe:} For a teacher, how would automated ratings of instruction affect human rating quality?  

\begin{wrapfigure}[23]{r}{0.4\textwidth}
    \begin{tcolorbox}[colback=Turquoise3!5!white,colframe=Turquoise3!75!black,title={\large \textbf{RQ6: \hypertarget{Helpfulness}{Helpfulness}}}]
          \begin{tcolorbox}[colback=Turquoise3!10!white,colframe=Turquoise3!90!black,title=\textbf{Metric}]
              \textbf{Human-in-the-Loop Dependability}: \par 
              \centering $\widetilde{\Phi}_{j,{\mathbb{F^\prime_\text{HIL}}} \sim \mathbf{K}}$
        \end{tcolorbox}
        \begin{tcolorbox}[colback=Turquoise3!1!white,colframe=Turquoise3!90!black,title=\textbf{Intuition:} $\widetilde{\Phi}_{{\mathbb{F^\prime_\text{HIL}}} \sim \mathbf{K}}$]
            By controlling variance contributions by source, we can estimate how changes (e.g., observing another lesson, using a different rater) would affect the dependability of a rating given to a teacher.
        \end{tcolorbox}
    \end{tcolorbox}
\end{wrapfigure}

\subsubsection{Mixed Decision Studies}
A Decision Study (D-study) estimates how reliabilities of ratings could improve by adjusting measured facets of variation, much like Ho and Kane did to motivate the case study. To estimate the reliability in a human-in-the-loop scenario, multiple g-studies and d-studies would need to be constructed to combine the variance contributions across a set rater families, $\mathbb{F}$. For this work, only two different types of families are consider in each d-study, and one of them will always be human, as automated rating models, even high-performing Encoders, are not yet ready to produce ratings independent from human confirmation. For a human-in-the-loop decision study, $\mathbb{F}$ would consist of families $\mathbb{f}$ that have humans only and models only, and a combined human-model family. For a $(S:O:i)\times R$ study estimated dependability of ratings provided to teachers $i$ on item $j$, $\tilde{\Phi}_{j}$ is, in the joined "universe" $\mathbb{F}^\prime$ where estimations are represented by $\mathbf{K}$, the collection of unique parameterizations and estimates, $\varkappa$, for the facets of variance in each D-study:
\begin{align}\label{eq:dstudyrho}
\widetilde{\Phi}_{j,{\mathbb{F^\prime_\varkappa}} \sim \mathbf{K}}= \frac{\sum_\mathbb{f}^\mathbb{F} {\sigma^2(i_{\varkappa})}_{j\mathbb{f}}}{\sum_\mathbb{f}^\mathbb{F} {\sigma^2(i_{\varkappa})}_{j\mathbb{f}} + {\sigma^2(\Delta_\varkappa)}_{j\mathbb{f}}}
\end{align}
where the summations in Equation \ref{eq:dstudyrho} combines the variation across the familial "universes", indexed by $\varkappa$, of different rater families in  $\mathbb{F}$ and ${\sigma^2(i_\varkappa)}_j$ and ${\sigma^2(\Delta_\varkappa)}_j$ represents the "universe" variability for teacher $i$ and the absolute error for dependability, respectively, at the teacher-year-level ($i$) across the combined parameterization set $\mathbf{K}$. Structurally, Equation \ref{eq:dstudyrho} shares similarities with the two-stage ICC calculation of Eq. \ref{eq:icc}. These values are represented in the ratio for calculating dependability, $\Phi_j$, as found in Equation \ref{eq:phi} ${\sigma^2(\Delta)}_j \equiv \nu_{o:ij}+ \nu_{s:o:ij} + \nu_{irj} + \nu_{rj} + \nu_{s:o:irj}$.  The absolute error for a rater family ($\mathbb{f}$) indexed by $\varkappa$ across any permutation of decision values in this study:  
\begin{align}\label{eq:dstudy_abserror}
    \sigma^2(\Delta_\varkappa) &= \frac{\sigma^2(r_{\varkappa})}{n_{r_{\varkappa}}'} + \frac{\sigma^2(o_{\varkappa}:i)}{n_{o_{\varkappa}}'} + \frac{\sigma^2(r_{\varkappa}i)}{n_{r_{\varkappa}}'} + \frac{\sigma^2(s_\varkappa:o_\varkappa:i)}{n_{s_{\varkappa}}'n_{o_{\varkappa}}'} + \frac{\sigma^2(s_\varkappa:o_\varkappa:ir_\varkappa)}{n_{s_{\varkappa}}'n_{o_{\varkappa}}'n_{r_{\varkappa}}'}
\end{align}
where the decision values vary across design facets and whose contribution is weighted by the combined count $n_k'$ of a given facet $k$ for ratings generated only by the family indexed by $\varkappa$, $n_{k_\varkappa}$ and those facets, if any, shared between families, $n_{k_{\mathbb{F}^\prime}}$: $n_{k_\varkappa}'  = n_{k_\varkappa} + n_{k_{\mathbb{F}^\prime}} \forall k \in \{ s,o,r\}, n_{r_{\mathbb{F}^\prime}} =0$. These distinct sets of parameter values for each design study are represented in Equation \ref{eq:dstudyrho}. 
For human-in-the-loop \textit{only} use cases, $\varkappa_\text{HIL}$, the value $n_{k_{\mathbb{F}^\prime}}$ represents those sources of variation that are shared between rater families, and for a model family $\mathbb{f} = \mathbb{m}$, where there would be no observations made by a model without a human, the model would not have any independent observations $n_{o_\mathbb{m}} = 0 $. To represent these $n$ values where a human $\mathbb{h}$ observes a classroom for 15 minutes\footnote{For the MQI instrument, observation segments are 7.5 minutes long.} with a model and where a single model $\mathbb{m}$ continues to observe for the remainder of the class (an additional 45 minutes), $\mathbf{K}_{n \in \varkappa_\text{HIL}} = \{n_{o_\mathbb{m}} = 0, n_{o_\mathbb{h}} = 0, n_{o_{\mathbb{F}^\prime}} = 1, n_{s_\mathbb{m}} = 6, n_{s_\mathbb{h}} = 0, n_{s_{\mathbb{F}^\prime}} = 2, n_{r_\mathbb{m}} = 1, n_{r_\mathbb{h}} = 1, n_{o_{\mathbb{F}^\prime}} = 0 \}$ and where the variance components are solved similarly to the coefficients of Eq. \ref{eq:gstudy}. 

\subsubsection{Results} 
Estimates of impacts of model use can be reconstructed from measurable variances. The estimates for $\widetilde{\Phi}_{j,{\mathbb{F^\prime}}}$ are in Figure \ref{fig:panels} panel (f) with complete results for all items in Figure \ref{fig:dstudyfullcombined}.  As conducting actual human annotated classroom observation ratings is immensely expensive, the decision study analyses of Section \ref{sec:dstudy} offer methods for estimating the improvement gained by using a model or model family. Parameterizing the decision conditions to reflect "human-in-the-loop" scenarios can even offer insight into whether the variation offered from automated ratings adds or detracts from human rating quality, offering a means of estimating research questions before more expensive trials. 

Constructs that are relatively infrequent, such as \cmtt{LANGIMP}, could greatly benefit automated ratings, since sufficient human observations for identifying that construct would be expensive. Having encoder models listen in for three entire classes yields reliabilities for that construct that are \textbf{\textit{twice}} that of the combined efforts of multiple human raters stopping by a teacher's classroom 10 times, fifteen minutes each time—a net savings of two hours for the principal and a potential savings of over 10 hours if such a level of reliability were desire and were these trends to continue. \textbf{\textit{Implications}}: Not all variance contributes equally, and its careful deconstruction and reconstruction can anticipate future effects before setting up more expensive studies.

\begin{table}[h]
\centering
    \renewcommand{\arraystretch}{2}
    \setlength{\tabcolsep}{1.5pt} 
\begin{tabularx}{0.89\textwidth}{llrcccccccccc} 
\toprule
\multirow{2}{*}{} & \multirow{2}{*}{\textbf{Category}} & \multirow{2}{*}{\textbf{Metric}} 
& \multicolumn{4}{c}{\textbf{GPTs}} & \multicolumn{4}{c}{\textbf{Encoders}} \\ \cmidrule(lr){4-7} \cmidrule(lr){8-11}
& & & {\footnotesize \textbf{\textit{EXPL}}} & {\footnotesize \textbf{\textit{LANGIMP}}} & {\footnotesize \textbf{\textit{REMED}}} & {\footnotesize \textbf{\textit{SMQR}}} 
& {\footnotesize \textbf{\textit{EXPL}}} & {\footnotesize \textbf{\textit{LANGIMP}}} & {\footnotesize \textbf{\textit{REMED}}} & {\footnotesize \textbf{\textit{SMQR}}}  \\ \midrule

\hyperlink{rq1}{\textbf{RQ1}} & \hyperlink{Concordance}{\textbf{Concordance}} & \textit{IRRs} 
& \textcolor{Red3}{\faTimesCircle[regular]} & \textcolor{Red3}{\faTimesCircle[regular]} & \textcolor{Red3}{\faTimesCircle[regular]} & \textcolor{Red3}{\faTimesCircle[regular]} &
\textcolor{Chartreuse4}{\faCheckCircle} & 
\textcolor{Gold3}{\faQuestionCircle[regular]} & 
\textcolor{Chartreuse4}{\faCheckCircle} &  
\textcolor{Chartreuse4}{\faCheckCircle} & 
\\ 

& & \textit{\(r, \rho, \tau\)} 
& \textcolor{Red3}{\faTimesCircle[regular]} & \textcolor{Red3}{\faTimesCircle[regular]} & \textcolor{Red3}{\faTimesCircle[regular]} & \textcolor{Red3}{\faTimesCircle[regular]} &
\textcolor{Chartreuse4}{\faCheckCircle} & 
\textcolor{Chartreuse4}{\faCheckCircle} & 
\textcolor{Chartreuse4}{\faCheckCircle} &  
\textcolor{Chartreuse4}{\faCheckCircle} & 
\\ 

\hyperlink{rq2}{\textbf{RQ2}}& \hyperlink{Confidence}{\textbf{Confidence}} & $\mathbf{E}\rho^2$ 
& \textcolor{Red3}{\faTimesCircle[regular]} & \textcolor{Red3}{\faTimesCircle[regular]} & \textcolor{Red3}{\faTimesCircle[regular]} & \textcolor{Red3}{\faTimesCircle[regular]}  &
\textcolor{Red3}{\faTimesCircle[regular]} & 
\textcolor{Chartreuse4}{\faCheckCircle} & 
\textcolor{Red3}{\faTimesCircle[regular]} &  
\textcolor{Red3}{\faTimesCircle[regular]} & 
\\ 

& & $\Phi$ 
& \textcolor{Red3}{\faTimesCircle[regular]} & \textcolor{Red3}{\faTimesCircle[regular]} & \textcolor{Red3}{\faTimesCircle[regular]} & \textcolor{Red3}{\faTimesCircle[regular]}  &
\textcolor{Red3}{\faTimesCircle[regular]} & 
\textcolor{Chartreuse4}{\faCheckCircle} & 
\textcolor{Red3}{\faTimesCircle[regular]} &  
\textcolor{Red3}{\faTimesCircle[regular]} & 

\\ 

\hyperlink{rq3}{\textbf{RQ3}} & \hyperlink{Validity}{\textbf{Validity}} & $\varrho_{\mathbb{hm}}^{(j)}$ 
& \textcolor{Red3}{\faTimesCircle[regular]} & \textcolor{Red3}{\faTimesCircle[regular]} & \textcolor{Red3}{\faTimesCircle[regular]} & \textcolor{Red3}{\faTimesCircle[regular]}  &
\textcolor{Red3}{\faTimesCircle[regular]} & 
\textcolor{Chartreuse4}{\faCheckCircle} & 
\textcolor{Chartreuse4}{\faCheckCircle} &  
\textcolor{Gold3}{\faQuestionCircle[regular]} & 

\\ 

\hyperlink{rq4}{\textbf{RQ4}} & \hyperlink{Bias}{\textbf{Bias}} & $\phi_r$ 
& \textcolor{Red3}{\faTimesCircle[regular]} & \textcolor{Red3}{\faTimesCircle[regular]} & \textcolor{Red3}{\faTimesCircle[regular]} & \textcolor{Red3}{\faTimesCircle[regular]}  &
\textcolor{Chartreuse4}{\faCheckCircle} & 
\textcolor{Chartreuse4}{\faCheckCircle} & 
\textcolor{Chartreuse4}{\faCheckCircle} &  
\textcolor{Gold3}{\faQuestionCircle[regular]} & 

\\ 

\hyperlink{rq5}{\textbf{RQ5}} & \hyperlink{Fairness}{\textbf{Fairness}} & $X \perp \varsigma$ 
& \textcolor{Gold3}{\faQuestionCircle[regular]} & \textcolor{Red3}{\faTimesCircle[regular]} & \textcolor{Red3}{\faTimesCircle[regular]} & \textcolor{Red3}{\faTimesCircle[regular]}  &
\textcolor{Chartreuse4}{\faCheckCircle} & 
\textcolor{Chartreuse4}{\faCheckCircle} & 
\textcolor{Chartreuse4}{\faCheckCircle} &  
\textcolor{Chartreuse4}{\faCheckCircle} & 

\\ 

\hyperlink{rq6}{\textbf{RQ6}} & \hyperlink{Helpfulness}{\textbf{Helpfulness}} & $\widetilde{\Phi}_{{\mathbb{F^\prime_\text{HIL}}} \sim \mathbf{K}}$ 
& \textcolor{Red3}{\faTimesCircle[regular]} & \textcolor{Red3}{\faTimesCircle[regular]} & \textcolor{Red3}{\faTimesCircle[regular]} & \textcolor{Red3}{\faTimesCircle[regular]}  &
\textcolor{Red3}{\faTimesCircle[regular]} & 
\textcolor{Chartreuse4}{\faCheckCircle} & 
\textcolor{Chartreuse4}{\faCheckCircle} &  
\textcolor{Gold3}{\faQuestionCircle[regular]} & 

\\  \bottomrule \\

\end{tabularx}
\caption{Summary Table for Item-level Metrics and Relative Performance for Model Families on four focus items. \textbf{GPTs} are from \citeauthor{wang_is_2023} and \textbf{Encoders} are from the present study. For each metric, symbols represent whether the model family generally performs as good as or better than humans \textcolor{Chartreuse4}{\faCheckCircle}, worse than humans \textcolor{Red3}{\faTimesCircle[regular]}, or if performance relative to humans is unclear \textcolor{Gold3}{\faQuestionCircle[regular]}. The results for all MQI items can be found in Table \ref{tab:allmqisummary}. \textit{IRRs} refers to the Inter-rater Agreement metrics from Section \ref{sec:reliabilities}.
}\label{tab:focussummary}
\end{table}

\section{Overall Results and Discussion} 
At the outset we asked \textit{How can we know when the behaviors of models are good enough to be used lieu of the humans estimated by \citeauthor{ho_reliability_2013}?} This question, which is a question of validity, is unanswerable by purely empirical means. While reliability (and accuracy) are measurable, validity is a case made from argument. Thus, the answer to that question is not a binary, but one of quality; it is about knowing when the behaviors of models  are "good enough" on \textit{some} item on \textit{some} instrument for \textit{some} population of classrooms against \textit{some} standard of performance. Even though the Encoder family in this study outperform humans, we need to be wary of the validity of the construct being measured, as humans have exhibited the tendency to collaborate poorly with LLM/AI models in their current state \cite{vaccaro_when_2024, agarwal_combining_2023,zhou_relying_2024,azaria_chatgpt_2024,uplevel_gen_2024}. The constraints of human uses demand arguments to validity that are beyond the scope of this work, despite the intentional wording of the primary research question.

The overall results relative to human performance corresponding to each of the research questions and their respective metrics for the four focus MQI items can be found in Table \ref{tab:focussummary} and Table \ref{tab:allmqisummary} has all MQI items. 

For the four focus MQI items, contrasting panel (b) with panel (c) in Figure \ref{fig:panels} reveals commonly used evaluation metrics can obscure important aspects of model performance. However, as demonstrated in panels (c)-(f), there are methods that can be used to improve evaluation under label uncertainty. Many of these methods could be applied to annotated data prior to model training to improve data quality and support training \citep{gordon_jury_2022}.

Encoder models, on most items and in general, outperformed human raters in terms of reduced biases, improved performance metrics, and anticipated cost savings. They represent the best performing models for automated rating of classroom instruction using an authentic measurement instrument of which we are aware at the time of writing, showing large gains over human performance and even larger compared to other models, across metrics discussed herein.  While not the focus of this study, the best reported single metric by \citeauthor{whitehill_automated_2024} on the CLASS rubric across all items and models, $R =0.48$, is contrasted with the average CLASS item performance of the encoders, $\bar{R}=0.60$, and the single worst item for any Encoder model $\min R=0.50$, as reported in the online materials. Thus, the Encoder family models offer a pathway forward for supporting the expensive research task of instructional annotation, regardless of whether they are ready for actual deployment teachers.

This is in stark contrast to the GPT models, which perform much worse than human raters. GPT models likely performed poorly in part due to the prompt length \citep{liu_lost_2023}, the out-of-distribution inputs of elementary school classroom discourse and task of instructional assessment \citep{mccoy_embers_2023}: hypotheses which could be investigated with future research. As GPT-style models increase in popularity, in use, and in sophistication, these methods can help identify sophistry and speciousness in third-party models even in the presence of low reliability. Like humans, models tended to choose a preferred rating value, and their deviations, conditionally informed by billions of fixed parameters at inference, are non-random.\footnote{Variables like `temperature` can increase stochasticity of model outputs.} 

Being able to identify biases in cases of unreliable annotations is important, and researchers should resist the urge to withhold evaluable results from foundation models even if the data fail to reject a null hypothesis. By performing more rigorous evaluations, researchers could crowdsource measuring model biases and behavior tendencies to help all users be more discerning of speciousness, especially as these models' poor behaviors get harder to detect \citep{azaria_chatgpt_2024,hosking_human_2024,zhou_relying_2024} and as researchers make bolder claims about their abilities (see \citealt{binz_centaur_2024}, \textit{inter alia}).

The Encoder models' designs, by contrast, were constructed to allow for multiple methods of interpretability and use by evaluating continuous windows of classroom discourse. This could be used for real-time diagnosis, interpretation, and supporting common understanding between teacher and coach. An example of such use can be found in Figure \ref{fig:heartbeats}, where the continuous predictions for all encoder models are displayed next to average human rating scores. Improvements to this process, combined with successful feature attribution, could boost validity and trust in model use for these high-stake scenarios. If various performance measures continue to display performance Feature attribution (see Appendix \ref{apx:feature_attribution}) could then be used in the future for augmenting transcripts of classroom instruction to support model training and inference.

Automated encoder LLMs could reduce the high costs of improving classroom observers' annotations and serve as a stepping stone to quality teacher development.\footnote{Code for statistical models is available in the appendix and free for use.} Education technologists and EdTech enthusiasts should be wary of foundation models' abilities to do out-of-distribution tasks. These "stochastic parrots" \citep{bender_dangers_2021} might start fires with their "embers of autoregression" \citep{mccoy_embers_2023} when trying to perform tasks for data so far from their training distribution, which is certainly the case with authentic fourth and fifth grade mathematics classroom discourse.

\section{Limitations}

The methods serve as a proof of concept for enhancing reliability in widespread and costly classroom evaluation tasks. Even though these models can perform better than a human given many accepted metrics, much more analysis and technological development is needed. Despite being best in class, these models should not be used in production in their current state. Even with a human in the loop, much more work must be done to ensure their readiness for possible assumed capabilities by end users. Far more important is that GPT style models are not used similarly, and this paper does not endorse their use for this or similar tasks.

Demonstrating multiple methods in a paper with suggestion towards their flexibility evokes the \textit{Garden of Forking Paths Problem}. This study chose to follow the same parameterizations in Section \ref{sec:reliabilities} and data aggregations as the original study \citep{kane_national_2015} in order to preserve comparability with the original data and human raters by using more familiar methods for the context. However, this parameterization has its limitations. An example of where aggregating and calculating reliabilities at the segment level (as was demonstrated in Section \ref{sec:dstudy}) would be to look at reliability and validity issues at the utterance level—something uniquely available to the Encoder model family herein that is not available to other raters or models. Figure \ref{fig:heartbeats} illustrates this capability, underexplored in this paper. Such analyses could be bolstered further by authentic feature attribution for improving interpretability. (See Appendix \ref{apx:feature_attribution} for more on directions for future work implied here.)

While they do demonstrate the claims, the methods of this paper might not be the best implementation of available methods. Rather it is intended to illustrate the potential for better quantifying behaviors in both labelers and models when we have uncertainty in labels. For example, if more understanding of rater perceptions and behaviors of labeling tasks is needed, using a more expressive substitution of Equation \ref{eq:MHRM_SDM} \citep{decarlo_hierarchical_2011,decarlo_classical_2023,decarlo_studies_2008} could give greater insight, especially in the case where models may perceive label category thresholds differently. 

Psychometric models generally assume that the underlying latent variables are distributed normally across a population, which is usually a reasonable assumption with humans. But this assumption need not be true for models nor for all tasks. In this study, few models were estimated alongside humans to demonstrate how differently they behave under this assumption, but this paper provides no evidence that model abilities would be normally distributed for LLMs (e.g., latent constructs could follow multimodal distributions, depending on a family and pretraining, or follow a Normal-exponential-gamma distribution for shifts in metric-specific emergent behaviors). Were researchers interested in modeling learning in a larger population of models, other methods, such as, unipolar IRT models \citep{huang_unipolar_2023}, could potentially help for understanding between-model behaviors for the case where the rating instrument is purely an issue of detection and then magnitude. The usefulness of basic psychometric models presented is based on usefulness of the anthropomorphic distributional comparisons we can reasonably make in the presence of uncertain labels.

The parameters and variables selected for reporting decision study results presented do not represent all use cases and algorithms. While the assumption that models like GPT would have their labels treated as if they were human is a reasonable assumption , it is still an assumption. 
For example, the decision study of Section \ref{sec:dstudy} does not have a within-observation-longitudinal parameterization and thus assumes that humans observing multiple segments of a class period do not necessarily need to observe the segments consecutively. While the MQI rubric is worded so as to be robust to within-lesson autocorrelation, actual lessons are obviously autocorrelated. Longitudinality could likewise support more accurate versions of Equation \ref{eq:MHRM_IRT}.

While many studies cited herein seek to generalize similar research across all classrooms, we acknowledge that this cannot be done with the transcript data we use for this presented work, as it only consists of fourth and fifth-grade mathematics classrooms from the United States. While the methods potentially possess broad applicability across all grades and subject areas, the current models lack generalizability beyond elementary mathematics classrooms in U.S. public schools, highlighting the need for more publicly available data in this area. Furthermore, the associated ratings and reliability metrics pertain solely to a subset of rating items on the MQI rubric\footnote{The full set of items from MQI and CLASS rubrics are available in Appendices and in the online materials.}, which may introduce limitations when addressing the more universal task of automated instruction ratings. This is associated with the limitations of the instruments themselves, as imperfect tools for even calibrated and trained raters.

Similarly, as the focus of this paper is to demonstrate evaluation techniques in the presence of unreliable labels, the generalizability of models is low. Encoder models, while each is powerful and individually able to produce automated scores for 25 different authentic measures of classroom instruction (in contrast to the models of \citeauthor{xu_promises_2024}, which used 11 separate fine-tuned models for the MQI items evaluated), were built specifically for this task and would not generalize further without data or architecture changes. GPT models represent available autoregressive decoder in-context learning via prompt engineering in 2023. Models have scaled and improved since then and it is possible that performance would improve, but issues of underlying racial biases (Section \ref{sec:fairness}) continue to exist, even with more current models \citep{hofmann_dialect_2024,hofmann_ai_2024, warr_implicit_2024,shieh_laissez-faire_2024,nghiem_you_2024, henderson_safety_2024}. 

The Encoder models were trained under the assumptions that the actual expert human ratings are not very reliable, that the alignment of the coordination of timing across rubrics and across transcripts is imperfect, that the discourse transcripts are imperfect, and that information is lost by keeping fixed sentence-level embeddings. While the methods outlined worked to extract a meaningful signal despite these challenges, it should be noted that the signal is still trained on noisy human ratings. If, on average, the raters had a particular bias, the model would carry that bias. For example,  this is particularly true with the CLASS item ratings, as there were only 19 different raters used, compared to the 63 used for the MQI rubric items, and only had one rater per classroom observation. Results are included for comparability and generalizability, but they likely carry more human raters' idiosyncrasies. 

The encoder models removed transcription notes and intentionally did not use transcription information (such as identification of speaker) to best emulate what the functionality would be in a audio-input-only setup. While this is an authentic interpretation of the task, the transcription process was still done with humans. While direct input from audio would capture even more information (such as tone or long breaks in speaking for independent work), these models have not been trained to work with automated transcription. 

The encoder models could be improved through metalearning training, so they could be more adaptive to new instructional rubrics and classrooms. Without metalearning across tasks, transferability is limited by the training regime and architecture as well as the data. Future work will include metalearning, allowing the model to take advantage of 72\% more observations.   

Finally, while the paper reported on "GPT" family performance, it only used the performance corresponding to a since study, which used only prompt engineering and which used ChatGPT 3.5. Perhaps with fine-tuning, multi-agent prompting, and other enhanced uses of such models, performance might improve. However, it is not clear that, even as models continue to improve on general use tasks, that they will improve on their ability to understand and respond to text that is outside of their training distribution (i.e., classroom discourse). Even if the text were within the training distribution, this study has demonstrated that evaluation of such text is non-trivial and, thus, the task would still be more challenging for such models \citep{mccoy_embers_2023}.

\section{Authorship and Positionality Statement}\label{sec:auth}
Michael Hardy is the sole author of this work. Prior to his research work, he worked in public education as a teacher, principal, superintendent, and a state chief, where he evaluated and improved instructional materials and practices across many contexts. With more than decade of successful coaching instruction and as a former Educator of the Year for Texas, he is compelled by his passion and expertise to improve and support classroom teachers so that all students can have access to an excellent education. Third-party generative language models, such as ChatGPT, were not used for any aspect of the study, except where explicitly stated. 

\newgeometry{margin=2 cm}
\begin{landscape}

\begin{table}[h]
\centering
    \renewcommand{\arraystretch}{2}
    \setlength{\tabcolsep}{1.5pt} 
\begin{tabularx}{0.85\paperheight}{llrccccccccccccccccccc} 
\toprule
\multirow{2}{*}{} & \multirow{2}{*}{\textbf{Category}} & \multirow{2}{*}{\textbf{Metric}} 
& \multicolumn{4}{c}{\textbf{GPTs}} & \multicolumn{13}{c}{\textbf{Encoders}} \\ \cmidrule(lr){4-7} \cmidrule(lr){8-20}
& & & {\footnotesize \textbf{\textit{EXPL}}} & {\footnotesize \textbf{\textit{LANGIMP}}} & {\footnotesize \textbf{\textit{REMED}}} & {\footnotesize \textbf{\textit{SMQR}}} 
& {\footnotesize \textbf{\textit{EXPL}}} & {\footnotesize \textbf{\textit{LANGIMP}}} & {\footnotesize \textbf{\textit{REMED}}} & {\footnotesize \textbf{\textit{SMQR}}} 
& {\footnotesize ETCA} & {\footnotesize LCP} & {\footnotesize LINK} & {\footnotesize MAJERR} 
& {\footnotesize MGEN} & {\footnotesize MLANG} & {\footnotesize MMETH} & {\footnotesize STEXPL} & {\footnotesize USEPROD} \\ \midrule

\hyperlink{rq1}{\textbf{RQ1}} & \hyperlink{Concordance}{\textbf{Concordance}} & \textit{IRRs} 
& \textcolor{Red3}{\faTimesCircle[regular]} & \textcolor{Red3}{\faTimesCircle[regular]} & \textcolor{Red3}{\faTimesCircle[regular]} & \textcolor{Red3}{\faTimesCircle[regular]} &
\textcolor{Chartreuse4}{\faCheckCircle} & 
\textcolor{Gold3}{\faQuestionCircle[regular]} & 
\textcolor{Chartreuse4}{\faCheckCircle} &  
\textcolor{Chartreuse4}{\faCheckCircle} & 
\textcolor{Chartreuse4}{\faCheckCircle} & 
\textcolor{Chartreuse4}{\faCheckCircle} & 
\textcolor{Chartreuse4}{\faCheckCircle} & 
\textcolor{Gold3}{\faQuestionCircle[regular]} & 
\textcolor{Chartreuse4}{\faCheckCircle} & 
\textcolor{Chartreuse4}{\faCheckCircle} &  
\textcolor{Gold3}{\faQuestionCircle[regular]} &  
\textcolor{Gold3}{\faQuestionCircle[regular]} & 
\textcolor{Chartreuse4}{\faCheckCircle} 
\\ 

& & \textit{\(r, \rho, \tau\)} 
& \textcolor{Red3}{\faTimesCircle[regular]} & \textcolor{Red3}{\faTimesCircle[regular]} & \textcolor{Red3}{\faTimesCircle[regular]} & \textcolor{Red3}{\faTimesCircle[regular]} &
\textcolor{Chartreuse4}{\faCheckCircle} & 
\textcolor{Chartreuse4}{\faCheckCircle} & 
\textcolor{Chartreuse4}{\faCheckCircle} &  
\textcolor{Chartreuse4}{\faCheckCircle} & 
\textcolor{Chartreuse4}{\faCheckCircle} & 
\textcolor{Chartreuse4}{\faCheckCircle} & 
\textcolor{Chartreuse4}{\faCheckCircle} & 
\textcolor{Chartreuse4}{\faCheckCircle} & 
\textcolor{Chartreuse4}{\faCheckCircle} & 
\textcolor{Chartreuse4}{\faCheckCircle} &  
\textcolor{Chartreuse4}{\faCheckCircle} &  
\textcolor{Chartreuse4}{\faCheckCircle} & 
\textcolor{Chartreuse4}{\faCheckCircle} 
\\ 

\hyperlink{rq2}{\textbf{RQ2}}& \hyperlink{Confidence}{\textbf{Confidence}} & $\mathbf{E}\rho^2$ 
& \textcolor{Red3}{\faTimesCircle[regular]} & \textcolor{Red3}{\faTimesCircle[regular]} & \textcolor{Red3}{\faTimesCircle[regular]} & \textcolor{Red3}{\faTimesCircle[regular]}  &
\textcolor{Red3}{\faTimesCircle[regular]} & 
\textcolor{Chartreuse4}{\faCheckCircle} & 
\textcolor{Red3}{\faTimesCircle[regular]} &  
\textcolor{Red3}{\faTimesCircle[regular]} & 
\textcolor{Chartreuse4}{\faCheckCircle} & 
\textcolor{Chartreuse4}{\faCheckCircle} & 
\textcolor{Chartreuse4}{\faCheckCircle} & 
\textcolor{Red3}{\faTimesCircle[regular]} & 
\textcolor{Chartreuse4}{\faCheckCircle} & 
\textcolor{Chartreuse4}{\faCheckCircle} &  
\textcolor{Chartreuse4}{\faCheckCircle} &  
\textcolor{Red3}{\faTimesCircle[regular]} & 
\textcolor{Chartreuse4}{\faCheckCircle} 
\\ 

& & $\Phi$ 
& \textcolor{Red3}{\faTimesCircle[regular]} & \textcolor{Red3}{\faTimesCircle[regular]} & \textcolor{Red3}{\faTimesCircle[regular]} & \textcolor{Red3}{\faTimesCircle[regular]}  &
\textcolor{Red3}{\faTimesCircle[regular]} & 
\textcolor{Chartreuse4}{\faCheckCircle} & 
\textcolor{Red3}{\faTimesCircle[regular]} &  
\textcolor{Red3}{\faTimesCircle[regular]} & 
\textcolor{Chartreuse4}{\faCheckCircle} & 
\textcolor{Chartreuse4}{\faCheckCircle} & 
\textcolor{Chartreuse4}{\faCheckCircle} & 
\textcolor{Red3}{\faTimesCircle[regular]}& 
\textcolor{Chartreuse4}{\faCheckCircle} & 
\textcolor{Chartreuse4}{\faCheckCircle} &  
\textcolor{Chartreuse4}{\faCheckCircle} &  
\textcolor{Red3}{\faTimesCircle[regular]} & 
\textcolor{Chartreuse4}{\faCheckCircle} 
\\ 

\hyperlink{rq3}{\textbf{RQ3}} & \hyperlink{Validity}{\textbf{Validity}} & $\varrho_{\mathbb{hm}}^{(j)}$ 
& \textcolor{Red3}{\faTimesCircle[regular]} & \textcolor{Red3}{\faTimesCircle[regular]} & \textcolor{Red3}{\faTimesCircle[regular]} & \textcolor{Red3}{\faTimesCircle[regular]}  &
\textcolor{Red3}{\faTimesCircle[regular]} & 
\textcolor{Chartreuse4}{\faCheckCircle} & 
\textcolor{Chartreuse4}{\faCheckCircle} &  
\textcolor{Gold3}{\faQuestionCircle[regular]} & 
\textcolor{Chartreuse4}{\faCheckCircle} & 
\textcolor{Chartreuse4}{\faCheckCircle} & 
\textcolor{Chartreuse4}{\faCheckCircle} & 
\textcolor{Chartreuse4}{\faCheckCircle} & 
\textcolor{Chartreuse4}{\faCheckCircle} & 
\textcolor{Chartreuse4}{\faCheckCircle} &  
\textcolor{Chartreuse4}{\faCheckCircle} &  
\textcolor{Red3}{\faTimesCircle[regular]} & 
\textcolor{Chartreuse4}{\faCheckCircle} 
\\ 

\hyperlink{rq4}{\textbf{RQ4}} & \hyperlink{Bias}{\textbf{Bias}} & $\phi_r$ 
& \textcolor{Red3}{\faTimesCircle[regular]} & \textcolor{Red3}{\faTimesCircle[regular]} & \textcolor{Red3}{\faTimesCircle[regular]} & \textcolor{Red3}{\faTimesCircle[regular]}  &
\textcolor{Chartreuse4}{\faCheckCircle} & 
\textcolor{Chartreuse4}{\faCheckCircle} & 
\textcolor{Chartreuse4}{\faCheckCircle} &  
\textcolor{Gold3}{\faQuestionCircle[regular]} & 
\textcolor{Chartreuse4}{\faCheckCircle} & 
\textcolor{Gold3}{\faQuestionCircle[regular]} & 
\textcolor{Chartreuse4}{\faCheckCircle} & 
\textcolor{Red3}{\faTimesCircle[regular]} & 
\textcolor{Red3}{\faTimesCircle[regular]} & 
\textcolor{Chartreuse4}{\faCheckCircle} &  
\textcolor{Gold3}{\faQuestionCircle[regular]} &  
\textcolor{Chartreuse4}{\faCheckCircle} & 
\textcolor{Gold3}{\faQuestionCircle[regular]} 
\\ 

\hyperlink{rq5}{\textbf{RQ5}} & \hyperlink{Fairness}{\textbf{Fairness}} & $X \perp \varsigma$ 
& \textcolor{Gold3}{\faQuestionCircle[regular]} & \textcolor{Red3}{\faTimesCircle[regular]} & \textcolor{Red3}{\faTimesCircle[regular]} & \textcolor{Red3}{\faTimesCircle[regular]}  &
\textcolor{Chartreuse4}{\faCheckCircle} & 
\textcolor{Chartreuse4}{\faCheckCircle} & 
\textcolor{Chartreuse4}{\faCheckCircle} &  
\textcolor{Chartreuse4}{\faCheckCircle} & 
\textcolor{Chartreuse4}{\faCheckCircle} & 
\textcolor{Chartreuse4}{\faCheckCircle} & 
\textcolor{Chartreuse4}{\faCheckCircle} & 
\textcolor{Chartreuse4}{\faCheckCircle} & 
\textcolor{Red3}{\faTimesCircle[regular]} & 
\textcolor{Chartreuse4}{\faCheckCircle} &  
\textcolor{Gold3}{\faQuestionCircle[regular]} &  
\textcolor{Chartreuse4}{\faCheckCircle} & 
\textcolor{Chartreuse4}{\faCheckCircle} 
\\ 

\hyperlink{rq6}{\textbf{RQ6}} & \hyperlink{Helpfulness}{\textbf{Helpfulness}} & $\widetilde{\Phi}_{{\mathbb{F^\prime_\text{HIL}}} \sim \mathbf{K}}$ 
& \textcolor{Red3}{\faTimesCircle[regular]} & \textcolor{Red3}{\faTimesCircle[regular]} & \textcolor{Red3}{\faTimesCircle[regular]} & \textcolor{Red3}{\faTimesCircle[regular]}  &
\textcolor{Red3}{\faTimesCircle[regular]} & 
\textcolor{Chartreuse4}{\faCheckCircle} & 
\textcolor{Chartreuse4}{\faCheckCircle} &  
\textcolor{Gold3}{\faQuestionCircle[regular]} & 
\textcolor{Chartreuse4}{\faCheckCircle} & 
\textcolor{Chartreuse4}{\faCheckCircle} & 
\textcolor{Chartreuse4}{\faCheckCircle} & 
\textcolor{Gold3}{\faQuestionCircle[regular]} & 
\textcolor{Red3}{\faTimesCircle[regular]} & 
\textcolor{Chartreuse4}{\faCheckCircle} &  
\textcolor{Chartreuse4}{\faCheckCircle} &  
\textcolor{Red3}{\faTimesCircle[regular]} & 
\textcolor{Chartreuse4}{\faCheckCircle} 
\\  \bottomrule \\

\end{tabularx}
\caption{Summary Table for All MQI Item-level Metrics and Relative Performance for Model Families. \textbf{GPTs} are from \citeauthor{wang_is_2023} and \textbf{Encoders} are from the present study. For each metric, symbols represent whether the model family generally performs as good as or better than humans \textcolor{Chartreuse4}{\faCheckCircle}, worse than humans \textcolor{Red3}{\faTimesCircle[regular]}, or if performance relative to humans is unclear \textcolor{Gold3}{\faQuestionCircle[regular]}. \textbf{\textit{Bold Italicized Items}} represent the four MQI items tested by \citeauthor{wang_is_2023}.
} \label{tab:allmqisummary}
\end{table}

\end{landscape}
\restoregeometry

\bibliographystyle{acl_natbib}
\bibliography{references, anthology_p1, anthology_p2} 

\begin{thebibliography}{147}
\providecommand{\natexlab}[1]{#1}

\bibitem[{Abercrombie et~al.(2023)Abercrombie, Rieser, and Hovy}]{abercrombie_consistency_2023}
Gavin Abercrombie, Verena Rieser, and Dirk Hovy. 2023.
\newblock \href {https://doi.org/10.48550/arXiv.2301.10684} {Consistency is {Key}: {Disentangling} {Label} {Variation} in {Natural} {Language} {Processing} with {Intra}-{Annotator} {Agreement}}.
\newblock \emph{arXiv preprint}.
\newblock ArXiv:2301.10684 [cs].

\bibitem[{Adams et~al.(1997)Adams, Wilson, and Wang}]{adams_multidimensional_1997}
Raymond~J. Adams, Mark Wilson, and Wen-chung Wang. 1997.
\newblock \href {https://doi.org/10.1177/0146621697211001} {The multidimensional random coefficients multinomial logit model}.
\newblock \emph{Applied Psychological Measurement}, 21(1):1--23.
\newblock Place: US Publisher: Sage Publications.

\bibitem[{Adebayo et~al.(2020)Adebayo, Gilmer, Muelly, Goodfellow, Hardt, and Kim}]{adebayo_sanity_2020}
Julius Adebayo, Justin Gilmer, Michael Muelly, Ian Goodfellow, Moritz Hardt, and Been Kim. 2020.
\newblock \href {https://doi.org/10.48550/arXiv.1810.03292} {Sanity {Checks} for {Saliency} {Maps}}.
\newblock \emph{arXiv preprint}.
\newblock ArXiv:1810.03292 [cs, stat].

\bibitem[{Agarwal et~al.(2023)Agarwal, Moehring, Rajpurkar, and Salz}]{agarwal_combining_2023}
Nikhil Agarwal, Alex Moehring, Pranav Rajpurkar, and Tobias Salz. 2023.
\newblock \href {https://doi.org/10.3386/w31422} {Combining {Human} {Expertise} with {Artificial} {Intelligence}: {Experimental} {Evidence} from {Radiology}}.

\bibitem[{Aguilar(2013)}]{aguilar_developing_2013}
Elena Aguilar. 2013.
\newblock Developing a {Work} {Plan}: {How} {Do} {I} {Determine} {What} to {Do}?
\newblock In \emph{The art of coaching: effective strategies for school transformation}, pages 119--144. Jossey-Bass, A Wiley Brand, San Francisco.

\bibitem[{Alic et~al.(2022)Alic, Demszky, Mancenido, Liu, Hill, and Jurafsky}]{alic-etal-2022-computationally}
Sterling Alic, Dorottya Demszky, Zid Mancenido, Jing Liu, Heather Hill, and Dan Jurafsky. 2022.
\newblock \href {https://doi.org/10.18653/v1/2022.bea-1.27} {Computationally identifying funneling and focusing questions in classroom discourse}.
\newblock In \emph{Proceedings of the 17th Workshop on Innovative Use of NLP for Building Educational Applications (BEA 2022)}, pages 224--233, Seattle, Washington. Association for Computational Linguistics.

\bibitem[{Azaria et~al.(2024)Azaria, Azoulay, and Reches}]{azaria_chatgpt_2024}
Amos Azaria, Rina Azoulay, and Shulamit Reches. 2024.
\newblock \href {https://doi.org/10.1162/dint_a_00235} {{ChatGPT} is a {Remarkable} {Tool}—{For} {Experts}}.
\newblock \emph{Data Intelligence}, 6(1):240--296.

\bibitem[{Baan et~al.(2022)Baan, Aziz, Plank, and Fernández}]{baan_stop_2022}
Joris Baan, Wilker Aziz, Barbara Plank, and Raquel Fernández. 2022.
\newblock \href {https://doi.org/10.48550/arXiv.2210.16133} {Stop {Measuring} {Calibration} {When} {Humans} {Disagree}}.
\newblock \emph{arXiv preprint}.
\newblock ArXiv:2210.16133 [cs].

\bibitem[{Baan et~al.(2024)Baan, Fernández, Plank, and Aziz}]{baan_interpreting_2024}
Joris Baan, Raquel Fernández, Barbara Plank, and Wilker Aziz. 2024.
\newblock \href {https://doi.org/10.48550/arXiv.2402.16102} {Interpreting {Predictive} {Probabilities}: {Model} {Confidence} or {Human} {Label} {Variation}?}
\newblock \emph{arXiv preprint}.
\newblock ArXiv:2402.16102 [cs] version: 1.

\bibitem[{Bacher-Hicks et~al.(2017)Bacher-Hicks, Chin, Kane, and Staiger}]{bacher-hicks_evaluation_2017}
Andrew Bacher-Hicks, Mark~J. Chin, Thomas~J. Kane, and Douglas~O. Staiger. 2017.
\newblock \href {https://doi.org/10.3386/w23478} {An {Evaluation} of {Bias} in {Three} {Measures} of {Teacher} {Quality}: {Value}-{Added}, {Classroom} {Observations}, and {Student} {Surveys}}.

\bibitem[{Bacher-Hicks et~al.(2019)Bacher-Hicks, Chin, Kane, and Staiger}]{bacher-hicks_experimental_2019}
Andrew Bacher-Hicks, Mark~J. Chin, Thomas~J. Kane, and Douglas~O. Staiger. 2019.
\newblock \href {https://doi.org/10.1016/j.econedurev.2019.101919} {An experimental evaluation of three teacher quality measures: {Value}-added, classroom observations, and student surveys}.
\newblock \emph{Economics of Education Review}, 73:101919.

\bibitem[{Bambrick-Santoyo(2016)}]{bambrick-santoyo_get_2016}
Paul Bambrick-Santoyo. 2016.
\newblock \emph{Get better faster: a 90-day plan for coaching new teachers}.
\newblock Jossey-Bass, A Wiley Brand, San Francisco, CA.

\bibitem[{Bambrick-Santoyo(2018)}]{bambrick-santoyo_leverage_2018}
Paul Bambrick-Santoyo. 2018.
\newblock \emph{Leverage leadership 2.0: a practical guide to building exceptional schools}.
\newblock Jossey-Bass, San Francisco, CA.

\bibitem[{Bates et~al.(2015)Bates, Mächler, Bolker, and Walker}]{bates_fitting_2015}
Douglas Bates, Martin Mächler, Ben Bolker, and Steve Walker. 2015.
\newblock \href {https://doi.org/10.18637/jss.v067.i01} {Fitting {Linear} {Mixed}-{Effects} {Models} {Using} lme4}.
\newblock \emph{Journal of Statistical Software}, 67:1--48.

\bibitem[{Bejar et~al.(2006)Bejar, Williamson, and Mislevy}]{bejar_human_2006}
Isaac~1 Bejar, David~M. Williamson, and {and} Robert~J. Mislevy. 2006.
\newblock Human {Scoring}.
\newblock In \emph{Automated {Scoring} of {Complex} {Tasks} in {Computer}-{Based} {Testing}}. Routledge.
\newblock Num Pages: 34.

\bibitem[{Belz et~al.(2020)Belz, Mille, and Howcroft}]{belz_disentangling_2020}
Anya Belz, Simon Mille, and David~M. Howcroft. 2020.
\newblock \href {https://doi.org/10.18653/v1/2020.inlg-1.24} {Disentangling the {Properties} of {Human} {Evaluation} {Methods}: {A} {Classification} {System} to {Support} {Comparability}, {Meta}-{Evaluation} and {Reproducibility} {Testing}}.
\newblock In \emph{Proceedings of the 13th {International} {Conference} on {Natural} {Language} {Generation}}, pages 183--194, Dublin, Ireland. Association for Computational Linguistics.

\bibitem[{Belz et~al.(2023)Belz, Thomson, Reiter, and Mille}]{belz_non-repeatable_2023}
Anya Belz, Craig Thomson, Ehud Reiter, and Simon Mille. 2023.
\newblock \href {https://doi.org/10.18653/v1/2023.findings-acl.226} {Non-{Repeatable} {Experiments} and {Non}-{Reproducible} {Results}: {The} {Reproducibility} {Crisis} in {Human} {Evaluation} in {NLP}}.
\newblock In \emph{Findings of the {Association} for {Computational} {Linguistics}: {ACL} 2023}, pages 3676--3687, Toronto, Canada. Association for Computational Linguistics.

\bibitem[{Bender et~al.(2021)Bender, Gebru, McMillan-Major, and Shmitchell}]{bender_dangers_2021}
Emily~M. Bender, Timnit Gebru, Angelina McMillan-Major, and Shmargaret Shmitchell. 2021.
\newblock \href {https://doi.org/10.1145/3442188.3445922} {On the {Dangers} of {Stochastic} {Parrots}: {Can} {Language} {Models} {Be} {Too} {Big}?}
\newblock In \emph{Proceedings of the 2021 {ACM} {Conference} on {Fairness}, {Accountability}, and {Transparency}}, {FAccT} '21, pages 610--623, New York, NY, USA. Association for Computing Machinery.

\bibitem[{Binz et~al.(2024)Binz, Akata, Bethge, Brändle, Callaway, Coda-Forno, Dayan, Demircan, Eckstein, Éltető, Griffiths, Haridi, Jagadish, Ji-An, Kipnis, Kumar, Ludwig, Mathony, Mattar, Modirshanechi, Nath, Peterson, Rmus, Russek, Saanum, Scharfenberg, Schubert, Buschoff, Singhi, Sui, Thalmann, Theis, Truong, Udandarao, Voudouris, Wilson, Witte, Wu, Wulff, Xiong, and Schulz}]{binz_centaur_2024}
Marcel Binz, Elif Akata, Matthias Bethge, Franziska Brändle, Fred Callaway, Julian Coda-Forno, Peter Dayan, Can Demircan, Maria~K. Eckstein, Noémi Éltető, Thomas~L. Griffiths, Susanne Haridi, Akshay~K. Jagadish, Li~Ji-An, Alexander Kipnis, Sreejan Kumar, Tobias Ludwig, Marvin Mathony, Marcelo Mattar, Alireza Modirshanechi, Surabhi~S. Nath, Joshua~C. Peterson, Milena Rmus, Evan~M. Russek, Tankred Saanum, Natalia Scharfenberg, Johannes~A. Schubert, Luca M.~Schulze Buschoff, Nishad Singhi, Xin Sui, Mirko Thalmann, Fabian Theis, Vuong Truong, Vishaal Udandarao, Konstantinos Voudouris, Robert Wilson, Kristin Witte, Shuchen Wu, Dirk Wulff, Huadong Xiong, and Eric Schulz. 2024.
\newblock \href {https://doi.org/10.48550/arXiv.2410.20268} {Centaur: a foundation model of human cognition}.
\newblock \emph{arXiv preprint}.
\newblock ArXiv:2410.20268.

\bibitem[{Birhane et~al.(2022)Birhane, Kalluri, Card, Agnew, Dotan, and Bao}]{birhane_values_2022}
Abeba Birhane, Pratyusha Kalluri, Dallas Card, William Agnew, Ravit Dotan, and Michelle Bao. 2022.
\newblock \href {https://doi.org/10.48550/arXiv.2106.15590} {The {Values} {Encoded} in {Machine} {Learning} {Research}}.
\newblock \emph{arXiv preprint}.
\newblock ArXiv:2106.15590.

\bibitem[{Blazar(2018)}]{blazar_validating_2018}
David Blazar. 2018.
\newblock \href {https://doi.org/10.1162/edfp_a_00251} {Validating {Teacher} {Effects} on {Students}’ {Attitudes} and {Behaviors}: {Evidence} from {Random} {Assignment} of {Teachers} to {Students}}.
\newblock \emph{Education Finance and Policy}, 13(3):281--309.

\bibitem[{Blazar et~al.(2017)Blazar, Braslow, Charalambous, and Hill}]{blazar_attending_2017}
David Blazar, David Braslow, Charalambos~Y. Charalambous, and Heather~C. Hill. 2017.
\newblock \href {https://doi.org/10.1080/10627197.2017.1309274} {Attending to {General} and {Mathematics}-{Specific} {Dimensions} of {Teaching}: {Exploring} {Factors} {Across} {Two} {Observation} {Instruments}}.
\newblock \emph{Educational Assessment}, 22(2):71--94.
\newblock Publisher: Routledge \_eprint: https://doi.org/10.1080/10627197.2017.1309274.

\bibitem[{Blazar and Pollard(2022)}]{blazar_challenges_2022}
David Blazar and Cynthia Pollard. 2022.
\newblock \href {https://www.edworkingpapers.com/ai22-591} {Challenges and {Tradeoffs} of “{Good}” {Teaching}: {The} {Pursuit} of {Multiple} {Educational} {Outcomes}}.
\newblock Technical report, Annenberg Institute at Brown University.
\newblock Publication Title: EdWorkingPapers.com.

\bibitem[{Brennan(2001{\natexlab{a}})}]{brennan_generalizability_2001}
Robert~L. Brennan. 2001{\natexlab{a}}.
\newblock \href {https://doi.org/10.1007/978-1-4757-3456-0} {\emph{Generalizability {Theory}}}.
\newblock Springer, New York, NY.

\bibitem[{Brennan(2001{\natexlab{b}})}]{brennan_variability_2001}
Robert~L. Brennan. 2001{\natexlab{b}}.
\newblock \href {https://doi.org/10.1007/978-1-4757-3456-0_6} {Variability of {Statistics} in {Generalizability} {Theory}}.
\newblock In Robert~L. Brennan, editor, \emph{Generalizability {Theory}}, Statistics for {Social} {Sciences} and {Public} {Policy}, pages 179--213. Springer, New York, NY.

\bibitem[{Brennan(2013)}]{brennan_generalizability_2013}
Robert~L. Brennan. 2013.
\newblock \emph{Generalizability {Theory}}.
\newblock Springer Science \& Business Media.
\newblock Google-Books-ID: nbHbBwAAQBAJ.

\bibitem[{Briggs and Wilson(2007)}]{briggs_generalizability_2007}
Derek~C. Briggs and Mark Wilson. 2007.
\newblock \href {https://doi.org/10.1111/j.1745-3984.2007.00031.x} {Generalizability in item response modeling}.
\newblock \emph{Journal of Educational Measurement}, 44(2):131--155.
\newblock Place: United Kingdom Publisher: Blackwell Publishing.

\bibitem[{Casabianca(2021)}]{casabianca_digital_2021}
Jodi~M. Casabianca. 2021.
\newblock \href {https://doi.org/10.1111/emip.12478} {Digital {Module} 27: {Hierarchical} {Rater} {Models}}.
\newblock \emph{Educational Measurement: Issues and Practice}, 40(4):103--104.
\newblock \_eprint: https://onlinelibrary.wiley.com/doi/pdf/10.1111/emip.12478.

\bibitem[{Casabianca et~al.(2013)Casabianca, McCaffrey, Gitomer, Bell, Hamre, and Pianta}]{casabianca_effect_2013}
Jodi~M. Casabianca, Daniel~F. McCaffrey, Drew~H. Gitomer, Courtney~A. Bell, Bridget~K. Hamre, and Robert~C. Pianta. 2013.
\newblock \href {https://doi.org/10.1177/0013164413486987} {Effect of {Observation} {Mode} on {Measures} of {Secondary} {Mathematics} {Teaching}}.
\newblock \emph{Educational and Psychological Measurement}, 73(5):757--783.
\newblock Publisher: SAGE Publications Inc.

\bibitem[{Charalambous and Delaney(2019)}]{charalambous_13_2019}
Charalambos~Y. Charalambous and Seán Delaney. 2019.
\newblock \href {https://doi.org/10.1163/9789004418875_014} {13 {Mathematics} {Teaching} {Practices} and {Practice}-{Based} {Pedagogies}}.
\newblock Brill.
\newblock Section: International Handbook of Mathematics Teacher Education: Volume 1.

\bibitem[{Charles(2005)}]{charles_correction_2005}
Eric~P. Charles. 2005.
\newblock \href {https://doi.org/10.1037/1082-989X.10.2.206} {The {Correction} for {Attenuation} {Due} to {Measurement} {Error}: {Clarifying} {Concepts} and {Creating} {Confidence} {Sets}}.
\newblock \emph{Psychological Methods}, 10(2):206--226.
\newblock Place: US Publisher: American Psychological Association.

\bibitem[{Corbett-Davies et~al.(2023)Corbett-Davies, Gaebler, Nilforoshan, Shroff, and Goel}]{corbett-davies_measure_2023}
Sam Corbett-Davies, Johann~D. Gaebler, Hamed Nilforoshan, Ravi Shroff, and Sharad Goel. 2023.
\newblock \href {https://doi.org/10.48550/arXiv.1808.00023} {The {Measure} and {Mismeasure} of {Fairness}}.
\newblock \emph{arXiv preprint}.
\newblock ArXiv:1808.00023 [cs].

\bibitem[{Cui et~al.(2024)Cui, Wang, and Xu}]{cui_variational_2024}
Chengyu Cui, Chun Wang, and Gongjun Xu. 2024.
\newblock \href {https://doi.org/10.1007/s11336-024-09955-8} {Variational {Estimation} for {Multidimensional} {Generalized} {Partial} {Credit} {Model}}.
\newblock \emph{Psychometrika}.

\bibitem[{D'Amour et~al.(2020)D'Amour, Heller, Moldovan, Adlam, Alipanahi, Beutel, Chen, Deaton, Eisenstein, Hoffman, Hormozdiari, Houlsby, Hou, Jerfel, Karthikesalingam, Lucic, Ma, McLean, Mincu, Mitani, Montanari, Nado, Natarajan, Nielson, Osborne, Raman, Ramasamy, Sayres, Schrouff, Seneviratne, Sequeira, Suresh, Veitch, Vladymyrov, Wang, Webster, Yadlowsky, Yun, Zhai, and Sculley}]{damour_underspecification_2020}
Alexander D'Amour, Katherine Heller, Dan Moldovan, Ben Adlam, Babak Alipanahi, Alex Beutel, Christina Chen, Jonathan Deaton, Jacob Eisenstein, Matthew~D. Hoffman, Farhad Hormozdiari, Neil Houlsby, Shaobo Hou, Ghassen Jerfel, Alan Karthikesalingam, Mario Lucic, Yian Ma, Cory McLean, Diana Mincu, Akinori Mitani, Andrea Montanari, Zachary Nado, Vivek Natarajan, Christopher Nielson, Thomas~F. Osborne, Rajiv Raman, Kim Ramasamy, Rory Sayres, Jessica Schrouff, Martin Seneviratne, Shannon Sequeira, Harini Suresh, Victor Veitch, Max Vladymyrov, Xuezhi Wang, Kellie Webster, Steve Yadlowsky, Taedong Yun, Xiaohua Zhai, and D.~Sculley. 2020.
\newblock \href {https://doi.org/10.48550/arXiv.2011.03395} {Underspecification {Presents} {Challenges} for {Credibility} in {Modern} {Machine} {Learning}}.
\newblock \emph{arXiv preprint}.
\newblock ArXiv:2011.03395 [cs, stat].

\bibitem[{Darling-Hammond(2014)}]{darling-hammond_what_2014}
Linda Darling-Hammond. 2014.
\newblock \href {https://scholarworks.umb.edu/nejpp/vol26/iss1/4} {What {Can} {PISA} {Tell} {Us} about {U}.{S}. {Education} {Policy}?}
\newblock \emph{New England Journal of Public Policy}, 26(1).

\bibitem[{Darling-Hammond et~al.(2020)Darling-Hammond, Flook, Cook-Harvey, Barron, and Osher}]{darling-hammond_implications_2020}
Linda Darling-Hammond, Lisa Flook, Channa Cook-Harvey, Brigid Barron, and David Osher. 2020.
\newblock \href {https://doi.org/10.1080/10888691.2018.1537791} {Implications for educational practice of the science of learning and development}.
\newblock \emph{Applied Developmental Science}, 24(2):97--140.
\newblock Publisher: Routledge \_eprint: https://doi.org/10.1080/10888691.2018.1537791.

\bibitem[{Decarlo(2003)}]{decarlo_using_2003}
Lawrence~T. Decarlo. 2003.
\newblock \href {https://doi.org/10.3758/BF03195496} {Using the {PLUM} procedure of {SPSS} to fit unequal variance and generalized signal detection models}.
\newblock \emph{Behavior Research Methods, Instruments, \& Computers}, 35(1):49--56.

\bibitem[{DeCarlo(2008)}]{decarlo_studies_2008}
Lawrence~T. DeCarlo. 2008.
\newblock \href {https://doi.org/10.1002/j.2333-8504.2008.tb02149.x} {Studies of a {Latent}-{Class} {Signal}-{Detection} {Model} for {Constructed}-{Response} {Scoring}}.
\newblock \emph{ETS Research Report Series}, 2008(2):i--55.
\newblock \_eprint: https://onlinelibrary.wiley.com/doi/pdf/10.1002/j.2333-8504.2008.tb02149.x.

\bibitem[{DeCarlo(2023)}]{decarlo_classical_2023}
Lawrence~T. DeCarlo. 2023.
\newblock \href {https://doi.org/10.1111/jedm.12358} {Classical {Item} {Analysis} from a {Signal} {Detection} {Perspective}}.
\newblock \emph{Journal of Educational Measurement}, 60(3):520--547.
\newblock \_eprint: https://onlinelibrary.wiley.com/doi/pdf/10.1111/jedm.12358.

\bibitem[{DeCarlo et~al.(2011)DeCarlo, Kim, and Johnson}]{decarlo_hierarchical_2011}
Lawrence~T. DeCarlo, YoungKoung Kim, and Matthew~S. Johnson. 2011.
\newblock \href {https://www.jstor.org/stable/23018150} {A {Hierarchical} {Rater} {Model} for {Constructed} {Responses}, with a {Signal} {Detection} {Rater} {Model}}.
\newblock \emph{Journal of Educational Measurement}, 48(3):333--356.
\newblock Publisher: National Council on Measurement in Education.

\bibitem[{Demszky and Hill(2022)}]{demszky_ncte_2022}
Dorottya Demszky and Heather Hill. 2022.
\newblock \href {https://doi.org/10.48550/ARXIV.2211.11772} {The {NCTE} {Transcripts}: {A} {Dataset} of {Elementary} {Math} {Classroom} {Transcripts}}.
\newblock Publisher: arXiv Version Number: 1.

\bibitem[{Demszky and Hill(2023)}]{demszky_ncte_2023}
Dorottya Demszky and Heather Hill. 2023.
\newblock \href {https://doi.org/10.18653/v1/2023.bea-1.44} {The {NCTE} {Transcripts}: {A} {Dataset} of {Elementary} {Math} {Classroom} {Transcripts}}.
\newblock In \emph{Proceedings of the 18th {Workshop} on {Innovative} {Use} of {NLP} for {Building} {Educational} {Applications} ({BEA} 2023)}, pages 528--538, Toronto, Canada. Association for Computational Linguistics.

\bibitem[{Demszky and Liu(2023)}]{demszky_m-powering_2023}
Dorottya Demszky and Jing Liu. 2023.
\newblock \href {https://doi.org/10.1145/3573051.3593379} {M-{Powering} {Teachers}: {Natural} {Language} {Processing} {Powered} {Feedback} {Improves} 1:1 {Instruction} and {Student} {Outcomes}}.
\newblock In \emph{Proceedings of the {Tenth} {ACM} {Conference} on {Learning} @ {Scale}}, L@{S} '23, pages 59--69, New York, NY, USA. Association for Computing Machinery.
\newblock Event-place: Copenhagen, Denmark.

\bibitem[{Demszky et~al.(2023)Demszky, Liu, Hill, Sanghi, and Chung}]{demszky_improving_2023}
Dorottya Demszky, Jing Liu, Heather~C. Hill, Shyamoli Sanghi, and Ariel Chung. 2023.
\newblock \href {https://edworkingpapers.com/ai23-875} {Improving {Teachers}’ {Questioning} {Quality} through {Automated} {Feedback}: {A} {Mixed}-{Methods} {Randomized} {Controlled} {Trial} in {Brick}-and-{Mortar} {Classrooms}}.
\newblock Technical report, Annenberg Institute at Brown University.
\newblock Publication Title: EdWorkingPapers.com.

\bibitem[{Demszky et~al.(2021)Demszky, Liu, Mancenido, Cohen, Hill, Jurafsky, and Hashimoto}]{demszky_measuring_2021}
Dorottya Demszky, Jing Liu, Zid Mancenido, Julie Cohen, Heather Hill, Dan Jurafsky, and Tatsunori Hashimoto. 2021.
\newblock \href {https://doi.org/10.48550/ARXIV.2106.03873} {Measuring {Conversational} {Uptake}: {A} {Case} {Study} on {Student}-{Teacher} {Interactions}}.
\newblock Publisher: arXiv Version Number: 1.

\bibitem[{Demszky et~al.(2024)Demszky, Wang, Geraghty, and Yu}]{demszky_does_2024}
Dorottya Demszky, Rose Wang, Sean Geraghty, and Carol Yu. 2024.
\newblock \href {https://doi.org/10.1145/3636555.3636924} {Does {Feedback} on {Talk} {Time} {Increase} {Student} {Engagement}? {Evidence} from a {Randomized} {Controlled} {Trial} on a {Math} {Tutoring} {Platform}}.
\newblock In \emph{Proceedings of the 14th {Learning} {Analytics} and {Knowledge} {Conference}}, {LAK} '24, pages 632--644, New York, NY, USA. Association for Computing Machinery.

\bibitem[{Devlin et~al.(2019)Devlin, Chang, Lee, and Toutanova}]{devlin_bert_2019}
Jacob Devlin, Ming-Wei Chang, Kenton Lee, and Kristina Toutanova. 2019.
\newblock \href {https://doi.org/10.18653/v1/N19-1423} {{BERT}: {Pre}-training of {Deep} {Bidirectional} {Transformers} for {Language} {Understanding}}.
\newblock In \emph{Proceedings of the 2019 {Conference} of the {North} {American} {Chapter} of the {Association} for {Computational} {Linguistics}: {Human} {Language} {Technologies}, {Volume} 1 ({Long} and {Short} {Papers})}, pages 4171--4186, Minneapolis, Minnesota. Association for Computational Linguistics.

\bibitem[{Ding et~al.(2022)Ding, Hardt, Miller, and Schmidt}]{ding_retiring_2022}
Frances Ding, Moritz Hardt, John Miller, and Ludwig Schmidt. 2022.
\newblock \href {https://doi.org/10.48550/arXiv.2108.04884} {Retiring {Adult}: {New} {Datasets} for {Fair} {Machine} {Learning}}.
\newblock \emph{arXiv preprint}.
\newblock ArXiv:2108.04884 [cs, stat].

\bibitem[{Donnelly et~al.(2017)Donnelly, Blanchard, Olney, Kelly, Nystrand, and D'Mello}]{donnelly_words_2017}
Patrick~J. Donnelly, Nathaniel Blanchard, Andrew~M. Olney, Sean Kelly, Martin Nystrand, and Sidney~K. D'Mello. 2017.
\newblock \href {https://doi.org/10.1145/3027385.3027417} {Words matter: automatic detection of teacher questions in live classroom discourse using linguistics, acoustics, and context}.
\newblock In \emph{Proceedings of the {Seventh} {International} {Learning} {Analytics} \& {Knowledge} {Conference}}, {LAK} '17, pages 218--227, New York, NY, USA. Association for Computing Machinery.

\bibitem[{Dwork et~al.(2012)Dwork, Hardt, Pitassi, Reingold, and Zemel}]{dwork_fairness_2012}
Cynthia Dwork, Moritz Hardt, Toniann Pitassi, Omer Reingold, and Richard Zemel. 2012.
\newblock \href {https://doi.org/10.1145/2090236.2090255} {Fairness through awareness}.
\newblock In \emph{Proceedings of the 3rd {Innovations} in {Theoretical} {Computer} {Science} {Conference}}, {ITCS} '12, pages 214--226, New York, NY, USA. Association for Computing Machinery.

\bibitem[{Eckes and Jin()}]{eckes_detecting_nodate}
Thomas Eckes and Kuan-Yu Jin.
\newblock Detecting {Illusory} {Halo} {Effects} in {Rater}- {Mediated} {Assessment}: {A} {Mixture} {Rasch} {Facets} {Modeling} {Approach}.

\bibitem[{Field et~al.(2021)Field, Blodgett, Waseem, and Tsvetkov}]{field_survey_2021}
Anjalie Field, Su~Lin Blodgett, Zeerak Waseem, and Yulia Tsvetkov. 2021.
\newblock \href {https://doi.org/10.18653/v1/2021.acl-long.149} {A {Survey} of {Race}, {Racism}, and {Anti}-{Racism} in {NLP}}.
\newblock In \emph{Proceedings of the 59th {Annual} {Meeting} of the {Association} for {Computational} {Linguistics} and the 11th {International} {Joint} {Conference} on {Natural} {Language} {Processing} ({Volume} 1: {Long} {Papers})}, pages 1905--1925, Online. Association for Computational Linguistics.

\bibitem[{Fleisig et~al.(2024)Fleisig, Smith, Bossi, Rustagi, Yin, and Klein}]{fleisig_linguistic_2024}
Eve Fleisig, Genevieve Smith, Madeline Bossi, Ishita Rustagi, Xavier Yin, and Dan Klein. 2024.
\newblock \href {https://doi.org/10.48550/arXiv.2406.08818} {Linguistic {Bias} in {ChatGPT}: {Language} {Models} {Reinforce} {Dialect} {Discrimination}}.
\newblock \emph{arXiv preprint}.
\newblock ArXiv:2406.08818 [cs] version: 1.

\bibitem[{Gao(2022)}]{gao-2022-systeme}
Shuai Gao. 2022.
\newblock \href {https://aclanthology.org/2022.jeptalnrecital-recital.8} {Syst{\`e}me de traduction automatique neuronale fran{\c{c}}ais-mongol (historique, mise en place et {\'e}valuations) ({F}rench-{M}ongolian neural machine translation system (history, implementation, and evaluations) machine translation (hereafter abbreviated {MT}) is currently undergoing rapid development, during which less-resourced languages nevertheless seem to be less developed)}.
\newblock In \emph{Actes de la 29e Conf{\'e}rence sur le Traitement Automatique des Langues Naturelles. Volume 2 : 24e Rencontres Etudiants Chercheurs en Informatique pour le TAL (RECITAL)}, pages 97--110, Avignon, France. ATALA.

\bibitem[{Glaese et~al.(2022)Glaese, McAleese, Trębacz, Aslanides, Firoiu, Ewalds, Rauh, Weidinger, Chadwick, Thacker, Campbell-Gillingham, Uesato, Huang, Comanescu, Yang, See, Dathathri, Greig, Chen, Fritz, Elias, Green, Mokrá, Fernando, Wu, Foley, Young, Gabriel, Isaac, Mellor, Hassabis, Kavukcuoglu, Hendricks, and Irving}]{glaese_improving_2022}
Amelia Glaese, Nat McAleese, Maja Trębacz, John Aslanides, Vlad Firoiu, Timo Ewalds, Maribeth Rauh, Laura Weidinger, Martin Chadwick, Phoebe Thacker, Lucy Campbell-Gillingham, Jonathan Uesato, Po-Sen Huang, Ramona Comanescu, Fan Yang, Abigail See, Sumanth Dathathri, Rory Greig, Charlie Chen, Doug Fritz, Jaume~Sanchez Elias, Richard Green, Soňa Mokrá, Nicholas Fernando, Boxi Wu, Rachel Foley, Susannah Young, Iason Gabriel, William Isaac, John Mellor, Demis Hassabis, Koray Kavukcuoglu, Lisa~Anne Hendricks, and Geoffrey Irving. 2022.
\newblock \href {https://doi.org/10.48550/arXiv.2209.14375} {Improving alignment of dialogue agents via targeted human judgements}.
\newblock \emph{arXiv preprint}.
\newblock ArXiv:2209.14375.

\bibitem[{Gordon et~al.(2022)Gordon, Lam, Park, Patel, Hancock, Hashimoto, and Bernstein}]{gordon_jury_2022}
Mitchell~L. Gordon, Michelle~S. Lam, Joon~Sung Park, Kayur Patel, Jeff Hancock, Tatsunori Hashimoto, and Michael~S. Bernstein. 2022.
\newblock \href {https://doi.org/10.1145/3491102.3502004} {Jury {Learning}: {Integrating} {Dissenting} {Voices} into {Machine} {Learning} {Models}}.
\newblock In \emph{Proceedings of the 2022 {CHI} {Conference} on {Human} {Factors} in {Computing} {Systems}}, {CHI} '22, pages 1--19, New York, NY, USA. Association for Computing Machinery.

\bibitem[{Grissom et~al.(2013)Grissom, Loeb, and Master}]{grissom_effective_2013}
Jason Grissom, Susanna Loeb, and Benjamin Master. 2013.
\newblock \href {https://cepa.stanford.edu/content/effective-instructional-time-use-school-leaders-longitudinal-evidence-observations-principals} {Effective {Instructional} {Time} {Use} for {School} {Leaders}: {Longitudinal} {Evidence} from {Observations} of {Principals}}.
\newblock \emph{Educational Researcher}, 42(8)(42(8)):433.

\bibitem[{Guttman(1945)}]{guttman_basis_1945}
Louis Guttman. 1945.
\newblock \href {https://doi.org/10.1007/BF02288892} {A basis for analyzing test-retest reliability}.
\newblock \emph{Psychometrika}, 10(4):255--282.

\bibitem[{Hammond(2015)}]{hammond_culturally_2015}
Zaretta Hammond. 2015.
\newblock \emph{Culturally responsive teaching and the brain: promoting authentic engagement and rigor among culturally and linguistically diverse students}.
\newblock Corwin, a SAGE company, Thousand Oaks, California.
\newblock OCLC: ocn889185083.

\bibitem[{Hardt et~al.(2016)Hardt, Price, and Srebro}]{hardt_equality_2016}
Moritz Hardt, Eric Price, and Nathan Srebro. 2016.
\newblock \href {https://doi.org/10.48550/arXiv.1610.02413} {Equality of {Opportunity} in {Supervised} {Learning}}.
\newblock \emph{arXiv preprint}.
\newblock ArXiv:1610.02413 [cs].

\bibitem[{Hardy(2021)}]{hardy_toward_2021}
Mike Hardy. 2021.
\newblock \href {https://doi.org/10.48550/arXiv.2112.11973} {Toward {Educator}-focused {Automated} {Scoring} {Systems} for {Reading} and {Writing}}.
\newblock \emph{arXiv preprint}.
\newblock ArXiv:2112.11973 [cs].

\bibitem[{Hebert-Johnson et~al.(2018)Hebert-Johnson, Kim, Reingold, and Rothblum}]{hebert-johnson_multicalibration_2018}
Ursula Hebert-Johnson, Michael Kim, Omer Reingold, and Guy Rothblum. 2018.
\newblock \href {https://proceedings.mlr.press/v80/hebert-johnson18a.html} {Multicalibration: {Calibration} for the ({Computationally}-{Identifiable}) {Masses}}.
\newblock In \emph{Proceedings of the 35th {International} {Conference} on {Machine} {Learning}}, pages 1939--1948. PMLR.
\newblock ISSN: 2640-3498.

\bibitem[{Henderson et~al.(2024)Henderson, Qi, Zeng, Xie, Chen, Jia, and Mittal}]{henderson_safety_2024}
Peter Henderson, Xiangyu Qi, Yi~Zeng, Tinghao Xie, Pin-Yu Chen, Ruoxi Jia, and Prateek Mittal. 2024.
\newblock Safety {Risks} from {Customizing} {Foundation} {Models} via {Fine}-tuning.

\bibitem[{Heo et~al.(2024)Heo, Heinze-Deml, Elachqar, Ren, Nallasamy, Miller, Chan, and Narain}]{heo_llms_2024}
Juyeon Heo, Christina Heinze-Deml, Oussama Elachqar, Shirley Ren, Udhay Nallasamy, Andy Miller, Kwan Ho~Ryan Chan, and Jaya Narain. 2024.
\newblock \href {https://arxiv.org/abs/2410.14516v1} {Do {LLMs} "know" internally when they follow instructions?}

\bibitem[{Hill et~al.(2008)Hill, Blunk, Charalambous, Lewis, Phelps, Sleep, and Ball}]{hill_mathematical_2008}
Heather~C. Hill, Merrie~L. Blunk, Charalambos~Y. Charalambous, Jennifer~M. Lewis, Geoffrey~C. Phelps, Laurie Sleep, and Deborah~Loewenberg Ball. 2008.
\newblock \href {https://www.jstor.org/stable/27739893} {Mathematical {Knowledge} for {Teaching} and the {Mathematical} {Quality} of {Instruction}: {An} {Exploratory} {Study}}.
\newblock \emph{Cognition and Instruction}, 26(4):430--511.
\newblock Publisher: Taylor \& Francis, Ltd.

\bibitem[{Hill et~al.(2012{\natexlab{a}})Hill, Charalambous, Blazar, McGinn, Kraft, Beisiegel, Humez, Litke, and Lynch}]{hill_validating_2012}
Heather~C. Hill, Charalambos~Y. Charalambous, David Blazar, Daniel McGinn, Matthew~A. Kraft, Mary Beisiegel, Andrea Humez, Erica Litke, and Kathleen Lynch. 2012{\natexlab{a}}.
\newblock \href {https://doi.org/10.1080/10627197.2012.715019} {Validating {Arguments} for {Observational} {Instruments}: {Attending} to {Multiple} {Sources} of {Variation}}.
\newblock \emph{Educational Assessment}, 17(2-3):88--106.
\newblock Publisher: Routledge \_eprint: https://doi.org/10.1080/10627197.2012.715019.

\bibitem[{Hill et~al.(2012{\natexlab{b}})Hill, Charalambous, and Kraft}]{hill_when_2012}
Heather~C. Hill, Charalambos~Y. Charalambous, and Matthew~A. Kraft. 2012{\natexlab{b}}.
\newblock \href {https://doi.org/10.3102/0013189X12437203} {When {Rater} {Reliability} {Is} {Not} {Enough}: {Teacher} {Observation} {Systems} and a {Case} for the {Generalizability} {Study}}.
\newblock \emph{Educational Researcher}, 41(2):56--64.
\newblock Publisher: American Educational Research Association.

\bibitem[{Ho and Kane(2013)}]{ho_reliability_2013}
Andrew~D. Ho and Thomas~J. Kane. 2013.
\newblock \href {https://eric.ed.gov/?id=ED540957} {The {Reliability} of {Classroom} {Observations} by {School} {Personnel}. {Research} {Paper}. {MET} {Project}}.
\newblock Technical report, Bill \& Melinda Gates Foundation.
\newblock Publication Title: Bill \& Melinda Gates Foundation ERIC Number: ED540957.

\bibitem[{Hofmann et~al.(2024{\natexlab{a}})Hofmann, Kalluri, Jurafsky, and King}]{hofmann_ai_2024}
Valentin Hofmann, Pratyusha~Ria Kalluri, Dan Jurafsky, and Sharese King. 2024{\natexlab{a}}.
\newblock \href {https://doi.org/10.1038/s41586-024-07856-5} {{AI} generates covertly racist decisions about people based on their dialect}.
\newblock \emph{Nature}, pages 1--8.
\newblock Publisher: Nature Publishing Group.

\bibitem[{Hofmann et~al.(2024{\natexlab{b}})Hofmann, Kalluri, Jurafsky, and King}]{hofmann_dialect_2024}
Valentin Hofmann, Pratyusha~Ria Kalluri, Dan Jurafsky, and Sharese King. 2024{\natexlab{b}}.
\newblock \href {https://doi.org/10.48550/arXiv.2403.00742} {Dialect prejudice predicts {AI} decisions about people's character, employability, and criminality}.
\newblock \emph{arXiv preprint}.
\newblock ArXiv:2403.00742 [cs].

\bibitem[{Hosking et~al.(2024)Hosking, Blunsom, and Bartolo}]{hosking_human_2024}
Tom Hosking, Phil Blunsom, and Max Bartolo. 2024.
\newblock \href {https://doi.org/10.48550/arXiv.2309.16349} {Human {Feedback} is not {Gold} {Standard}}.
\newblock \emph{arXiv preprint}.
\newblock ArXiv:2309.16349.

\bibitem[{Hosseiny~Marani et~al.(2022)Hosseiny~Marani, Levine, and Baumer}]{hosseiny_marani_one_2022}
Amin Hosseiny~Marani, Joshua Levine, and Eric~P.S. Baumer. 2022.
\newblock \href {https://doi.org/10.1145/3511808.3557410} {One {Rating} to {Rule} {Them} {All}? {Evidence} of {Multidimensionality} in {Human} {Assessment} of {Topic} {Labeling} {Quality}}.
\newblock In \emph{Proceedings of the 31st {ACM} {International} {Conference} on {Information} \& {Knowledge} {Management}}, {CIKM} '22, pages 768--779, New York, NY, USA. Association for Computing Machinery.

\bibitem[{Huang and Bolt(2023)}]{huang_unipolar_2023}
Qi~(Helen) Huang and Daniel~M. Bolt. 2023.
\newblock \href {https://doi.org/10.3758/s13428-023-02275-2} {Unipolar {IRT} and the {Author} {Recognition} {Test} ({ART})}.
\newblock \emph{Behavior Research Methods}.

\bibitem[{Jacobs et~al.(2022)Jacobs, Hubbard, and Federmeier}]{jacobs-etal-2022-masked}
Cassandra~L. Jacobs, Ryan~J. Hubbard, and Kara~D. Federmeier. 2022.
\newblock \href {https://aclanthology.org/2022.scil-1.22} {Masked language models directly encode linguistic uncertainty}.
\newblock In \emph{Proceedings of the Society for Computation in Linguistics 2022}, pages 225--228, online. Association for Computational Linguistics.

\bibitem[{Ji(2023)}]{ji_using_2023}
Xuejun~(Ryan) Ji. 2023.
\newblock \href {https://doi.org/10.14288/1.0437518} {\emph{Using cross-classified mixed effects model for validation studies : a flexible and pragmatic validation method}}.
\newblock Ph.D. thesis, University of British Columbia.

\bibitem[{Jurenka et~al.(2024)Jurenka, Kunesch, McKee, Gillick, Zhu, Phal, Hermann, Kasenberg, Bhoopchand, Anand, Pîslar, Chan, Wang, She, Mahmoudieh, Ko, Huber, Wiltshire, Elidan, Rabin, Rubinovitz, McAllister, Wilkowski, Choi, Engelberg, Hackmon, Levin, Griffin, Sears, Bar, Mesar, Jabbour, Chaudhry, Cohan, Thiagarajan, Levine, Brown, Gorur, Grant, Hashimshoni, Hu, Chen, Dolecki, Akbulut, Bileschi, Culp, Dong, Marchal, Deman, Misra, Duah, Ambar, Caciularu, Lefdal, Summerfield, An, Kamienny, Mohdi, Strinopoulous, Hale, Anderson, Cobo, Efron, Ananda, Mohamed, Heymans, Ghahramani, Matias, Gomes, and Ibrahim}]{jurenka_towards_2024}
Irina Jurenka, Markus Kunesch, Kevin~R McKee, Daniel Gillick, Shaojian Zhu, Shubham~Milind Phal, Katherine Hermann, Daniel Kasenberg, Avishkar Bhoopchand, Ankit Anand, Miruna Pîslar, Stephanie Chan, Lisa Wang, Jennifer She, Parsa Mahmoudieh, Wei-Jen Ko, Andrea Huber, Brett Wiltshire, Gal Elidan, Roni Rabin, Jasmin Rubinovitz, Mac McAllister, Julia Wilkowski, David Choi, Roee Engelberg, Lidan Hackmon, Adva Levin, Rachel Griffin, Michael Sears, Filip Bar, Mia Mesar, Mana Jabbour, Arslan Chaudhry, James Cohan, Sridhar Thiagarajan, Nir Levine, Ben Brown, Dilan Gorur, Svetlana Grant, Rachel Hashimshoni, Jieru Hu, Dawn Chen, Kuba Dolecki, Canfer Akbulut, Maxwell Bileschi, Laura Culp, Wen-Xin Dong, Nahema Marchal, Kelsie~Van Deman, Hema~Bajaj Misra, Michael Duah, Moran Ambar, Avi Caciularu, Sandra Lefdal, Chris Summerfield, James An, Pierre-Alexandre Kamienny, Abhinit Mohdi, Theofilos Strinopoulous, Annie Hale, Wayne Anderson, Luis~C Cobo, Niv Efron, Muktha Ananda, Shakir Mohamed, Maureen Heymans, Zoubin
  Ghahramani, Yossi Matias, Ben Gomes, and Lila Ibrahim. 2024.
\newblock Towards {Responsible} {Development} of {Generative} {AI} for {Education}: {An} {Evaluation}-{Driven} {Approach}.

\bibitem[{Kane et~al.(2015)Kane, Hill, and Staiger}]{kane_national_2015}
Thomas Kane, Heather Hill, and Douglas Staiger. 2015.
\newblock \href {https://doi.org/10.3886/ICPSR36095.V4} {National {Center} for {Teacher} {Effectiveness} {Main} {Study}: {Version} 4}.

\bibitem[{Kane et~al.(2013)Kane, McCaffrey, Miller, and Staiger}]{kane_have_2013}
Thomas~J. Kane, Daniel~F. McCaffrey, Trey Miller, and Douglas~O. Staiger. 2013.
\newblock \href {https://eric.ed.gov/?id=ED540959} {Have {We} {Identified} {Effective} {Teachers}? {Validating} {Measures} of {Effective} {Teaching} {Using} {Random} {Assignment}. {Research} {Paper}. {MET} {Project}}.
\newblock Technical report, Bill \& Melinda Gates Foundation.
\newblock Publication Title: Bill \& Melinda Gates Foundation ERIC Number: ED540959.

\bibitem[{Kane and Staiger(2012)}]{kane_gathering_2012}
Thomas~J. Kane and Douglas~O. Staiger. 2012.
\newblock \href {https://eric.ed.gov/?id=ED540960} {Gathering {Feedback} for {Teaching}: {Combining} {High}-{Quality} {Observations} with {Student} {Surveys} and {Achievement} {Gains}. {Research} {Paper}. {MET} {Project}}.
\newblock Technical report, Bill \& Melinda Gates Foundation.
\newblock Publication Title: Bill \& Melinda Gates Foundation ERIC Number: ED540960.

\bibitem[{Kasy and Abebe(2021)}]{kasy_fairness_2021}
Maximilian Kasy and Rediet Abebe. 2021.
\newblock \href {https://doi.org/10.1145/3442188.3445919} {Fairness, {Equality}, and {Power} in {Algorithmic} {Decision}-{Making}}.
\newblock In \emph{Proceedings of the 2021 {ACM} {Conference} on {Fairness}, {Accountability}, and {Transparency}}, pages 576--586, Virtual Event Canada. ACM.

\bibitem[{Kazai et~al.(2013)Kazai, Kamps, and Milic-Frayling}]{kazai_analysis_2013}
Gabriella Kazai, Jaap Kamps, and Natasa Milic-Frayling. 2013.
\newblock \href {https://doi.org/10.1007/s10791-012-9205-0} {An analysis of human factors and label accuracy in crowdsourcing relevance judgments}.
\newblock \emph{Information Retrieval}, 16(2):138--178.

\bibitem[{Kelly et~al.(2018)Kelly, Olney, Donnelly, Nystrand, and D’Mello}]{kelly_automatically_2018}
Sean Kelly, Andrew~M. Olney, Patrick Donnelly, Martin Nystrand, and Sidney~K. D’Mello. 2018.
\newblock \href {https://doi.org/10.3102/0013189X18785613} {Automatically {Measuring} {Question} {Authenticity} in {Real}-{World} {Classrooms}}.
\newblock \emph{Educational Researcher}, 47(7):451--464.
\newblock Publisher: American Educational Research Association.

\bibitem[{Kiela et~al.(2021)Kiela, Bartolo, Nie, Kaushik, Geiger, Wu, Vidgen, Prasad, Singh, Ringshia, Ma, Thrush, Riedel, Waseem, Stenetorp, Jia, Bansal, Potts, and Williams}]{kiela_dynabench_2021}
Douwe Kiela, Max Bartolo, Yixin Nie, Divyansh Kaushik, Atticus Geiger, Zhengxuan Wu, Bertie Vidgen, Grusha Prasad, Amanpreet Singh, Pratik Ringshia, Zhiyi Ma, Tristan Thrush, Sebastian Riedel, Zeerak Waseem, Pontus Stenetorp, Robin Jia, Mohit Bansal, Christopher Potts, and Adina Williams. 2021.
\newblock \href {https://doi.org/10.48550/arXiv.2104.14337} {Dynabench: {Rethinking} {Benchmarking} in {NLP}}.
\newblock \emph{arXiv preprint}.
\newblock ArXiv:2104.14337 [cs].

\bibitem[{Kim et~al.(2018)Kim, Wattenberg, Gilmer, Cai, Wexler, Viegas, and Sayres}]{kim_interpretability_2018}
Been Kim, Martin Wattenberg, Justin Gilmer, Carrie Cai, James Wexler, Fernanda Viegas, and Rory Sayres. 2018.
\newblock \href {https://doi.org/10.48550/arXiv.1711.11279} {Interpretability {Beyond} {Feature} {Attribution}: {Quantitative} {Testing} with {Concept} {Activation} {Vectors} ({TCAV})}.
\newblock \emph{arXiv preprint}.
\newblock ArXiv:1711.11279 [stat].

\bibitem[{Kingma and Ba(2017)}]{kingma_adam_2017}
Diederik~P. Kingma and Jimmy Ba. 2017.
\newblock \href {https://doi.org/10.48550/arXiv.1412.6980} {Adam: {A} {Method} for {Stochastic} {Optimization}}.
\newblock \emph{arXiv preprint}.
\newblock ArXiv:1412.6980 [cs].

\bibitem[{Klahr(2013)}]{klahr_what_2013}
David Klahr. 2013.
\newblock \href {https://doi.org/10.1073/pnas.1212738110} {What do we mean? {On} the importance of not abandoning scientific rigor when talking about science education}.
\newblock \emph{Proceedings of the National Academy of Sciences}, 110(supplement\_3):14075--14080.
\newblock Publisher: Proceedings of the National Academy of Sciences.

\bibitem[{Kromrey et~al.(2008)Kromrey, Fay, and Bellara}]{kromrey_macro_2008}
J.~Kromrey, Robert~H. Fay, and Aarti~P. Bellara. 2008.
\newblock \href {https://www.semanticscholar.org/paper/Macro-for-Computing-Confidence-Intervals-for-Kromrey-Fay/62c7827b2d2ebb01dd9cd78a757001513f335141} {Macro for {Computing} {Confidence} {Intervals} for {Disattenuated} {Correlation} {Coefficients}}.

\bibitem[{Lemov(2021)}]{lemov_teach_2021}
Doug Lemov. 2021.
\newblock \emph{Teach like a champion 3.0: 63 techniques that put students on the path to college}, third edition edition.
\newblock Jossey-Bass, a Wiley imprint, Hoboken, NJ.

\bibitem[{Lemov and Atkins(2015)}]{lemov_teach_2015}
Doug Lemov and Norman Atkins. 2015.
\newblock \emph{Teach like a champion 2.0: 62 techniques that put students on the path to college}, second edition edition.
\newblock Jossey-Bass, San Francisco, CA.

\bibitem[{Li et~al.(2023)Li, Zhang, Zhang, Long, Xie, and Zhang}]{li_towards_2023}
Zehan Li, Xin Zhang, Yanzhao Zhang, Dingkun Long, Pengjun Xie, and Meishan Zhang. 2023.
\newblock \href {https://doi.org/10.48550/arXiv.2308.03281} {Towards {General} {Text} {Embeddings} with {Multi}-stage {Contrastive} {Learning}}.
\newblock \emph{arXiv preprint}.
\newblock ArXiv:2308.03281 [cs].

\bibitem[{Liljedahl et~al.(2021)Liljedahl, Zager, and Wheeler}]{liljedahl_building_2021}
Peter Liljedahl, Tracy~Johnston Zager, and Laura Wheeler. 2021.
\newblock \emph{Building thinking classrooms in mathematics: 14 teaching practices for enhancing learning: {Grades} {K}-12}.
\newblock Corwin {Mathematics}. Corwin, Thousand Oaks, California London New Delhi Singapore.

\bibitem[{Liu and Cohen(2021)}]{liu_measuring_2021}
Jing Liu and Julie Cohen. 2021.
\newblock \href {https://doi.org/10.3102/01623737211009267} {Measuring {Teaching} {Practices} at {Scale}: {A} {Novel} {Application} of {Text}-as-{Data} {Methods}}.
\newblock \emph{Educational Evaluation and Policy Analysis}, 43(4):587--614.
\newblock Publisher: American Educational Research Association.

\bibitem[{Liu et~al.(2023{\natexlab{a}})Liu, Lin, Hewitt, Paranjape, Bevilacqua, Petroni, and Liang}]{liu_lost_2023}
Nelson~F. Liu, Kevin Lin, John Hewitt, Ashwin Paranjape, Michele Bevilacqua, Fabio Petroni, and Percy Liang. 2023{\natexlab{a}}.
\newblock \href {https://doi.org/10.48550/arXiv.2307.03172} {Lost in the {Middle}: {How} {Language} {Models} {Use} {Long} {Contexts}}.
\newblock \emph{arXiv preprint}.
\newblock ArXiv:2307.03172 [cs].

\bibitem[{Liu et~al.(2023{\natexlab{b}})Liu, Iter, Xu, Wang, Xu, and Zhu}]{liu_g-eval_2023}
Yang Liu, Dan Iter, Yichong Xu, Shuohang Wang, Ruochen Xu, and Chenguang Zhu. 2023{\natexlab{b}}.
\newblock \href {https://doi.org/10.48550/arXiv.2303.16634} {G-{Eval}: {NLG} {Evaluation} using {GPT}-4 with {Better} {Human} {Alignment}}.
\newblock \emph{arXiv preprint}.
\newblock ArXiv:2303.16634 [cs].

\bibitem[{Liu et~al.(2019)Liu, Ott, Goyal, Du, Joshi, Chen, Levy, Lewis, Zettlemoyer, and Stoyanov}]{liu_roberta_2019}
Yinhan Liu, Myle Ott, Naman Goyal, Jingfei Du, Mandar Joshi, Danqi Chen, Omer Levy, Mike Lewis, Luke Zettlemoyer, and Veselin Stoyanov. 2019.
\newblock \href {https://doi.org/10.48550/arXiv.1907.11692} {{RoBERTa}: {A} {Robustly} {Optimized} {BERT} {Pretraining} {Approach}}.
\newblock \emph{arXiv preprint}.
\newblock ArXiv:1907.11692 [cs].

\bibitem[{Lundberg and Lee(2017)}]{lundberg_unified_2017}
Scott Lundberg and Su-In Lee. 2017.
\newblock \href {https://doi.org/10.48550/arXiv.1705.07874} {A {Unified} {Approach} to {Interpreting} {Model} {Predictions}}.
\newblock \emph{arXiv preprint}.
\newblock ArXiv:1705.07874 [cs, stat].

\bibitem[{Mantzicopoulos et~al.(2018)Mantzicopoulos, French, and Patrick}]{mantzicopoulos_mathematical_2018}
Panayota Mantzicopoulos, Brian~F. French, and Helen Patrick. 2018.
\newblock \href {https://doi.org/10.1080/10409289.2018.1477903} {The {Mathematical} {Quality} of {Instruction} ({MQI}) in {Kindergarten}: {An} {Evaluation} of the {Stability} of the {MQI} {Using} {Generalizability} {Theory}}.
\newblock \emph{Early Education and Development}, 29(6):893--908.
\newblock Publisher: Routledge \_eprint: https://doi.org/10.1080/10409289.2018.1477903.

\bibitem[{Mariano and Junker(2007)}]{mariano_covariates_2007}
Louis~T. Mariano and Brian~W. Junker. 2007.
\newblock \href {https://doi.org/10.3102/1076998606298033} {Covariates of the {Rating} {Process} in {Hierarchical} {Models} for {Multiple} {Ratings} of {Test} {Items}}.
\newblock \emph{Journal of Educational and Behavioral Statistics}, 32(3):287--314.

\bibitem[{McCoy et~al.(2023)McCoy, Yao, Friedman, Hardy, and Griffiths}]{mccoy_embers_2023}
R.~Thomas McCoy, Shunyu Yao, Dan Friedman, Matthew Hardy, and Thomas~L. Griffiths. 2023.
\newblock \href {https://arxiv.org/abs/2309.13638v1} {Embers of {Autoregression}: {Understanding} {Large} {Language} {Models} {Through} the {Problem} {They} are {Trained} to {Solve}}.

\bibitem[{Messick(1998)}]{messick_test_1998}
Samuel Messick. 1998.
\newblock \href {https://www.jstor.org/stable/27522333} {Test {Validity}: {A} {Matter} of {Consequence}}.
\newblock \emph{Social Indicators Research}, 45(1/3):35--44.
\newblock Publisher: Springer.

\bibitem[{Muchinsky(1996)}]{muchinsky_correction_1996}
Paul~M. Muchinsky. 1996.
\newblock \href {https://doi.org/10.1177/0013164496056001004} {The {Correction} for {Attenuation}}.
\newblock \emph{Educational and Psychological Measurement}, 56(1):63--75.
\newblock Publisher: SAGE Publications Inc.

\bibitem[{Muraki(1992)}]{muraki_generalized_1992}
Eiji Muraki. 1992.
\newblock \href {https://doi.org/10.1177/014662169201600206} {A {Generalized} {Partial} {Credit} {Model}: {Application} of an {EM} {Algorithm}}.
\newblock \emph{Applied Psychological Measurement}, 16(2):159--176.
\newblock Publisher: SAGE Publications Inc.

\bibitem[{Murphy and Beretvas(2015)}]{murphy_comparison_2015}
Daniel~L. Murphy and S.~Natasha Beretvas. 2015.
\newblock \href {https://doi.org/10.1080/08957347.2015.1042158} {A {Comparison} of {Teacher} {Effectiveness} {Measures} {Calculated} {Using} {Three} {Multilevel} {Models} for {Raters} {Effects}}.
\newblock \emph{Applied Measurement in Education}, 28(3):219--236.
\newblock Publisher: Routledge \_eprint: https://doi.org/10.1080/08957347.2015.1042158.

\bibitem[{Nghiem et~al.(2024)Nghiem, Prindle, Zhao, and III}]{nghiem_you_2024}
Huy Nghiem, John Prindle, Jieyu Zhao, and Hal~Daumé III. 2024.
\newblock \href {https://doi.org/10.48550/arXiv.2406.12232} {"{You} {Gotta} be a {Doctor}, {Lin}": {An} {Investigation} of {Name}-{Based} {Bias} of {Large} {Language} {Models} in {Employment} {Recommendations}}.
\newblock \emph{arXiv preprint}.
\newblock ArXiv:2406.12232.

\bibitem[{Patz et~al.(2002)Patz, Junker, Johnson, and Mariano}]{patz_hierarchical_2002}
Richard~J. Patz, Brian~W. Junker, Matthew~S. Johnson, and Louis~T. Mariano. 2002.
\newblock \href {https://www.jstor.org/stable/3648122} {The {Hierarchical} {Rater} {Model} for {Rated} {Test} {Items} and {Its} {Application} to {Large}-{Scale} {Educational} {Assessment} {Data}}.
\newblock \emph{Journal of Educational and Behavioral Statistics}, 27(4):341--384.
\newblock Publisher: [American Educational Research Association, Sage Publications, Inc., American Statistical Association].

\bibitem[{Pianta and Hamre(2009)}]{pianta_conceptualization_2009}
Robert~C. Pianta and Bridget~K. Hamre. 2009.
\newblock \href {https://doi.org/10.3102/0013189X09332374} {Conceptualization, {Measurement}, and {Improvement} of {Classroom} {Processes}: {Standardized} {Observation} {Can} {Leverage} {Capacity}}.
\newblock \emph{Educational Researcher}, 38(2):109--119.
\newblock Publisher: American Educational Research Association.

\bibitem[{Pianta et~al.(2008)Pianta, Paro, and Hamre}]{pianta_classroom_2008}
Robert~C. Pianta, Karen M.~La Paro, and Bridget~K. Hamre. 2008.
\newblock \emph{Classroom {Assessment} {Scoring} {System} ({CLASS}) {Manual}, {K}-3}.
\newblock Paul H. Brookes Publishing Company.
\newblock Google-Books-ID: NBeaGgAACAAJ.

\bibitem[{Pleiss et~al.(2017)Pleiss, Raghavan, Wu, Kleinberg, and Weinberger}]{pleiss_fairness_2017}
Geoff Pleiss, Manish Raghavan, Felix Wu, Jon Kleinberg, and Kilian~Q. Weinberger. 2017.
\newblock \href {https://doi.org/10.48550/arXiv.1709.02012} {On {Fairness} and {Calibration}}.
\newblock \emph{arXiv preprint}.
\newblock ArXiv:1709.02012 [cs, stat].

\bibitem[{Plummer(2003)}]{plummer_jags_2003}
Martyn Plummer. 2003.
\newblock {JAGS}: {A} program for analysis of {Bayesian} graphical models using {Gibbs} sampling.
\newblock \emph{Working Papers}.

\bibitem[{Qi et~al.(2023)Qi, Zeng, Xie, Chen, Jia, Mittal, and Henderson}]{qi_fine-tuning_2023}
Xiangyu Qi, Yi~Zeng, Tinghao Xie, Pin-Yu Chen, Ruoxi Jia, Prateek Mittal, and Peter Henderson. 2023.
\newblock \href {https://doi.org/10.48550/arXiv.2310.03693} {Fine-tuning {Aligned} {Language} {Models} {Compromises} {Safety}, {Even} {When} {Users} {Do} {Not} {Intend} {To}!}
\newblock \emph{arXiv preprint}.
\newblock ArXiv:2310.03693.

\bibitem[{Ribeiro et~al.(2020)Ribeiro, Wu, Guestrin, and Singh}]{ribeiro_beyond_2020}
Marco~Tulio Ribeiro, Tongshuang Wu, Carlos Guestrin, and Sameer Singh. 2020.
\newblock \href {https://doi.org/10.18653/v1/2020.acl-main.442} {Beyond {Accuracy}: {Behavioral} {Testing} of {NLP} {Models} with {CheckList}}.
\newblock In \emph{Proceedings of the 58th {Annual} {Meeting} of the {Association} for {Computational} {Linguistics}}, pages 4902--4912, Online. Association for Computational Linguistics.

\bibitem[{Rickford and King(2016)}]{rickford_language_2016}
John~R. Rickford and Sharese King. 2016.
\newblock \href {https://doi.org/10.1353/lan.2016.0078} {Language and linguistics on trial: {Hearing} {Rachel} {Jeantel} (and other vernacular speakers) in the courtroom and beyond}.
\newblock \emph{Language}, 92(4):948--988.

\bibitem[{Rudin(2019)}]{rudin_stop_2019}
Cynthia Rudin. 2019.
\newblock \href {https://doi.org/10.48550/arXiv.1811.10154} {Stop {Explaining} {Black} {Box} {Machine} {Learning} {Models} for {High} {Stakes} {Decisions} and {Use} {Interpretable} {Models} {Instead}}.
\newblock \emph{arXiv preprint}.
\newblock ArXiv:1811.10154 [cs, stat].

\bibitem[{Samei et~al.(2014)Samei, Olney, Kelly, Nystrand, D'Mello, Blanchard, Sun, Glaus, and Graesser}]{samei_domain_2014}
Borhan Samei, Andrew~M. Olney, Sean Kelly, Martin Nystrand, Sidney D'Mello, Nathan Blanchard, Xiaoyi Sun, Marcy Glaus, and Art Graesser. 2014.
\newblock \href {https://eric.ed.gov/?id=ED566380} {Domain {Independent} {Assessment} of {Dialogic} {Properties} of {Classroom} {Discourse}}.
\newblock Technical report.
\newblock Publication Title: Grantee Submission ERIC Number: ED566380.

\bibitem[{Sanh et~al.(2020)Sanh, Debut, Chaumond, and Wolf}]{sanh_distilbert_2020}
Victor Sanh, Lysandre Debut, Julien Chaumond, and Thomas Wolf. 2020.
\newblock \href {https://doi.org/10.48550/arXiv.1910.01108} {{DistilBERT}, a distilled version of {BERT}: smaller, faster, cheaper and lighter}.
\newblock \emph{arXiv preprint}.
\newblock ArXiv:1910.01108.

\bibitem[{Saphier et~al.(2008)Saphier, Haley-Speca, and Gower}]{saphier_skillful_2008}
Jon Saphier, Mary~Ann Haley-Speca, and Robert Gower. 2008.
\newblock \emph{The skillful teacher: building your teaching skills}, 6th ed edition.
\newblock Research for Better Teaching, Acton, Mass.

\bibitem[{Schwartz et~al.(2016)Schwartz, Tsang, and Blair}]{schwartz_abcs_2016}
Daniel~L. Schwartz, Jessica~M. Tsang, and Kristen~P. Blair. 2016.
\newblock \emph{The {ABCs} of how we learn: 26 scientifically proven approaches, how they work, and when to use them}, first edition edition.
\newblock Norton books in education. W.W. Norton \& Company, New York.

\bibitem[{Shermis(2014)}]{shermis_state---art_2014}
Mark~D. Shermis. 2014.
\newblock \href {https://doi.org/10.1016/j.asw.2013.04.001} {State-of-the-art automated essay scoring: {Competition}, results, and future directions from a {United} {States} demonstration}.
\newblock \emph{Assessing Writing}, 20:53--76.

\bibitem[{Shieh et~al.(2024)Shieh, Vassel, Sugimoto, and Monroe-White}]{shieh_laissez-faire_2024}
Evan Shieh, Faye-Marie Vassel, Cassidy Sugimoto, and Thema Monroe-White. 2024.
\newblock \href {https://doi.org/10.48550/arXiv.2404.07475} {Laissez-{Faire} {Harms}: {Algorithmic} {Biases} in {Generative} {Language} {Models}}.
\newblock \emph{arXiv preprint}.
\newblock ArXiv:2404.07475.

\bibitem[{Slavin(2002)}]{slavin_evidence-based_2002}
Robert~E. Slavin. 2002.
\newblock \href {https://doi.org/10.3102/0013189X031007015} {Evidence-{Based} {Education} {Policies}: {Transforming} {Educational} {Practice} and {Research}}.
\newblock \emph{Educational Researcher}, 31(7):15--21.
\newblock Publisher: American Educational Research Association.

\bibitem[{Song et~al.(2020)Song, Kalluri, Grover, Zhao, and Ermon}]{song_learning_2020}
Jiaming Song, Pratyusha Kalluri, Aditya Grover, Shengjia Zhao, and Stefano Ermon. 2020.
\newblock \href {https://doi.org/10.48550/arXiv.1812.04218} {Learning {Controllable} {Fair} {Representations}}.
\newblock \emph{arXiv preprint}.
\newblock ArXiv:1812.04218 [cs, stat].

\bibitem[{Sundararajan et~al.(2017)Sundararajan, Taly, and Yan}]{sundararajan_axiomatic_2017}
Mukund Sundararajan, Ankur Taly, and Qiqi Yan. 2017.
\newblock \href {https://proceedings.mlr.press/v70/sundararajan17a.html} {Axiomatic {Attribution} for {Deep} {Networks}}.
\newblock In \emph{Proceedings of the 34th {International} {Conference} on {Machine} {Learning}}, pages 3319--3328. PMLR.
\newblock ISSN: 2640-3498.

\bibitem[{Suresh et~al.(2022)Suresh, Jacobs, Harty, Perkoff, Martin, and Sumner}]{suresh-etal-2022-talkmoves}
Abhijit Suresh, Jennifer Jacobs, Charis Harty, Margaret Perkoff, James~H. Martin, and Tamara Sumner. 2022.
\newblock \href {https://aclanthology.org/2022.lrec-1.497} {The {T}alk{M}oves dataset: K-12 mathematics lesson transcripts annotated for teacher and student discursive moves}.
\newblock In \emph{Proceedings of the Thirteenth Language Resources and Evaluation Conference}, pages 4654--4662, Marseille, France. European Language Resources Association.

\bibitem[{Tack et~al.(2023)Tack, Kochmar, Yuan, Bibauw, and Piech}]{tack_bea_2023}
Anaïs Tack, Ekaterina Kochmar, Zheng Yuan, Serge Bibauw, and Chris Piech. 2023.
\newblock \href {https://doi.org/10.48550/arXiv.2306.06941} {The {BEA} 2023 {Shared} {Task} on {Generating} {AI} {Teacher} {Responses} in {Educational} {Dialogues}}.
\newblock \emph{arXiv preprint}.
\newblock ArXiv:2306.06941.

\bibitem[{Team()}]{r_core_team_r_nodate}
R~Core Team.
\newblock \href {https://www.r-project.org/} {R: {A} {Language} and {Environment} for {Statistical} {Computing}}.

\bibitem[{Touvron et~al.(2023)Touvron, Martin, Stone, Albert, Almahairi, Babaei, Bashlykov, Batra, Bhargava, Bhosale, Bikel, Blecher, Ferrer, Chen, Cucurull, Esiobu, Fernandes, Fu, Fu, Fuller, Gao, Goswami, Goyal, Hartshorn, Hosseini, Hou, Inan, Kardas, Kerkez, Khabsa, Kloumann, Korenev, Koura, Lachaux, Lavril, Lee, Liskovich, Lu, Mao, Martinet, Mihaylov, Mishra, Molybog, Nie, Poulton, Reizenstein, Rungta, Saladi, Schelten, Silva, Smith, Subramanian, Tan, Tang, Taylor, Williams, Kuan, Xu, Yan, Zarov, Zhang, Fan, Kambadur, Narang, Rodriguez, Stojnic, Edunov, and Scialom}]{touvron_llama_2023}
Hugo Touvron, Louis Martin, Kevin Stone, Peter Albert, Amjad Almahairi, Yasmine Babaei, Nikolay Bashlykov, Soumya Batra, Prajjwal Bhargava, Shruti Bhosale, Dan Bikel, Lukas Blecher, Cristian~Canton Ferrer, Moya Chen, Guillem Cucurull, David Esiobu, Jude Fernandes, Jeremy Fu, Wenyin Fu, Brian Fuller, Cynthia Gao, Vedanuj Goswami, Naman Goyal, Anthony Hartshorn, Saghar Hosseini, Rui Hou, Hakan Inan, Marcin Kardas, Viktor Kerkez, Madian Khabsa, Isabel Kloumann, Artem Korenev, Punit~Singh Koura, Marie-Anne Lachaux, Thibaut Lavril, Jenya Lee, Diana Liskovich, Yinghai Lu, Yuning Mao, Xavier Martinet, Todor Mihaylov, Pushkar Mishra, Igor Molybog, Yixin Nie, Andrew Poulton, Jeremy Reizenstein, Rashi Rungta, Kalyan Saladi, Alan Schelten, Ruan Silva, Eric~Michael Smith, Ranjan Subramanian, Xiaoqing~Ellen Tan, Binh Tang, Ross Taylor, Adina Williams, Jian~Xiang Kuan, Puxin Xu, Zheng Yan, Iliyan Zarov, Yuchen Zhang, Angela Fan, Melanie Kambadur, Sharan Narang, Aurelien Rodriguez, Robert Stojnic, Sergey Edunov, and Thomas
  Scialom. 2023.
\newblock \href {https://doi.org/10.48550/arXiv.2307.09288} {Llama 2: {Open} {Foundation} and {Fine}-{Tuned} {Chat} {Models}}.
\newblock \emph{arXiv preprint}.
\newblock ArXiv:2307.09288 [cs].

\bibitem[{UpLevel(2024)}]{uplevel_gen_2024}
UpLevel. 2024.
\newblock \href {https://resources.uplevelteam.com/gen-ai-for-coding} {Gen {AI} for {Coding} {Research} {Report}}.
\newblock Technical report, Uplevel Data Labs.

\bibitem[{Vaccaro et~al.(2024)Vaccaro, Almaatouq, and Malone}]{vaccaro_when_2024}
Michelle Vaccaro, Abdullah Almaatouq, and Thomas Malone. 2024.
\newblock \href {https://doi.org/10.48550/arXiv.2405.06087} {When {Are} {Combinations} of {Humans} and {AI} {Useful}?}
\newblock \emph{arXiv preprint}.
\newblock ArXiv:2405.06087 [cs].

\bibitem[{van~der Lee et~al.(2019)van~der Lee, Gatt, van Miltenburg, Wubben, and Krahmer}]{van_der_lee_best_2019}
Chris van~der Lee, Albert Gatt, Emiel van Miltenburg, Sander Wubben, and Emiel Krahmer. 2019.
\newblock \href {https://doi.org/10.18653/v1/W19-8643} {Best practices for the human evaluation of automatically generated text}.
\newblock In \emph{Proceedings of the 12th {International} {Conference} on {Natural} {Language} {Generation}}, pages 355--368, Tokyo, Japan. Association for Computational Linguistics.

\bibitem[{Wang et~al.(2022)Wang, Zhang, Chen, Kim, and Mao}]{wang-etal-2022-text}
Jiarui Wang, Richong Zhang, Junfan Chen, Jaein Kim, and Yongyi Mao. 2022.
\newblock \href {https://doi.org/10.18653/v1/2022.emnlp-main.521} {Text style transferring via adversarial masking and styled filling}.
\newblock In \emph{Proceedings of the 2022 Conference on Empirical Methods in Natural Language Processing}, pages 7654--7663, Abu Dhabi, United Arab Emirates. Association for Computational Linguistics.

\bibitem[{Wang and Demszky(2023)}]{wang_is_2023}
Rose Wang and Dorottya Demszky. 2023.
\newblock \href {https://doi.org/10.18653/v1/2023.bea-1.53} {Is {ChatGPT} a {Good} {Teacher} {Coach}? {Measuring} {Zero}-{Shot} {Performance} {For} {Scoring} and {Providing} {Actionable} {Insights} on {Classroom} {Instruction}}.
\newblock In \emph{Proceedings of the 18th {Workshop} on {Innovative} {Use} of {NLP} for {Building} {Educational} {Applications} ({BEA} 2023)}, pages 626--667, Toronto, Canada. Association for Computational Linguistics.

\bibitem[{Warr et~al.(2024)Warr, Oster, and Isaac}]{warr_implicit_2024}
Melissa Warr, Nicole~Jakubczyk Oster, and Roger Isaac. 2024.
\newblock \href {https://doi.org/10.1080/15391523.2024.2395295} {Implicit bias in large language models: {Experimental} proof and implications for education}.
\newblock \emph{Journal of Research on Technology in Education}, 0(0):1--24.
\newblock Publisher: Routledge \_eprint: https://doi.org/10.1080/15391523.2024.2395295.

\bibitem[{Waseem(2016)}]{waseem_are_2016}
Zeerak Waseem. 2016.
\newblock \href {https://doi.org/10.18653/v1/W16-5618} {Are {You} a {Racist} or {Am} {I} {Seeing} {Things}? {Annotator} {Influence} on {Hate} {Speech} {Detection} on {Twitter}}.
\newblock In \emph{Proceedings of the {First} {Workshop} on {NLP} and {Computational} {Social} {Science}}, pages 138--142, Austin, Texas. Association for Computational Linguistics.

\bibitem[{Webson et~al.(2023)Webson, Loo, Yu, and Pavlick}]{webson_are_2023}
Albert Webson, Alyssa Loo, Qinan Yu, and Ellie Pavlick. 2023.
\newblock \href {https://doi.org/10.18653/v1/2023.findings-emnlp.514} {Are {Language} {Models} {Worse} than {Humans} at {Following} {Prompts}? {It}'s {Complicated}}.
\newblock In \emph{Findings of the {Association} for {Computational} {Linguistics}: {EMNLP} 2023}, pages 7662--7686, Singapore. Association for Computational Linguistics.

\bibitem[{Webson and Pavlick(2022)}]{webson_prompt-based_2022}
Albert Webson and Ellie Pavlick. 2022.
\newblock \href {https://doi.org/10.18653/v1/2022.naacl-main.167} {Do {Prompt}-{Based} {Models} {Really} {Understand} the {Meaning} of {Their} {Prompts}?}
\newblock In \emph{Proceedings of the 2022 {Conference} of the {North} {American} {Chapter} of the {Association} for {Computational} {Linguistics}: {Human} {Language} {Technologies}}, pages 2300--2344, Seattle, United States. Association for Computational Linguistics.

\bibitem[{Whitehill and LoCasale-Crouch(2024)}]{whitehill_automated_2024}
Jacob Whitehill and Jennifer LoCasale-Crouch. 2024.
\newblock \href {https://doi.org/10.48550/arXiv.2310.01132} {Automated {Evaluation} of {Classroom} {Instructional} {Support} with {LLMs} and {BoWs}: {Connecting} {Global} {Predictions} to {Specific} {Feedback}}.
\newblock \emph{arXiv preprint}.
\newblock ArXiv:2310.01132 [cs].

\bibitem[{Whitehurst et~al.(2014)Whitehurst, Chingos, and Lindquist}]{whitehurst_evaluating_2014}
Grover~J. Whitehurst, Matthew~M. Chingos, and Katharine~M. Lindquist. 2014.
\newblock Evaluating {Teachers} with {Classroom} {Observations}: {Lessons} {Learned} in {Four} {Districts}.
\newblock Technical report, Brookings Institution.
\newblock Publication Title: Brookings Institution ERIC Number: ED553815.

\bibitem[{Wind(2019)}]{wind_nonparametric_2019}
Stefanie~A. Wind. 2019.
\newblock \href {https://doi.org/10.1111/jedm.12222} {Nonparametric {Evidence} of {Validity}, {Reliability}, and {Fairness} for {Rater}-{Mediated} {Assessments}: {An} {Illustration} {Using} {Mokken} {Scale} {Analysis}}.
\newblock \emph{Journal of Educational Measurement}, 56(3):478--504.
\newblock \_eprint: https://onlinelibrary.wiley.com/doi/pdf/10.1111/jedm.12222.

\bibitem[{Wind and Guo(2019)}]{wind_exploring_2019}
Stefanie~A. Wind and Wenjing Guo. 2019.
\newblock \href {https://doi.org/10.1177/0013164419834613} {Exploring the {Combined} {Effects} of {Rater} {Misfit} and {Differential} {Rater} {Functioning} in {Performance} {Assessments}}.
\newblock \emph{Educational and Psychological Measurement}, 79(5):962--987.
\newblock Publisher: SAGE Publications Inc.

\bibitem[{Xu et~al.(2024)Xu, Liu, Jones, Cohen, and Ai}]{xu_promises_2024}
Paiheng Xu, Jing Liu, Nathan Jones, Julie Cohen, and Wei Ai. 2024.
\newblock \href {https://doi.org/10.48550/arXiv.2404.02444} {The {Promises} and {Pitfalls} of {Using} {Language} {Models} to {Measure} {Instruction} {Quality} in {Education}}.
\newblock \emph{arXiv preprint}.
\newblock ArXiv:2404.02444.

\bibitem[{Yang et~al.(2020)Yang, Dai, Yang, Carbonell, Salakhutdinov, and Le}]{yang_xlnet_2020}
Zhilin Yang, Zihang Dai, Yiming Yang, Jaime Carbonell, Ruslan Salakhutdinov, and Quoc~V. Le. 2020.
\newblock \href {https://doi.org/10.48550/arXiv.1906.08237} {{XLNet}: {Generalized} {Autoregressive} {Pretraining} for {Language} {Understanding}}.
\newblock \emph{arXiv preprint}.
\newblock ArXiv:1906.08237.

\bibitem[{Zemel et~al.(2013)Zemel, Wu, Swersky, Pitassi, and Dwork}]{zemel_learning_2013}
Rich Zemel, Yu~Wu, Kevin Swersky, Toni Pitassi, and Cynthia Dwork. 2013.
\newblock \href {https://proceedings.mlr.press/v28/zemel13.html} {Learning {Fair} {Representations}}.
\newblock In \emph{Proceedings of the 30th {International} {Conference} on {Machine} {Learning}}, pages 325--333. PMLR.
\newblock ISSN: 1938-7228.

\bibitem[{Zhao and Ermon(2021)}]{zhao_right_2021}
Shengjia Zhao and Stefano Ermon. 2021.
\newblock \href {https://doi.org/10.48550/arXiv.2011.07476} {Right {Decisions} from {Wrong} {Predictions}: {A} {Mechanism} {Design} {Alternative} to {Individual} {Calibration}}.
\newblock \emph{arXiv preprint}.
\newblock ArXiv:2011.07476 [cs, math, stat].

\bibitem[{Zhou et~al.(2024)Zhou, Hwang, Ren, and Sap}]{zhou_relying_2024}
Kaitlyn Zhou, Jena~D. Hwang, Xiang Ren, and Maarten Sap. 2024.
\newblock \href {https://doi.org/10.48550/arXiv.2401.06730} {Relying on the {Unreliable}: {The} {Impact} of {Language} {Models}' {Reluctance} to {Express} {Uncertainty}}.
\newblock \emph{arXiv preprint}.
\newblock ArXiv:2401.06730 [cs].

\bibitem[{Zhou et~al.(2023)Zhou, Kok, Quintana, Delahay, and Wang}]{zhou_how_2023}
Xiaofei Zhou, Christopher Kok, Rebecca~M. Quintana, Anita Delahay, and Xu~Wang. 2023.
\newblock \href {https://doi.org/10.1145/3573051.3593388} {How {Learning} {Experience} {Designers} {Make} {Design} {Decisions}: {The} {Role} of {Data}, the {Reliance} on {Subject} {Matter} {Expertise}, and the {Opportunities} for {Data}-{Driven} {Support}}.
\newblock In \emph{Proceedings of the {Tenth} {ACM} {Conference} on {Learning} @ {Scale}}, L@{S} '23, pages 132--143, New York, NY, USA. Association for Computing Machinery.
\newblock Event-place: Copenhagen, Denmark.

\bibitem[{Zijlmans et~al.(2018{\natexlab{a}})Zijlmans, Tijmstra, van~der Ark, and Sijtsma}]{zijlmans_item-score_2018}
Eva A.~O. Zijlmans, Jesper Tijmstra, L.~Andries van~der Ark, and Klaas Sijtsma. 2018{\natexlab{a}}.
\newblock \href {https://doi.org/10.1177/0013164417728358} {Item-{Score} {Reliability} in {Empirical}-{Data} {Sets} and {Its} {Relationship} {With} {Other} {Item} {Indices}}.
\newblock \emph{Educational and Psychological Measurement}, 78(6):998--1020.
\newblock Publisher: SAGE Publications Inc.

\bibitem[{Zijlmans et~al.(2018{\natexlab{b}})Zijlmans, van~der Ark, Tijmstra, and Sijtsma}]{zijlmans_methods_2018}
Eva A.~O. Zijlmans, L.~Andries van~der Ark, Jesper Tijmstra, and Klaas Sijtsma. 2018{\natexlab{b}}.
\newblock \href {https://doi.org/10.1177/0146621618758290} {Methods for {Estimating} {Item}-{Score} {Reliability}}.
\newblock \emph{Applied Psychological Measurement}, 42(7):553--570.
\newblock Publisher: SAGE Publications Inc.

\end{thebibliography}

\appendix
\section{NCTE Population Descriptive Statistics}\label{apdx:popdesciptives}
\begin{table}[h]
\centering
\begin{tabular}{l c}
\toprule
 & NCTE sample means\\
\hline
Female & 0.85 \\
African-American & 0.22 \\
Asian & 0.03 \\
Hispanic & 0.03 \\
White & 0.65 \\
Teaching Experience (Years) & 10.59 \\
\hline
Teachers & N=309 \\
\hline
Female & 0.50 \\
African-American & 0.41 \\
Asian & 0.08 \\
Hispanic & 0.24 \\
White & 0.24 \\
Free or Reduced Price Lunch& 0.65 \\
Special Education& 0.11 \\
English Language Learners& 0.21 \\
Prior Year State Math Test (Standardized) & 0.08 \\
Prior Year State ELA Test (Standardized) & 0.07 \\
\hline
Students & N=9,141 \\
\bottomrule
\end{tabular}
    \caption{Teacher and student descriptive statistics.}
    \label{tab:populations}
\end{table}

\section{Observation Instrument Item Descriptions and Distributions}\label{apdx:item-descriptions}
\begin{landscape}
    \begin{figure}[h]
        \centering
        \includegraphics[width=0.8\paperheight]{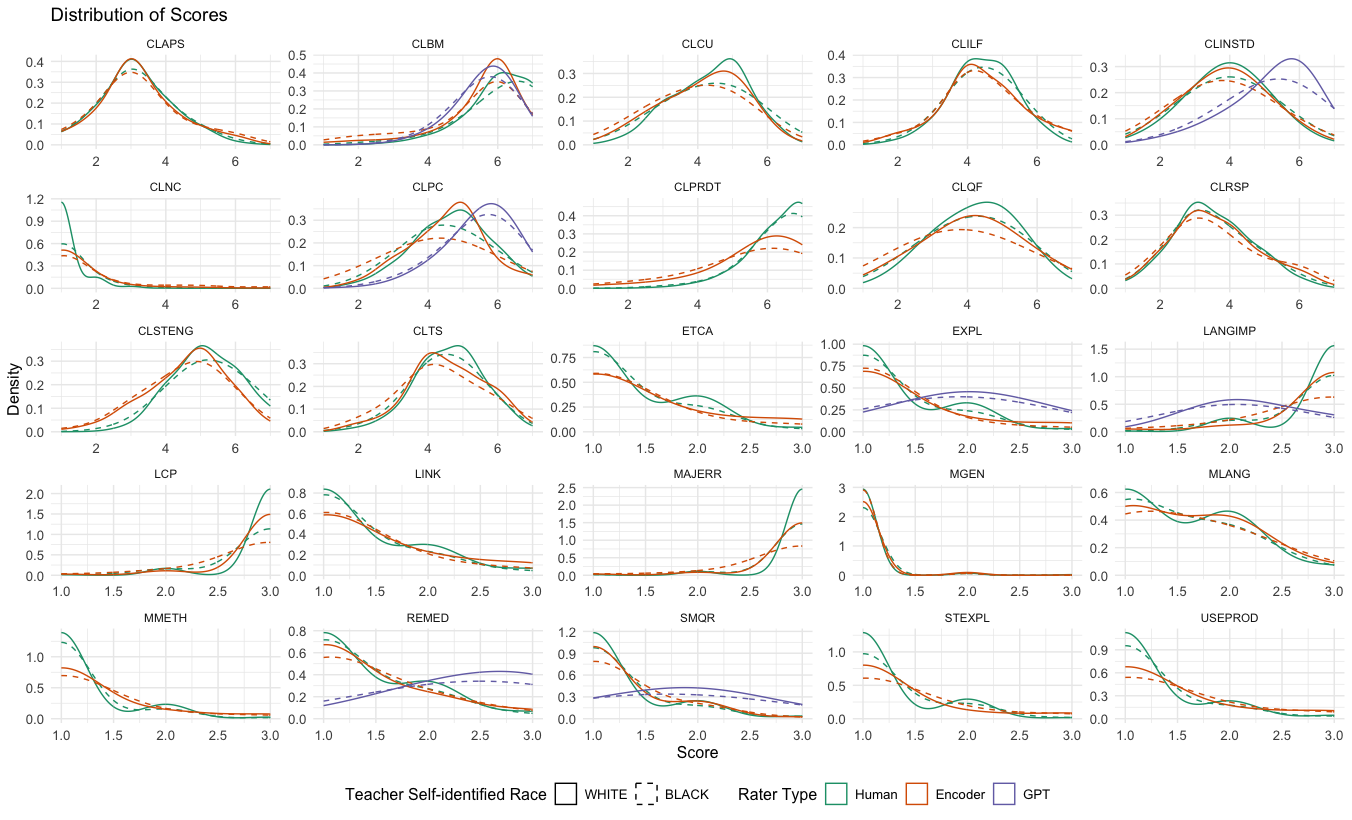}
        \caption{Distribution of rater scores for each of the 25 instrument items for all rater families.}
        \label{fig:rating-dist}
    \end{figure}
\end{landscape}

For each of the observation instruments, the abbreviation codes used in this study are listed with the expanded names in Table \ref{table:items}. The distributions of scores across all items for all rater families are in Figure \ref{fig:rating-dist}. The CLASS rubric has 12 items on a scale from 1 to 7, rated at 15 minute intervals. The MQI rubric has 13 items on a scale from 1 to 3, rated at 7.5 minute intervals.

\begin{figure}[h]
    \centering
    \includegraphics[width=0.75\linewidth]{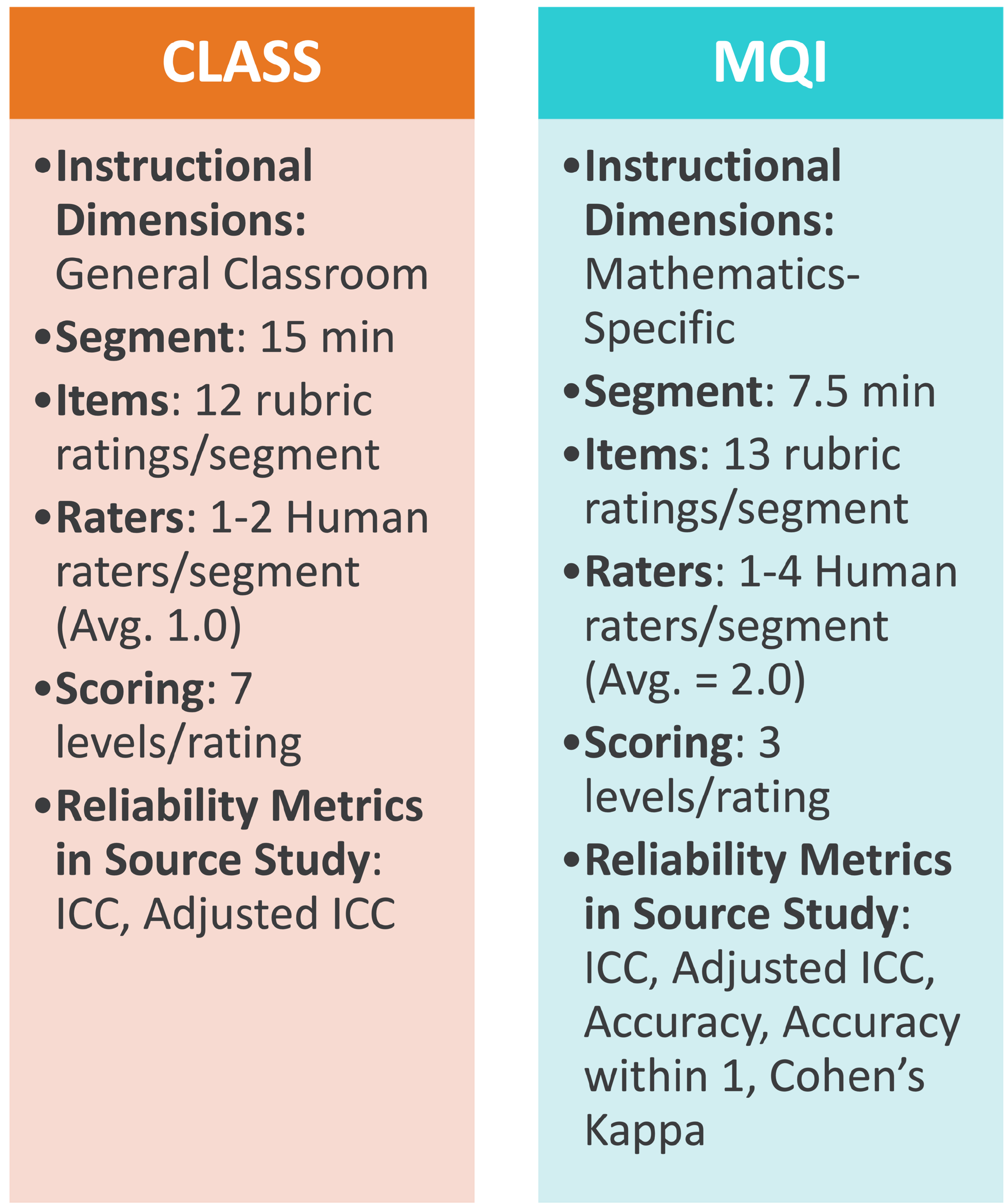}
    \caption{Overview of technical details the two instructional frameworks used for evaluating instruction.}
    \label{fig:obs-instruments}
\end{figure}

\begin{table*}[h]
    \centering
    \setlength\tabcolsep{2pt}\renewcommand\defaultaddspace{1.5ex}
    \begin{tabularx}{1\textwidth}{@{} l>{\hsize=0.65\hsize}X>{\hsize=1.30\hsize}X>{\hsize=1.05\hsize\arraybackslash} X @{}}
    \toprule
    \textbf{Abbreviation} & \textbf{Item} & \textbf{Item Description} \\
    \hline
    \hline
    \underline{\textbf{MQI Instrument}} & \\
    ETCA & \textit{Enacted Task Cognitive Activation} & Task cognitive demand, such as drawing connections among different representations, concepts, or solution methods; identifying and explaining patterns. \\
    \textbf{EXPL} &\textit{Teacher Explanations} & Teacher explanations that give meaning to ideas, procedures, steps, or solution methods. \\
    \textbf{LANGIMP}\dag & \textit{Imprecision in Language or Notation} & Imprecision in language or notation, with regard to mathematical symbols and technical or general mathematical language.  \\
    LCP\dag & \textit{Lack of Clarity in Presentation of Mathematical Content} & Lack of clarity in teachers’ launching of tasks or presentation of the content. \\
    LINK & \textit{Linking and Connections} & Linking and connections of mathematical representations, ideas, and procedures. \\
     MAJERR\dag & \textit{Major Mathematical Errors} & Major mathematical errors, such as solving problems incorrectly, defining terms incorrectly, forgetting a key condition in a definition, equating two non-identical mathematical terms. \\
    MGEN & \textit{Developing Mathematical Generalizations} & Developing generalizations based on multiple examples. \\
    MLANG & \textit{Mathematical Language} & Mathematical language is dense and precise and is used fluently and consistently. \\
    MMETH &\textit{Multiple Procedures or Solution Methods} & Multiple procedures or solution methods for a single problem.  \\
    \textbf{REMED} & \textit{Remediation of Student Errors and Difficulties} & Remediation of student errors and difficulties addressed in a substantive manner. \\
    \textbf{SMQR} & \textit{Student Mathematical Questioning and Reasoning} & Student mathematical questioning and reasoning, such as posing mathematically motivated questions, offering mathematical claims or counterclaims. \\
    STEXPL & \textit{Students Provide Explanations} & Student explanations that give meaning to ideas, procedures, steps, or solution methods. \\
    USEPROD & \textit{Responding to Student Mathematical Productions} & Responding to student mathematical productions in instruction, such as appropriately identifying mathematical insight in specific student questions, comments, or work; building instruction on student ideas or methods. \\
    \midrule
    \underline{\textbf{CLASS Instrument}}  & \\
    \textbf{CLPC} & \textit{Classroom Positive Climate} &  \\
    CLNC\dag & \textit{Classroom Negative Climate} &   \\
    CLTS & \textit{Teacher Sensitivity} &  \\
    CLRSP & \textit{Regard for Student Perspective} &   \\
    \textbf{CLBM} & \textit{Behavior Management} &   \\
    CLPRDT & \textit{Productivity} &   \\
    CLILF & \textit{Instructional Learning Formats} &   \\
    CLCU & \textit{Content Understanding} &   \\
    CLAPS & \textit{Applied Problem Solving} &   \\
    CLQF & \textit{Quality of Feedback} &    \\
    \textbf{CLINSTD} & \textit{Instructional Dialogue} &   \\
    CLSTENG & \textit{Student Engagement} &   \\
    \bottomrule
    \end{tabularx}
    \caption{CLASS and MQI item descriptions and corresponding abbreviations. \dag denotes items that are reverse coded due to being negatively worded with respect to the construct of teacher ability. Bolded items are those evaluated by the \textbf{GPT} family of raters and reported by \citeauthor{wang_is_2023}. Each member of the Human and Encoder families of raters evaluated all 25 items.}
    \label{table:items}
\end{table*}

\section{MQI Instrument}\label{apdx:mqi}
\subsection{MQI Instrument Properties}
For our purposes, the MQI instrument has a few unique properties that warrant further analysis, as the instrument may have some qualitative attributes that may influence human raters. 

The MQI ratings are written to identify the presence of a behavior and then, if present, report the magnitude or quality of its presence, doing so repeatedly at regular intervals throughout the lesson (in this case, 7.5 minutes). This shortened window with simpler targets provides an opportunity for training a model for real-time use (rather than an arbitrary interval) to find different features across a single lesson, as shown in Figure \ref{fig:heartbeats}. 

The version of the MQI for which data is available in the NCTE dataset is ternary, in contrast to the current MQI version, which is quaternary. The lowest rating on the ternary MQI scale is a combination of the lowest two ratings on the quaternary, meaning the present data cannot distinguish between whether the attribute described in each item is “Not present” or “Low”. \footnote{There is one exception, which the original authors of the Appendix adjusted for: the \textbf{\cmtt{USEPROD}} item is replaced by the \textbf{\cmtt{MATCON}} item, with the correction of combining the lowest two categories.} This ternary classification scheme creates non-normal distributions as seen in Figure \ref{fig:rating-dist}, which will need to inform models and methods during quantitative analysis. 

This is unfortunate because the difference between these two categories are “None.” And “Brief {content error, instance of imprecision, lack of clarity}. Does not obscure the mathematics of the segment,” respectively (for the Errors and Imprecision domain in \citeauthor{hill_mathematical_2008} and second MQI-only factor in \citeauthor{blazar_attending_2017}: {\textbf{\cmtt{MAJERR}}, \textbf{\cmtt{LANGIMP}}, \textbf{\cmtt{LCP}}}). 

\subsection{Possible Effects of Negative-worded Items}\label{apx:neg-mqi}
The MQI is unique in having a separate domain of items that try to capture aspects of \textit{poor} mathematical instruction. Unlike most items in observation rubrics, the MQI has three items that are worded in the negative direction, specifically, higher scores on the \textbf{\cmtt{MAJERR}}, \textbf{\cmtt{LANGIMP}}, and \textbf{\cmtt{LCP}} items indicate worse performance.\footnote{In the analyses of this paper, these will be reverse coded, as will the one negative CLASS item \textbf{CLNC}} It is possible that looking for negative attributes may make these items more susceptible to different rater biases. A partial description of the potential impact of this rubric attribute for the \cmtt{LCP} item found in Appendix \ref{apx:neg-mqi} with further details.

Of note, the \cmtt{LCP} item is particularly subjective. In the documentation and training provided for the MQI, You have to ask: “What, mathematically, was the teacher trying to say?” This is already problematic, as it is asking for observers to use their judgment to determine what the teacher was “trying to say.” The subjectivity increases further for observers who may not be as familiar with African-American Vernacular English (AAVE).  The subjectivity is further mixes lack of content clarity (lack of clarity explaining math) with lack of directional clarity (unclear instructions for an activity, which is typically associated with items addressing classroom management), as stated in the MQI rubric: 

\begin{displayquote}
\textit{Teacher’s launch of a task/activity lacks clarity (the “launch” is the teacher’s effort to get the mathematical tasks/activities into play). If the launch is problematic, score for the launch plus amount of time students are confused/off-task/engaging in non-productive explorations…[Example:] Garbling a task launch, e.g., by asking initially “How much TV is watched in the US?” when students really must draw a graph to show “How many TVs in US vs. Europe vs. rest of the world?}
\end{displayquote}

Instructing observers to score based on the “amount of time students are confused/off-task/engaging in non-productive explorations”, is more likely to capture problems with classroom management and directional lack of clarity, not mathematical lack of clarity, compounded by the request for raters to guess what the teachers were trying to say and training instructions that let raters "code Lack of Clarity even with correction". This mix of observational cues and overlapping constructs makes this item particularly susceptible to individual rater biases.\footnote{As a note, the skill of providing clear directions, foundational to establishing a well-managed classroom, is also not included the CLASS instrument's "Behavior Management" item, suggesting that neither of these instruments is perfectly designed to address root causes of instructional shortcomings and thus may be inadequate as tools for coaching and developing skills in teachers.}

Indeed, while not reported in this paper explicitly, we identified that one rater in particular rated Black teachers much more harshly on these, especially on \cmtt{LCP}, providing some evidence that some items can be more prone to rater biases, even with research-quality observers and calibration.

\subsection{Prior work on Rater Fairness with MQI}\label{sec:mqifair}
Recent work has begun to look at rater biases, including racial bias, in these data and with the MQI instrument. \citet{ji_using_2023} uses cross-classified mixed effects models for analysis and evaluation, which seeks to answer similar questions through combining G-theory and IRT estimations \cite{briggs_generalizability_2007}. However, the helpfulness of this study is limited by data selection decisions: it eliminates 23\% of MQI items (all of the second MQI factor in \cite{blazar_attending_2017}) without explanation; it only uses 21\% of available classroom observations (from a single year) and by so doing also eliminates 43\% of the study's raters; it then truncates the class lengths to 45 minutes thus removing another 20\% of the remaining data observations, and when evaluating for differences in teacher race, combines all non-white races/ethnicities into a single category, removing meaningful inference from the contrast. These decisions to use only 13\% of available data would lead to a model with better fit, as all of those removals simplify trends in the data, indirectly suggesting that the mixed effects model constructions used are not robust to the complete set of observations \cite{murphy_comparison_2015} and are therefore inadequate for our purposes here. 

\begin{figure*}[ht]
    \centering
        \includegraphics[width=1\textwidth]{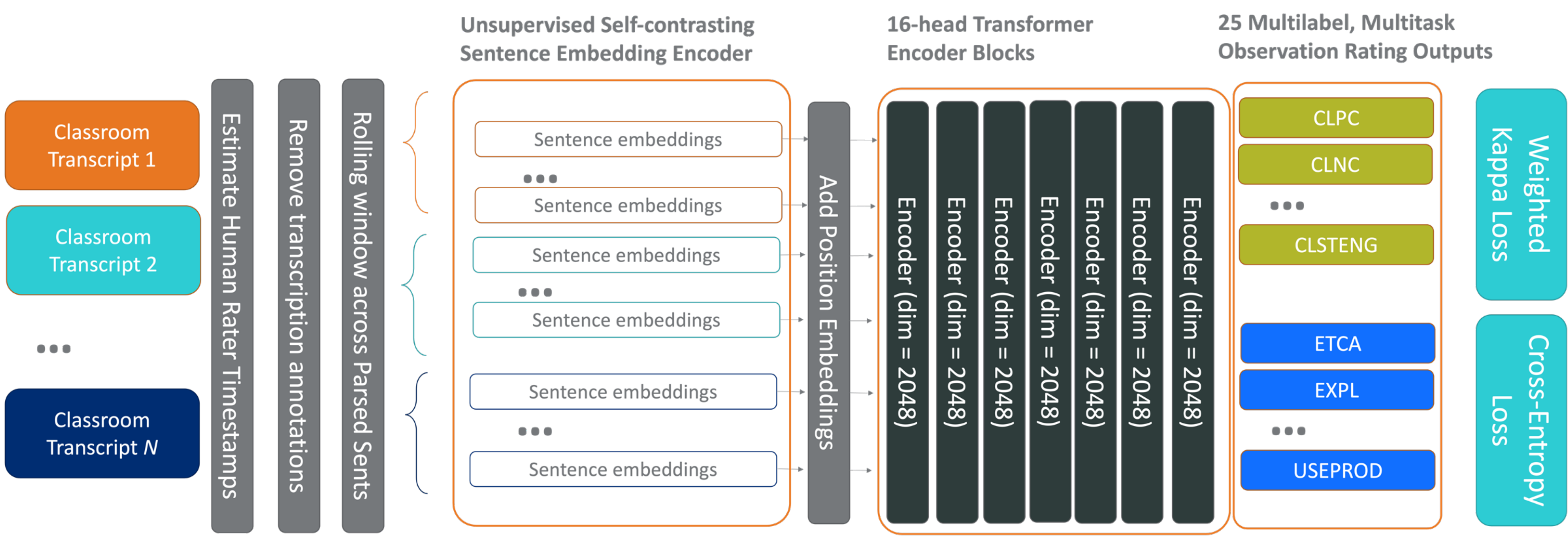}
        \caption{Model Pipeline: General sentence-encoder model architecture.}
        \label{fig:model-diagram}
\end{figure*}

\section{Encoder Family Construction}\label{apx:encoders}
Pretraining and training/fine-tuning regimes can have significant effects on model performance \cite{damour_underspecification_2020}, so our family of models sought to exploit this by using three different pretrainings for sentence-level embeddings and including variations on training regimes (e.g., different checkpoints), the summary of these variations can be found in Table \ref{tab:encoder}. Thus, the encoder family of models designed for this study share the same architecture,\footnote{One model, "un2", has a slightly different architecture, differing in the number of attention heads.} training and held-out test sets, differing only as outlined in Table \ref{tab:encoder}. 

[Another forthcoming paper to be under review] explores this protocol in greater depth, showing that the extreme training and treatment of data noise can achieve SOTA and "super-human" results on a variety of sentence embedding pretrainings, with a more complete set of training

\begin{table*}[ht]
    \centering
    \begin{tabular}{lcccc}
        \toprule
        \textbf{Model} & \textbf{Pretrained Embedding} & \textbf{Layer Attn. Heads} & \textbf{Train Epochs} & \textbf{Dropout} \\
        \midrule
        \textbf{un1} & Unsupervised SimCSE \citep{gao-2022-systeme} & 32 & 3 & 75 \\ 
        \textbf{un2} & Unsupervised SimCSE \citep{gao-2022-systeme} & 16 & 4 & 75 \\ 
        \textbf{un3} & Unsupervised SimCSE \citep{gao-2022-systeme} & 32 & 8 & 75 \\ 
        \textbf{e5}  & E5 \citep{wang-etal-2022-text} & 32 & 2 & 15 \\ 
        \textbf{gte} & GTE \citep{li_towards_2023} & 32 & 4 & 65 \\ 
        \bottomrule
    \end{tabular}
    \caption{Encoder Within-family differences: Summary of basic differences within the Encoder family of models. Detailed information about training and architecture can be found in Appendix \ref{apdx:training}.}
    \label{tab:encoder} 
\end{table*}

All results were run on a completely held out test set (Figure \ref{fig:test_set}) of entire classroom transcripts. No analyses were conducted using the held-out test set until after all models in the model family were trained, thus preserving the integrity of the study. 
\begin{table*}[ht]
    \centering
    \begin{tabularx}{\textwidth}{ccXl}
        \toprule
        \textbf{GPT Model} & \textbf{Name} & \textbf{Prompt Info} & \textbf{Output} \\
        \midrule
        \textbf{N}  & Numeric & Item Overview & Single Number \\ 
        \textbf{ND} & Numeric w/ Description & Rubric Descriptions of Score Categories & Single Number \\ 
        \textbf{NR} & Numeric after Reasoning & Item Overview and CoT instructions & Reasoning and Number \\ 
        \bottomrule
    \end{tabularx}
    \caption{GPT Within-family model differences: Details for the GPT/Decoder models can be found in the original paper \citep{wang_is_2023}.}
    \label{tab:gptdiff}
\end{table*}

\subsubsection{Encoder Model Preprocessing}
\begin{landscape}

\begin{figure}[ht]
        \centering
        \includegraphics[width=0.7\paperheight]{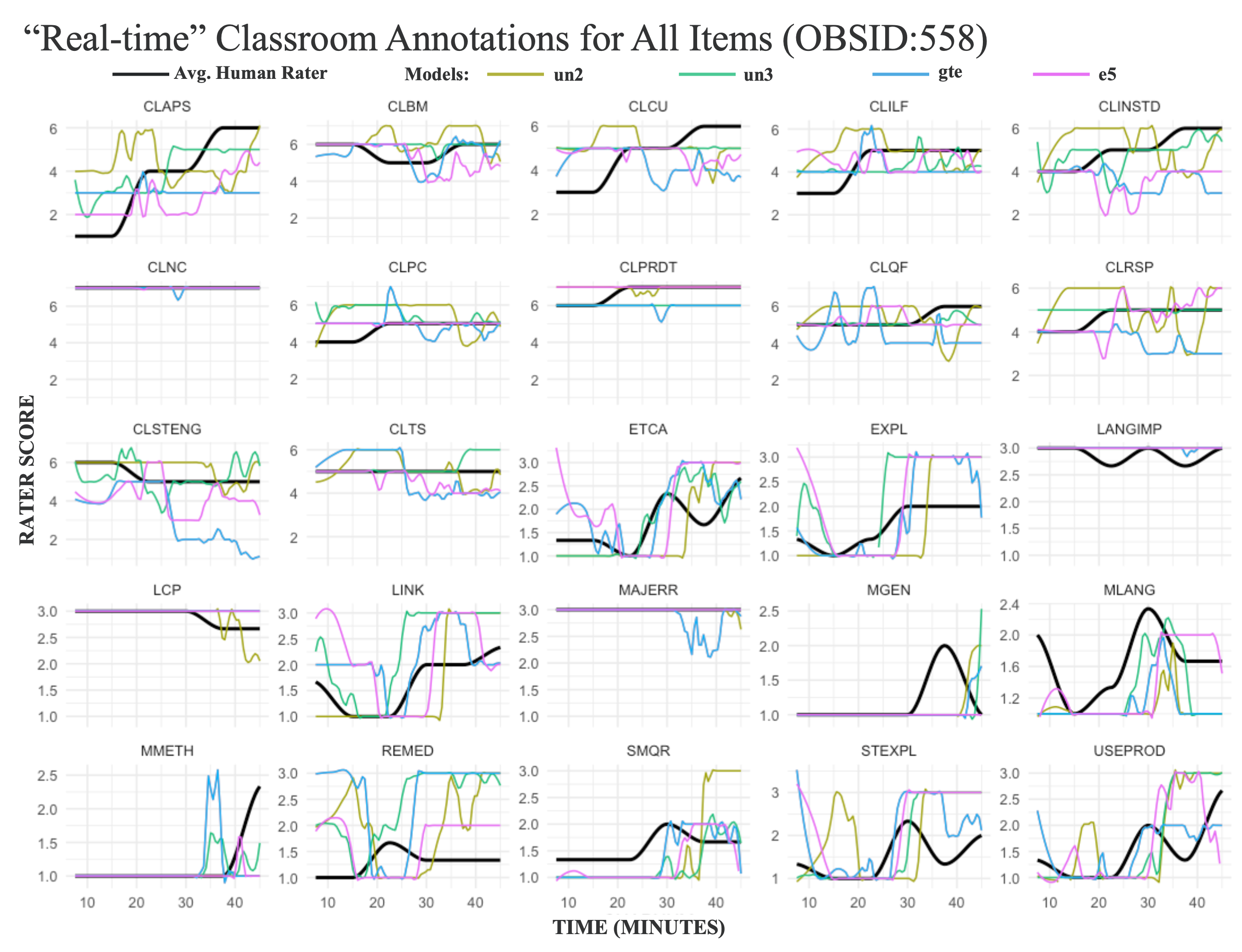}
        \caption{Real-time Evaluation: the X axis represents time in class (where 0 minutes is the start of class), each chart is one of the 25 items in the rubrics, the black lines are human evaluations (averaged, if multiple raters). The other lines are continuous model predictions for that item, using Loess smoothing where local fitting uses tricubic weighting of neighborhood points that span $\alpha = 0.1$.}
        \label{fig:heartbeats}
    \end{figure}

\end{landscape}

As mentioned in Section \ref{sec:modelfams}, preprocessing of the transcript data was intentionally minimal, replacing bracketed transcription notes (e.g. \texttt{[cross-talk]}) with \texttt{[inaudible]}. For this study, the transcript was not annotated denote whether a teacher or a student is speaking to reflect the broadest future use case of general classroom microphones. In other words, this family of models does not know who is speaking, and the results of this decision are evident in the models' relative underperformance in two MQI items that distinguish between teacher explanations (\cmtt{EXPL}) and student explanations (\cmtt{STEXPL}), a trend that might be evident in the validity demonstration in Figure \ref{fig:heartbeats}, where models may be responding nearly identically to/failing to distinguish between these two items.

To align transcripted class segments to human observation ratings, transcripts were equipartitioned at the word-level across the maximum number of lesson segments for which there were human annotations available, and estimated timestamps were made across sentences by linear interpolation weighted by word count.

\subsection{Sentence-level Embeddings }\label{apdx:sentence_embeds}
One key difference to other studies using these same transcripts is the choice to parse the utterances at the sentence level. Sentences, rather than individual words or long, uninterrupted utterances, are the key unit of meaning for interpretability of models for classroom discourse. The downstream tasks are a key decision for this choice. Sentence level parsing anticipates meaningful feature attribution studies \citep{sundararajan_axiomatic_2017} to further investigate construct validity. 

Parsing at the sentence level both augments the total number of unique observations in the data and, by creating more standardization in sequence lengths prior to sentence-embedding, the variation in the density of semantic information is reduced.

The model takes as input an approximate 12 min rolling window of class text (stepping at each sentence), and simultaneously predicts ratings for each of the 12 CLASS dimensions, 13 of the MQI dimensions for rounded-rolling average scores for that time window. Each model is multi-task predicting all 25 scores simultaneously for each of the MQI and CLASS items. This multi-task training takes advantage of the interrelated skills of teaching that may be implicit in human ratings. Over one million unique observations from fewer than 1,600 unique classroom transcripts were generated, with rolling windows representing each observation. Training-val-test splits of this data were 75/15/10, stratified at the classroom level. 

Classroom transcripts are extremely long, with thousands of sentences, and with classes having tokens in the hundreds of thousands. Sentence-level inputs could capture the relationship between something a teacher says and something a student says five minutes later without incurring large costs associated with sequence length. These long-range dependencies are needed to identify some of the instructional constructs being measured.

Raw class transcripts also have a lot of noise: content that is unrelated to any of the tasks, including fillers, self-corrections, interruptions and self-interruptions, sentences that are partially repeated or emphasized, text that requires being able to refer to a visual cue in the classroom, etc. While sentence level embeddings lose information relative to subword tokenizations, this loss of information may mitigate disproportionate effects of idiosyncratic speaking styles.

\subsubsection{Embedding Model Selection}
To save on compute, static embeddings were pre-computed. To represent the very noisy transcript data, we have to be careful in using sentence-embeddings, as they decrease the completeness of the information captured. We tested sentence-level embeddings using across different pretrained embedding models accessed through Huggingface on a subset of the training data for a small random selection of target measures:

\begin{itemize}

    \item \texttt{unsup-simcse-roberta-large}: from princeton-nlp \citep{gao-2022-systeme}, was pretrained using unsupervised contrastive sentence representations. \href{https://github.com/princeton-nlp/SimCSE?tab=readme-ov-file}{simCSE}
    \item \texttt{sup-simcse-roberta-large}: from princeton-nlp \citep{gao-2022-systeme}, was pretrained using supervised training. At the writing of this paper, we did not yet have a converged model with reportable results. \href{https://github.com/princeton-nlp/SimCSE?tab=readme-ov-file}{simCSE}
    \item \texttt{e5-large-v2}: from intfloat \cite{wang-etal-2022-text}, pretrained using weakly supervised contrastive sentence representations with sentence pair training. \href{https://huggingface.co/intfloat/e5-large-v2}{e5-large-v2}
    \item \texttt{gte-large}: from thenlper \cite{li_towards_2023}, pretrained using multistage contrastive sentence representations. \href{https://huggingface.co/thenlper/gte-large}{gte-large}

\end{itemize}

The first three models had significantly reduced performance, compared to our sentence embedding model of choice, SimCSE \citep{gao-2022-systeme}, which uses unsupervised self-contrasting learning to improve sentence-level representations of words.

\subsection{Model Architecture}
\subsection{Encoder Model Training and Description}\label{apdx:training}

Models were built and trained in pytorch,\footnote{\url{https://pytorch.org/docs/stable/generated/torch.nn.TransformerEncoder.html}} largely based on the Encoder modules available. Each model was trained on a single L4 GPU in Google Colab. Each epoch took about 4.25 hours: 
\begin{itemize}
    \item  8 transformer encoder layers
    \item 25 total classifier heads (with a single dense layer each) for each task (using double objective functions, results 50 total loss calculations backpropagated.)
    \item All encoder layer parameters are shared by objectives, but the trainable parameters of the single dense layer classification heads are specific to each item. 
    \item Attention heads: 32. Since a lot of semantic information were needed to be extracted from within each embedding and its neighbors, supporting an increase in multi-head self-attention mechanisms. 
    \item Hidden dimension: 2048

\end{itemize}

\subsubsection{Preventing Overfit within the Model}
An abnormally high 0.75 Dropout rate was the primary regularization technique to avoid overfit in a noisy, repetitively augmented dataset with non-gold labels. 
\begin{itemize}
    \item     \textbf{Optimizer: Adamax}: defined in the original paper by \citet{kingma_adam_2017}, this is a variant of Adam that replaces the L2 norm of the gradients with the L-infinity norm which provides stability in sparse gradients resulting from the dropout. Additionally, its initial momentum and second derivative momentum are limited slightly to 0.78 and 0.9, respectively, to prevent overfitting, but increasing training time, and increased the weight decay to 0.0003 similarly.
    \item \textbf{Learning Rate}: initial learning rate was set to 2.5e-5, within the learning rate schedule seen below.
    \item \textbf{Gradient clipping}: set to 4 (instead of the typical 1), since we did not want an unusual batch to explode, but recognizing the need to capture as much info as we can from our optimizer given dropout was a primary regularization to account for high level of repetition in the augmented transcript windows.
    \item \textbf{Learning rate schedule}: Using chaining, began linear from zero with warmup, a 1,000 step linear ramp, followed by exponential decay with gamma = 0.9995) multiplied with \texttt{CosineAnnealingWarmRestarts} from \texttt{pytorch}\footnote{\url{https://pytorch.org/}}, scheduling with annealing cycles cutting frequency by a third each time. We have initial data to suggest that using a cyclic learning rate improves model performance, but did not sufficiently ablate this additional level of complexity sufficiently to claim whether, without it, the models would still learn effectively.
    \item \textbf{Loss functions} In addition to cross-entropy loss, we use a custom loss function implementing Quadratic weighted kappa loss with fuzzy labels/label smoothing set at 0.2, to increase noise around the unreliable human ratings.

\end{itemize}

\subsection{Encoder Model Test Set}
\begin{landscape}

\begin{figure}[h]
    \centering
        \includegraphics[width=0.8\paperheight]{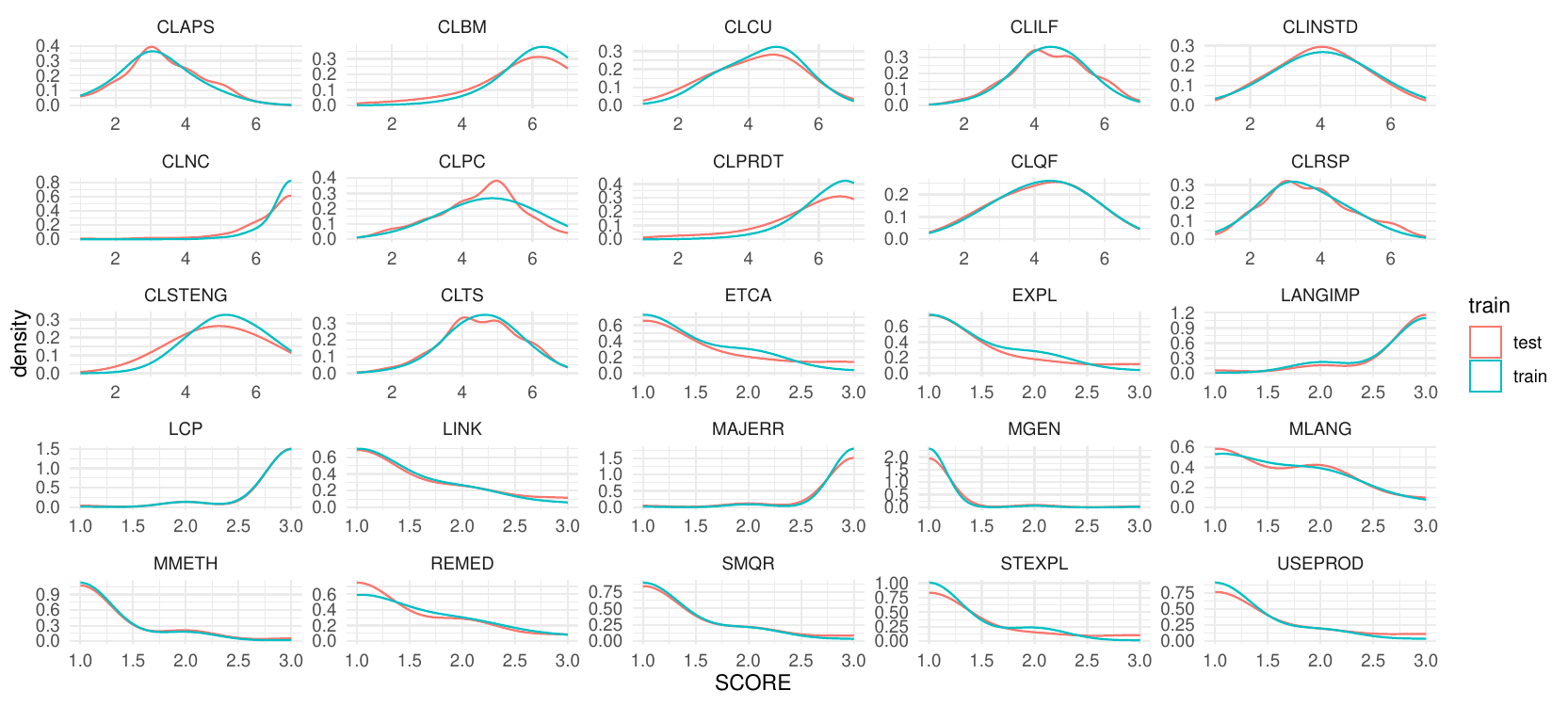}
        \caption{Test set label distributions compared to training and development sets, based on all human rater labels.
}
        \label{fig:test_set}
\end{figure}
\end{landscape}

The distributions for the held out test set for Encoder model can be found in Figure \ref{fig:test_set} compared to the training/development data. 

\newgeometry{margin=2 cm}
\restoregeometry

\section{GPT Model Family}

\subsection{Model construction}
Detailed descriptions of the three models and data generated by them can be found in the original paper and accompanying websites \citeauthor{wang_is_2023}\footnote{The automated rating data was retrieved from https://github.com/rosewang2008/zero-shot-teacher-feedback/tree/main} which  examples for how the three models differ. A brief summary of those differences can be found in Table \ref{tab:gptdiff}.

\subsubsection{GPT Model Preprocessing}
In contrast to the Encoder model preprocessing, a preliminary analysis was conducted by \citeauthor{wang_is_2023} to identify the highest quality 7.5-minute segments available in the dataset, as defined by fewest transcriber notes. The models are provided the discrourse from these selections and also information about the subset of items they provide ratings for, including four items from the MQI (\textbf{\cmtt{EXPL}}, \textbf{\cmtt{LANGIMP}}, \textbf{\cmtt{REMED}}, \textbf{\cmtt{SMQR}}).

\section{Reliability Metrics}\label{apx:rel_metrix_icc}

ICC calculations were reproduced using the following multilevel model where lesson $l$ scores for each rubric item are nested within teachers $k$:
\begin{align}\label{eq:iccrandeffects}
{ITEM}_{l k}=\beta_{0}+\mu_{k}+\varepsilon_{l k} \text{,}
\end{align}

and then calculate the ICC and Adjusted ICC
\begin{align}\label{eq:icc}
ICC = \frac{\operatorname{var}\left(\mu_{k}\right)}{\operatorname{var}\left(\mu_{k}\right)+\frac{\operatorname{var}\left(\varepsilon_{l k}\right)}{n_{l}}},
\end{align}

where $n_{l} = 1$ for ICC and where $n_{l}=6$ for Adjusted ICC following the original study. 
Full results of human baselines and comparisons against the various models can be found in Appendix \ref{apx:fullresult}.

\subsection{Full Results}\label{apx:fullresult}
Table \ref{tab:tab:full} contains the full results calculations referenced in Section \ref{sec:reliabilities}. The metric symbols found in the table are as follows: \textbf{C's $\kappa$}: Cohen's $\kappa$; \textbf{QWK}: Quadratic Weighted Kappa; \textbf{\%Agr}: percent exact agreement; \textbf{Agr±1}: percent  agreement within 1 category; \textbf{ICC} and \textbf{AdjICC}: intraclass correlation and adjusted intraclass correlation (with nested staging in Eq. \ref{eq:icc}; \textbf{$r$}: $r$, Pearson's correlation; $\mathbf{\rho}$:  $\mathbf{\rho}$, Spearman's rank correlation, $\mathbf{\tau}$:  $\mathbf{\tau}$, Kendall's rank correlation. *.low and *.hi are low and high 95\% confidence intervals at $\alpha = 0.05$, respectively. These results and full results for CLASS items can be found online.\footnote{\url{https://github.com/hardy-education/LLM-Psychometrics}}

\newpage

\begingroup\fontsize{8}{10}\selectfont

\begin{longtable}{lllrrrrrrrrrrr}
\caption{\label{tab:tab:full}Full Agreement Metrics}\\
\toprule
Instrument & Item & Metric & Human & \textbf{Encoders} & \textit{un1} & \textit{un2} & \textit{un3} & \textit{gte} & \textit{e5} & \textbf{GPTs} & \textit{N} & \textit{ND} & \textit{NR}\\
\midrule
\endfirsthead
\caption[]{Full Agreement Metrics \textit{(continued)}}\\
\toprule
Instrument & Item & Metric & Human & Encoder & un1 & un2 & un3 & gte & e5 & GPT & N & ND & NR\\
\midrule
\endhead

\endfoot
\bottomrule
\endlastfoot
MQI & LINK    & C's $\kappa$ & 0.31 & 0.39 & 0.41 & 0.33 & 0.44 & 0.39 & 0.39 &       &       &       &       \\
MQI & LINK    & QWK                              & 0.41 & 0.58 & 0.6  & 0.55 & 0.62 & 0.56 & 0.56 &       &       &       &       \\
MQI & LINK    & \%Agr                           & 0.7  & 0.73 & 0.74 & 0.71 & 0.75 & 0.71 & 0.71 &       &       &       &       \\
MQI & LINK    & Agr±1                            & 0.97 & 0.98 & 0.97 & 0.98 & 0.98 & 0.98 & 0.98 &       &       &       &       \\
MQI & LINK    & $r$                          & 0.41 & 0.58 & 0.61 & 0.56 & 0.63 & 0.56 & 0.56 &       &       &       &       \\
MQI & LINK    & $r$.low                      & 0.39 & 0.57 & 0.57 & 0.51 & 0.59 & 0.52 & 0.52 &       &       &       &       \\
MQI & LINK    & $r$.hi                       & 0.42 & 0.6  & 0.64 & 0.6  & 0.66 & 0.6  & 0.6  &       &       &       &       \\
MQI & LINK    & $\rho$       & 0.41 & 0.57 & 0.6  & 0.53 & 0.61 & 0.54 & 0.54 &       &       &       &       \\
MQI & LINK    & $\rho$.low   & 0.4  & 0.55 & 0.56 & 0.48 & 0.58 & 0.5  & 0.5  &       &       &       &       \\
MQI & LINK    & $\rho$.hi    & 0.43 & 0.58 & 0.64 & 0.58 & 0.65 & 0.58 & 0.58 &       &       &       &       \\
MQI & LINK    & $\tau$       & 0.4  & 0.54 & 0.57 & 0.51 & 0.59 & 0.51 & 0.51 &       &       &       &       \\
MQI & LINK    & $\tau$.low   & 0.38 & 0.52 & 0.53 & 0.46 & 0.55 & 0.47 & 0.47 &       &       &       &       \\
MQI & LINK    & $\tau$.hi    & 0.41 & 0.56 & 0.61 & 0.56 & 0.62 & 0.56 & 0.56 &       &       &       &       \\
MQI & LINK    & ICC                              & 0.15 & 0.14 & 0.14 & 0.14 & 0.14 & 0.14 & 0.14 &       &       &       &       \\
MQI & LINK    & AdjICC                           & 0.51 & 0.5  & 0.5  & 0.5  & 0.5  & 0.5  & 0.5  &       &       &       &       \\
MQI & EXPL    & C's $\kappa$ & 0.23 & 0.25 & 0.25 & 0.28 & 0.23 & 0.24 & 0.24 & 0.03  & 0.01  & 0.07  & 0.01  \\
MQI & EXPL    & QWK                              & 0.28 & 0.43 & 0.46 & 0.42 & 0.44 & 0.4  & 0.4  & 0.01  & 0.01  & 0.06  & -0.01 \\
MQI & EXPL    & \%Agr                           & 0.7  & 0.72 & 0.72 & 0.69 & 0.72 & 0.72 & 0.72 & 0.31  & 0.31  & 0.42  & 0.15  \\
MQI & EXPL    & Agr±1                            & 0.98 & 0.97 & 0.97 & 0.97 & 0.96 & 0.97 & 0.97 & 0.86  & 0.95  & 0.9   & 0.67  \\
MQI & EXPL    & $r$                          & 0.28 & 0.44 & 0.48 & 0.43 & 0.47 & 0.41 & 0.41 & 0.03  & 0.03  & 0.09  & -0.03 \\
MQI & EXPL    & $r$.low                      & 0.26 & 0.42 & 0.44 & 0.37 & 0.42 & 0.36 & 0.36 & -0.03 & -0.07 & -0.01 & -0.14 \\
MQI & EXPL    & $r$.hi                       & 0.29 & 0.46 & 0.52 & 0.48 & 0.51 & 0.46 & 0.46 & 0.08  & 0.13  & 0.19  & 0.09  \\
MQI & EXPL    & $\rho$       & 0.27 & 0.42 & 0.46 & 0.41 & 0.46 & 0.39 & 0.39 & 0.03  & 0.03  & 0.1   & -0.03 \\
MQI & EXPL    & $\rho$.low   & 0.25 & 0.4  & 0.42 & 0.35 & 0.41 & 0.34 & 0.34 & -0.03 & -0.07 & 0     & -0.14 \\
MQI & EXPL    & $\rho$.hi    & 0.29 & 0.44 & 0.51 & 0.46 & 0.5  & 0.43 & 0.43 & 0.09  & 0.13  & 0.19  & 0.08  \\
MQI & EXPL    & $\tau$       & 0.26 & 0.41 & 0.45 & 0.39 & 0.44 & 0.38 & 0.38 & 0.03  & 0.03  & 0.09  & -0.03 \\
MQI & EXPL    & $\tau$.low   & 0.25 & 0.39 & 0.4  & 0.33 & 0.4  & 0.32 & 0.32 & -0.03 & -0.07 & -0.01 & -0.14 \\
MQI & EXPL    & $\tau$.hi    & 0.28 & 0.43 & 0.49 & 0.45 & 0.49 & 0.42 & 0.42 & 0.08  & 0.12  & 0.19  & 0.08  \\
MQI & EXPL    & ICC                              & 0.17 & 0.17 & 0.17 & 0.17 & 0.17 & 0.17 & 0.17 & 0.17  & 0.17  & 0.17  & 0.17  \\
MQI & EXPL    & AdjICC                           & 0.55 & 0.56 & 0.56 & 0.56 & 0.56 & 0.56 & 0.56 & 0.55  & 0.55  & 0.55  & 0.55  \\
MQI & MMETH   & C's $\kappa$ & 0.42 & 0.33 & 0.46 & 0.39 & 0.33 & 0.27 & 0.27 &       &       &       &       \\
MQI & MMETH   & QWK                              & 0.47 & 0.49 & 0.48 & 0.53 & 0.54 & 0.46 & 0.46 &       &       &       &       \\
MQI & MMETH   & \%Agr                           & 0.85 & 0.82 & 0.88 & 0.86 & 0.84 & 0.78 & 0.78 &       &       &       &       \\
MQI & MMETH   & Agr±1                            & 0.99 & 0.98 & 0.99 & 0.98 & 0.98 & 0.97 & 0.97 &       &       &       &       \\
MQI & MMETH   & $r$                          & 0.47 & 0.52 & 0.51 & 0.58 & 0.57 & 0.51 & 0.51 &       &       &       &       \\
MQI & MMETH   & $r$.low                      & 0.46 & 0.5  & 0.46 & 0.53 & 0.53 & 0.47 & 0.47 &       &       &       &       \\
MQI & MMETH   & $r$.hi                       & 0.49 & 0.54 & 0.55 & 0.62 & 0.61 & 0.56 & 0.56 &       &       &       &       \\
MQI & MMETH   & $\rho$       & 0.47 & 0.5  & 0.51 & 0.57 & 0.57 & 0.48 & 0.48 &       &       &       &       \\
MQI & MMETH   & $\rho$.low   & 0.45 & 0.49 & 0.47 & 0.52 & 0.53 & 0.43 & 0.43 &       &       &       &       \\
MQI & MMETH   & $\rho$.hi    & 0.48 & 0.52 & 0.55 & 0.61 & 0.61 & 0.52 & 0.52 &       &       &       &       \\
MQI & MMETH   & $\tau$       & 0.46 & 0.49 & 0.51 & 0.56 & 0.56 & 0.46 & 0.46 &       &       &       &       \\
MQI & MMETH   & $\tau$.low   & 0.45 & 0.47 & 0.46 & 0.52 & 0.52 & 0.42 & 0.42 &       &       &       &       \\
MQI & MMETH   & $\tau$.hi    & 0.48 & 0.51 & 0.55 & 0.61 & 0.6  & 0.51 & 0.51 &       &       &       &       \\
MQI & MMETH   & ICC                              & 0.15 & 0.18 & 0.18 & 0.18 & 0.18 & 0.18 & 0.18 &       &       &       &       \\
MQI & MMETH   & AdjICC                           & 0.52 & 0.57 & 0.57 & 0.57 & 0.57 & 0.57 & 0.57 &       &       &       &       \\
MQI & MGEN    & C's $\kappa$ & 0.15 & 0.26 & 0.27 & 0.32 & 0.27 & 0.24 & 0.24 &       &       &       &       \\
MQI & MGEN    & QWK                              & 0.19 & 0.34 & 0.34 & 0.48 & 0.34 & 0.29 & 0.29 &       &       &       &       \\
MQI & MGEN    & \%Agr                           & 0.95 & 0.95 & 0.96 & 0.96 & 0.96 & 0.94 & 0.94 &       &       &       &       \\
MQI & MGEN    & Agr±1                            & 0.99 & 1    & 1    & 1    & 1    & 0.99 & 0.99 &       &       &       &       \\
MQI & MGEN    & $r$                          & 0.19 & 0.34 & 0.37 & 0.48 & 0.37 & 0.29 & 0.29 &       &       &       &       \\
MQI & MGEN    & $r$.low                      & 0.18 & 0.32 & 0.32 & 0.43 & 0.32 & 0.24 & 0.24 &       &       &       &       \\
MQI & MGEN    & $r$.hi                       & 0.21 & 0.37 & 0.42 & 0.53 & 0.42 & 0.34 & 0.34 &       &       &       &       \\
MQI & MGEN    & $\rho$       & 0.19 & 0.32 & 0.34 & 0.42 & 0.34 & 0.28 & 0.28 &       &       &       &       \\
MQI & MGEN    & $\rho$.low   & 0.17 & 0.29 & 0.29 & 0.37 & 0.29 & 0.22 & 0.22 &       &       &       &       \\
MQI & MGEN    & $\rho$.hi    & 0.2  & 0.34 & 0.39 & 0.48 & 0.39 & 0.33 & 0.33 &       &       &       &       \\
MQI & MGEN    & $\tau$       & 0.18 & 0.32 & 0.34 & 0.42 & 0.34 & 0.27 & 0.27 &       &       &       &       \\
MQI & MGEN    & $\tau$.low   & 0.17 & 0.29 & 0.29 & 0.37 & 0.29 & 0.22 & 0.22 &       &       &       &       \\
MQI & MGEN    & $\tau$.hi    & 0.2  & 0.34 & 0.39 & 0.48 & 0.39 & 0.33 & 0.33 &       &       &       &       \\
MQI & MGEN    & ICC                              & 0.04 & 0.03 & 0.03 & 0.03 & 0.03 & 0.03 & 0.03 &       &       &       &       \\
MQI & MGEN    & AdjICC                           & 0.19 & 0.16 & 0.16 & 0.16 & 0.16 & 0.16 & 0.16 &       &       &       &       \\
MQI & MLANG   & C's $\kappa$ & 0.23 & 0.37 & 0.4  & 0.44 & 0.43 & 0.31 & 0.31 &       &       &       &       \\
MQI & MLANG   & QWK                              & 0.33 & 0.48 & 0.49 & 0.55 & 0.51 & 0.44 & 0.44 &       &       &       &       \\
MQI & MLANG   & \%Agr                           & 0.59 & 0.65 & 0.68 & 0.69 & 0.7  & 0.6  & 0.6  &       &       &       &       \\
MQI & MLANG   & Agr±1                            & 0.98 & 0.98 & 0.99 & 0.99 & 0.99 & 0.98 & 0.98 &       &       &       &       \\
MQI & MLANG   & $r$                          & 0.33 & 0.48 & 0.5  & 0.55 & 0.52 & 0.46 & 0.46 &       &       &       &       \\
MQI & MLANG   & $r$.low                      & 0.32 & 0.46 & 0.45 & 0.5  & 0.47 & 0.41 & 0.41 &       &       &       &       \\
MQI & MLANG   & $r$.hi                       & 0.35 & 0.5  & 0.54 & 0.59 & 0.56 & 0.5  & 0.5  &       &       &       &       \\
MQI & MLANG   & $\rho$       & 0.32 & 0.47 & 0.48 & 0.54 & 0.5  & 0.43 & 0.43 &       &       &       &       \\
MQI & MLANG   & $\rho$.low   & 0.31 & 0.45 & 0.43 & 0.49 & 0.46 & 0.39 & 0.39 &       &       &       &       \\
MQI & MLANG   & $\rho$.hi    & 0.34 & 0.49 & 0.52 & 0.59 & 0.55 & 0.48 & 0.48 &       &       &       &       \\
MQI & MLANG   & $\tau$       & 0.31 & 0.45 & 0.46 & 0.52 & 0.49 & 0.41 & 0.41 &       &       &       &       \\
MQI & MLANG   & $\tau$.low   & 0.29 & 0.43 & 0.42 & 0.47 & 0.45 & 0.36 & 0.36 &       &       &       &       \\
MQI & MLANG   & $\tau$.hi    & 0.33 & 0.47 & 0.51 & 0.57 & 0.53 & 0.46 & 0.46 &       &       &       &       \\
MQI & MLANG   & ICC                              & 0.08 & 0.09 & 0.09 & 0.09 & 0.09 & 0.09 & 0.09 &       &       &       &       \\
MQI & MLANG   & AdjICC                           & 0.34 & 0.36 & 0.36 & 0.36 & 0.36 & 0.36 & 0.36 &       &       &       &       \\
MQI & REMED   & C's $\kappa$ & 0.27 & 0.3  & 0.27 & 0.34 & 0.35 & 0.27 & 0.27 & -0.01 & -0.01 & 0     & 0     \\
MQI & REMED   & QWK                              & 0.32 & 0.44 & 0.44 & 0.52 & 0.42 & 0.42 & 0.42 & 0.02  & 0     & 0.06  & 0.02  \\
MQI & REMED   & \%Agr                           & 0.66 & 0.69 & 0.68 & 0.68 & 0.74 & 0.67 & 0.67 & 0.16  & 0.1   & 0.27  & 0.08  \\
MQI & REMED   & Agr±1                            & 0.96 & 0.96 & 0.94 & 0.96 & 0.99 & 0.96 & 0.96 & 0.62  & 0.54  & 0.81  & 0.48  \\
MQI & REMED   & $r$                          & 0.32 & 0.44 & 0.46 & 0.52 & 0.45 & 0.42 & 0.42 & 0.05  & -0.01 & 0.11  & 0.11  \\
MQI & REMED   & $r$.low                      & 0.31 & 0.42 & 0.41 & 0.47 & 0.4  & 0.37 & 0.37 & 0     & -0.11 & 0.01  & -0.01 \\
MQI & REMED   & $r$.hi                       & 0.34 & 0.47 & 0.5  & 0.57 & 0.49 & 0.47 & 0.47 & 0.11  & 0.09  & 0.21  & 0.22  \\
MQI & REMED   & $\rho$       & 0.32 & 0.42 & 0.44 & 0.49 & 0.44 & 0.38 & 0.38 & 0.06  & 0     & 0.12  & 0.09  \\
MQI & REMED   & $\rho$.low   & 0.31 & 0.4  & 0.39 & 0.43 & 0.4  & 0.33 & 0.33 & 0     & -0.1  & 0.02  & -0.02 \\
MQI & REMED   & $\rho$.hi    & 0.34 & 0.44 & 0.48 & 0.54 & 0.49 & 0.43 & 0.43 & 0.12  & 0.1   & 0.22  & 0.2   \\
MQI & REMED   & $\tau$       & 0.31 & 0.4  & 0.42 & 0.46 & 0.43 & 0.37 & 0.37 & 0.06  & 0     & 0.11  & 0.09  \\
MQI & REMED   & $\tau$.low   & 0.3  & 0.38 & 0.37 & 0.41 & 0.39 & 0.32 & 0.32 & 0     & -0.1  & 0.02  & -0.02 \\
MQI & REMED   & $\tau$.hi    & 0.33 & 0.42 & 0.46 & 0.51 & 0.48 & 0.41 & 0.41 & 0.11  & 0.1   & 0.21  & 0.2   \\
MQI & REMED   & ICC                              & 0.16 & 0.17 & 0.17 & 0.17 & 0.17 & 0.17 & 0.17 & 0.14  & 0.14  & 0.14  & 0.14  \\
MQI & REMED   & AdjICC                           & 0.53 & 0.55 & 0.55 & 0.55 & 0.55 & 0.55 & 0.55 & 0.5   & 0.5   & 0.5   & 0.5   \\
MQI & USEPROD & C's $\kappa$ & 0.25 & 0.3  & 0.28 & 0.32 & 0.3  & 0.31 & 0.31 &       &       &       &       \\
MQI & USEPROD & QWK                              & 0.33 & 0.46 & 0.44 & 0.5  & 0.46 & 0.46 & 0.46 &       &       &       &       \\
MQI & USEPROD & \%Agr                           & 0.76 & 0.75 & 0.74 & 0.8  & 0.75 & 0.74 & 0.74 &       &       &       &       \\
MQI & USEPROD & Agr±1                            & 0.98 & 0.95 & 0.93 & 0.97 & 0.94 & 0.95 & 0.95 &       &       &       &       \\
MQI & USEPROD & $r$                          & 0.33 & 0.49 & 0.48 & 0.5  & 0.5  & 0.49 & 0.49 &       &       &       &       \\
MQI & USEPROD & $r$.low                      & 0.32 & 0.47 & 0.43 & 0.45 & 0.46 & 0.45 & 0.45 &       &       &       &       \\
MQI & USEPROD & $r$.hi                       & 0.35 & 0.51 & 0.52 & 0.55 & 0.54 & 0.53 & 0.53 &       &       &       &       \\
MQI & USEPROD & $\rho$       & 0.31 & 0.47 & 0.47 & 0.45 & 0.49 & 0.46 & 0.46 &       &       &       &       \\
MQI & USEPROD & $\rho$.low   & 0.29 & 0.45 & 0.42 & 0.39 & 0.45 & 0.42 & 0.42 &       &       &       &       \\
MQI & USEPROD & $\rho$.hi    & 0.32 & 0.49 & 0.51 & 0.5  & 0.53 & 0.51 & 0.51 &       &       &       &       \\
MQI & USEPROD & $\tau$       & 0.3  & 0.45 & 0.45 & 0.44 & 0.48 & 0.45 & 0.45 &       &       &       &       \\
MQI & USEPROD & $\tau$.low   & 0.29 & 0.43 & 0.41 & 0.38 & 0.43 & 0.4  & 0.4  &       &       &       &       \\
MQI & USEPROD & $\tau$.hi    & 0.32 & 0.47 & 0.5  & 0.49 & 0.52 & 0.49 & 0.49 &       &       &       &       \\
MQI & USEPROD & ICC                              & 0.24 & 0.23 & 0.23 & 0.23 & 0.23 & 0.23 & 0.23 &       &       &       &       \\
MQI & USEPROD & AdjICC                           & 0.65 & 0.64 & 0.64 & 0.64 & 0.64 & 0.64 & 0.64 &       &       &       &       \\
MQI & MAJERR  & C's $\kappa$ & 0.24 & 0.22 & 0.27 & 0.26 & 0.22 & 0.19 & 0.19 &       &       &       &       \\
MQI & MAJERR  & QWK                              & 0.28 & 0.35 & 0.35 & 0.45 & 0.43 & 0.29 & 0.29 &       &       &       &       \\
MQI & MAJERR  & \%Agr                           & 0.91 & 0.9  & 0.92 & 0.91 & 0.92 & 0.87 & 0.87 &       &       &       &       \\
MQI & MAJERR  & Agr±1                            & 0.99 & 0.99 & 1    & 0.99 & 0.99 & 0.98 & 0.98 &       &       &       &       \\
MQI & MAJERR  & $r$                          & 0.28 & 0.36 & 0.38 & 0.45 & 0.44 & 0.31 & 0.31 &       &       &       &       \\
MQI & MAJERR  & $r$.low                      & 0.26 & 0.34 & 0.33 & 0.4  & 0.39 & 0.26 & 0.26 &       &       &       &       \\
MQI & MAJERR  & $r$.hi                       & 0.29 & 0.38 & 0.43 & 0.5  & 0.49 & 0.36 & 0.36 &       &       &       &       \\
MQI & MAJERR  & $\rho$       & 0.28 & 0.31 & 0.34 & 0.43 & 0.38 & 0.27 & 0.27 &       &       &       &       \\
MQI & MAJERR  & $\rho$.low   & 0.26 & 0.29 & 0.28 & 0.37 & 0.33 & 0.21 & 0.21 &       &       &       &       \\
MQI & MAJERR  & $\rho$.hi    & 0.29 & 0.33 & 0.39 & 0.48 & 0.43 & 0.32 & 0.32 &       &       &       &       \\
MQI & MAJERR  & $\tau$       & 0.28 & 0.31 & 0.33 & 0.42 & 0.37 & 0.26 & 0.26 &       &       &       &       \\
MQI & MAJERR  & $\tau$.low   & 0.26 & 0.28 & 0.28 & 0.36 & 0.32 & 0.21 & 0.21 &       &       &       &       \\
MQI & MAJERR  & $\tau$.hi    & 0.29 & 0.33 & 0.38 & 0.47 & 0.42 & 0.32 & 0.32 &       &       &       &       \\
MQI & MAJERR  & ICC                              & 0.1  & 0.06 & 0.06 & 0.06 & 0.06 & 0.06 & 0.06 &       &       &       &       \\
MQI & MAJERR  & AdjICC                           & 0.39 & 0.29 & 0.29 & 0.29 & 0.29 & 0.29 & 0.29 &       &       &       &       \\
MQI & LANGIMP & C's $\kappa$ & 0.25 & 0.2  & 0.32 & 0.21 & 0.21 & 0.15 & 0.15 & 0     & 0     & -0.03 & 0.03  \\
MQI & LANGIMP & QWK                              & 0.29 & 0.34 & 0.36 & 0.43 & 0.39 & 0.29 & 0.29 & -0.01 & -0.01 & -0.05 & 0.03  \\
MQI & LANGIMP & \%Agr                           & 0.8  & 0.8  & 0.86 & 0.81 & 0.83 & 0.75 & 0.75 & 0.32  & 0.25  & 0.38  & 0.33  \\
MQI & LANGIMP & Agr±1                            & 0.99 & 0.98 & 1    & 0.99 & 0.98 & 0.97 & 0.97 & 0.98  & 0.97  & 0.98  & 0.99  \\
MQI & LANGIMP & $r$                          & 0.29 & 0.35 & 0.4  & 0.44 & 0.4  & 0.31 & 0.31 & -0.02 & -0.02 & -0.08 & 0.06  \\
MQI & LANGIMP & $r$.low                      & 0.27 & 0.33 & 0.35 & 0.38 & 0.35 & 0.26 & 0.26 & -0.08 & -0.12 & -0.17 & -0.05 \\
MQI & LANGIMP & $r$.hi                       & 0.3  & 0.37 & 0.45 & 0.49 & 0.45 & 0.36 & 0.36 & 0.04  & 0.07  & 0.02  & 0.17  \\
MQI & LANGIMP & $\rho$       & 0.28 & 0.31 & 0.38 & 0.4  & 0.37 & 0.26 & 0.26 & -0.02 & -0.03 & -0.08 & 0.05  \\
MQI & LANGIMP & $\rho$.low   & 0.26 & 0.29 & 0.33 & 0.34 & 0.32 & 0.21 & 0.21 & -0.08 & -0.13 & -0.17 & -0.06 \\
MQI & LANGIMP & $\rho$.hi    & 0.29 & 0.34 & 0.43 & 0.45 & 0.42 & 0.32 & 0.32 & 0.03  & 0.07  & 0.02  & 0.17  \\
MQI & LANGIMP & $\tau$       & 0.28 & 0.31 & 0.38 & 0.39 & 0.37 & 0.26 & 0.26 & -0.02 & -0.03 & -0.07 & 0.05  \\
MQI & LANGIMP & $\tau$.low   & 0.26 & 0.28 & 0.33 & 0.33 & 0.31 & 0.2  & 0.2  & -0.08 & -0.13 & -0.17 & -0.06 \\
MQI & LANGIMP & $\tau$.hi    & 0.29 & 0.33 & 0.43 & 0.45 & 0.41 & 0.31 & 0.31 & 0.03  & 0.07  & 0.02  & 0.16  \\
MQI & LANGIMP & ICC                              & 0.12 & 0.13 & 0.13 & 0.13 & 0.13 & 0.13 & 0.13 & 0.12  & 0.12  & 0.12  & 0.12  \\
MQI & LANGIMP & AdjICC                           & 0.44 & 0.47 & 0.47 & 0.47 & 0.47 & 0.47 & 0.47 & 0.45  & 0.45  & 0.45  & 0.45  \\
MQI & LCP     & C's $\kappa$ & 0.18 & 0.2  & 0.26 & 0.25 & 0.18 & 0.17 & 0.17 &       &       &       &       \\
MQI & LCP     & QWK                              & 0.23 & 0.32 & 0.32 & 0.44 & 0.36 & 0.25 & 0.25 &       &       &       &       \\
MQI & LCP     & \%Agr                           & 0.86 & 0.86 & 0.89 & 0.87 & 0.89 & 0.83 & 0.83 &       &       &       &       \\
MQI & LCP     & Agr±1                            & 0.99 & 0.98 & 0.99 & 0.98 & 0.98 & 0.98 & 0.98 &       &       &       &       \\
MQI & LCP     & $r$                          & 0.23 & 0.32 & 0.36 & 0.45 & 0.37 & 0.25 & 0.25 &       &       &       &       \\
MQI & LCP     & $r$.low                      & 0.22 & 0.3  & 0.31 & 0.39 & 0.32 & 0.2  & 0.2  &       &       &       &       \\
MQI & LCP     & $r$.hi                       & 0.25 & 0.34 & 0.41 & 0.5  & 0.42 & 0.31 & 0.31 &       &       &       &       \\
MQI & LCP     & $\rho$       & 0.22 & 0.28 & 0.33 & 0.41 & 0.34 & 0.21 & 0.21 &       &       &       &       \\
MQI & LCP     & $\rho$.low   & 0.2  & 0.25 & 0.28 & 0.35 & 0.29 & 0.15 & 0.15 &       &       &       &       \\
MQI & LCP     & $\rho$.hi    & 0.23 & 0.3  & 0.38 & 0.46 & 0.39 & 0.26 & 0.26 &       &       &       &       \\
MQI & LCP     & $\tau$       & 0.21 & 0.27 & 0.33 & 0.41 & 0.34 & 0.21 & 0.21 &       &       &       &       \\
MQI & LCP     & $\tau$.low   & 0.2  & 0.25 & 0.27 & 0.35 & 0.29 & 0.15 & 0.15 &       &       &       &       \\
MQI & LCP     & $\tau$.hi    & 0.23 & 0.3  & 0.38 & 0.46 & 0.39 & 0.26 & 0.26 &       &       &       &       \\
MQI & LCP     & ICC                              & 0.14 & 0.15 & 0.15 & 0.15 & 0.15 & 0.15 & 0.15 &       &       &       &       \\
MQI & LCP     & AdjICC                           & 0.5  & 0.51 & 0.51 & 0.51 & 0.51 & 0.51 & 0.51 &       &       &       &       \\
MQI & STEXPL  & C's $\kappa$ & 0.36 & 0.29 & 0.26 & 0.3  & 0.26 & 0.31 & 0.31 &       &       &       &       \\
MQI & STEXPL  & QWK                              & 0.4  & 0.45 & 0.45 & 0.45 & 0.48 & 0.45 & 0.45 &       &       &       &       \\
MQI & STEXPL  & \%Agr                           & 0.8  & 0.77 & 0.76 & 0.79 & 0.77 & 0.77 & 0.77 &       &       &       &       \\
MQI & STEXPL  & Agr±1                            & 0.99 & 0.97 & 0.97 & 0.97 & 0.98 & 0.97 & 0.97 &       &       &       &       \\
MQI & STEXPL  & $r$                          & 0.4  & 0.48 & 0.48 & 0.46 & 0.51 & 0.48 & 0.48 &       &       &       &       \\
MQI & STEXPL  & $r$.low                      & 0.38 & 0.46 & 0.44 & 0.4  & 0.47 & 0.43 & 0.43 &       &       &       &       \\
MQI & STEXPL  & $r$.hi                       & 0.41 & 0.5  & 0.53 & 0.51 & 0.56 & 0.52 & 0.52 &       &       &       &       \\
MQI & STEXPL  & $\rho$       & 0.39 & 0.47 & 0.47 & 0.46 & 0.5  & 0.47 & 0.47 &       &       &       &       \\
MQI & STEXPL  & $\rho$.low   & 0.38 & 0.45 & 0.42 & 0.4  & 0.46 & 0.43 & 0.43 &       &       &       &       \\
MQI & STEXPL  & $\rho$.hi    & 0.41 & 0.49 & 0.51 & 0.51 & 0.54 & 0.52 & 0.52 &       &       &       &       \\
MQI & STEXPL  & $\tau$       & 0.39 & 0.46 & 0.46 & 0.45 & 0.49 & 0.46 & 0.46 &       &       &       &       \\
MQI & STEXPL  & $\tau$.low   & 0.37 & 0.44 & 0.41 & 0.39 & 0.45 & 0.41 & 0.41 &       &       &       &       \\
MQI & STEXPL  & $\tau$.hi    & 0.4  & 0.48 & 0.5  & 0.5  & 0.53 & 0.51 & 0.51 &       &       &       &       \\
MQI & STEXPL  & ICC                              & 0.3  & 0.27 & 0.27 & 0.27 & 0.27 & 0.27 & 0.27 &       &       &       &       \\
MQI & STEXPL  & AdjICC                           & 0.72 & 0.69 & 0.69 & 0.69 & 0.69 & 0.69 & 0.69 &       &       &       &       \\
MQI & SMQR    & C's $\kappa$ & 0.25 & 0.3  & 0.23 & 0.29 & 0.35 & 0.32 & 0.32 & 0.07  & 0.1   & 0.09  & 0     \\
MQI & SMQR    & QWK                              & 0.3  & 0.41 & 0.45 & 0.41 & 0.41 & 0.37 & 0.37 & 0.08  & 0.09  & 0.07  & 0.06  \\
MQI & SMQR    & \%Agr                           & 0.76 & 0.76 & 0.75 & 0.77 & 0.78 & 0.75 & 0.75 & 0.4   & 0.42  & 0.48  & 0.25  \\
MQI & SMQR    & Agr±1                            & 0.98 & 0.99 & 0.97 & 0.97 & 0.99 & 0.99 & 0.99 & 0.9   & 0.91  & 0.88  & 0.93  \\
MQI & SMQR    & $r$                          & 0.3  & 0.41 & 0.46 & 0.41 & 0.43 & 0.38 & 0.38 & 0.13  & 0.16  & 0.11  & 0.13  \\
MQI & SMQR    & $r$.low                      & 0.29 & 0.39 & 0.42 & 0.36 & 0.38 & 0.33 & 0.33 & 0.07  & 0.06  & 0.01  & 0.02  \\
MQI & SMQR    & $r$.hi                       & 0.32 & 0.43 & 0.51 & 0.47 & 0.47 & 0.43 & 0.43 & 0.19  & 0.25  & 0.2   & 0.24  \\
MQI & SMQR    & $\rho$       & 0.29 & 0.39 & 0.41 & 0.4  & 0.42 & 0.37 & 0.37 & 0.12  & 0.16  & 0.11  & 0.12  \\
MQI & SMQR    & $\rho$.low   & 0.28 & 0.37 & 0.36 & 0.34 & 0.37 & 0.32 & 0.32 & 0.06  & 0.06  & 0.01  & 0.01  \\
MQI & SMQR    & $\rho$.hi    & 0.31 & 0.42 & 0.46 & 0.46 & 0.46 & 0.42 & 0.42 & 0.18  & 0.25  & 0.2   & 0.23  \\
MQI & SMQR    & $\tau$       & 0.29 & 0.38 & 0.4  & 0.39 & 0.41 & 0.37 & 0.37 & 0.12  & 0.15  & 0.1   & 0.11  \\
MQI & SMQR    & $\tau$.low   & 0.27 & 0.36 & 0.35 & 0.33 & 0.36 & 0.32 & 0.32 & 0.06  & 0.05  & 0     & 0     \\
MQI & SMQR    & $\tau$.hi    & 0.3  & 0.41 & 0.45 & 0.45 & 0.46 & 0.42 & 0.42 & 0.17  & 0.24  & 0.2   & 0.23  \\
MQI & SMQR    & ICC                              & 0.19 & 0.19 & 0.19 & 0.19 & 0.19 & 0.19 & 0.19 & 0.19  & 0.19  & 0.19  & 0.19  \\
MQI & SMQR    & AdjICC                           & 0.59 & 0.59 & 0.59 & 0.59 & 0.59 & 0.59 & 0.59 & 0.59  & 0.59  & 0.59  & 0.59  \\
MQI & ETCA    & C's $\kappa$ & 0.24 & 0.3  & 0.28 & 0.39 & 0.27 & 0.31 & 0.31 &       &       &       &       \\
MQI & ETCA    & QWK                              & 0.32 & 0.5  & 0.5  & 0.55 & 0.51 & 0.48 & 0.48 &       &       &       &       \\
MQI & ETCA    & \%Agr                           & 0.67 & 0.68 & 0.66 & 0.74 & 0.65 & 0.69 & 0.69 &       &       &       &       \\
MQI & ETCA    & Agr±1                            & 0.98 & 0.97 & 0.96 & 0.98 & 0.96 & 0.98 & 0.98 &       &       &       &       \\
MQI & ETCA    & $r$                          & 0.32 & 0.52 & 0.52 & 0.56 & 0.55 & 0.48 & 0.48 &       &       &       &       \\
MQI & ETCA    & $r$.low                      & 0.3  & 0.5  & 0.48 & 0.51 & 0.51 & 0.44 & 0.44 &       &       &       &       \\
MQI & ETCA    & $r$.hi                       & 0.33 & 0.54 & 0.56 & 0.6  & 0.59 & 0.53 & 0.53 &       &       &       &       \\
MQI & ETCA    & $\rho$       & 0.3  & 0.5  & 0.51 & 0.54 & 0.55 & 0.46 & 0.46 &       &       &       &       \\
MQI & ETCA    & $\rho$.low   & 0.28 & 0.48 & 0.47 & 0.49 & 0.51 & 0.42 & 0.42 &       &       &       &       \\
MQI & ETCA    & $\rho$.hi    & 0.31 & 0.52 & 0.55 & 0.59 & 0.59 & 0.51 & 0.51 &       &       &       &       \\
MQI & ETCA    & $\tau$       & 0.29 & 0.48 & 0.49 & 0.52 & 0.53 & 0.44 & 0.44 &       &       &       &       \\
MQI & ETCA    & $\tau$.low   & 0.27 & 0.46 & 0.45 & 0.47 & 0.48 & 0.4  & 0.4  &       &       &       &       \\
MQI & ETCA    & $\tau$.hi    & 0.31 & 0.5  & 0.54 & 0.57 & 0.57 & 0.49 & 0.49 &       &       &       &       \\
MQI & ETCA    & ICC                              & 0.21 & 0.22 & 0.22 & 0.22 & 0.22 & 0.22 & 0.22 &       &       &       &       \\
MQI & ETCA    & AdjICC                           & 0.61 & 0.63 & 0.63 & 0.63 & 0.63 & 0.63 & 0.63 &       &       &       &       \\*
\end{longtable}
\endgroup{}
\newpage

\newpage

\section{Disentangling Bias and Measuring Fairness}\label{apx:JAGS}
\begin{landscape}
    
\begin{figure}[h]
    \centering
        \includegraphics[width=0.85\paperheight]{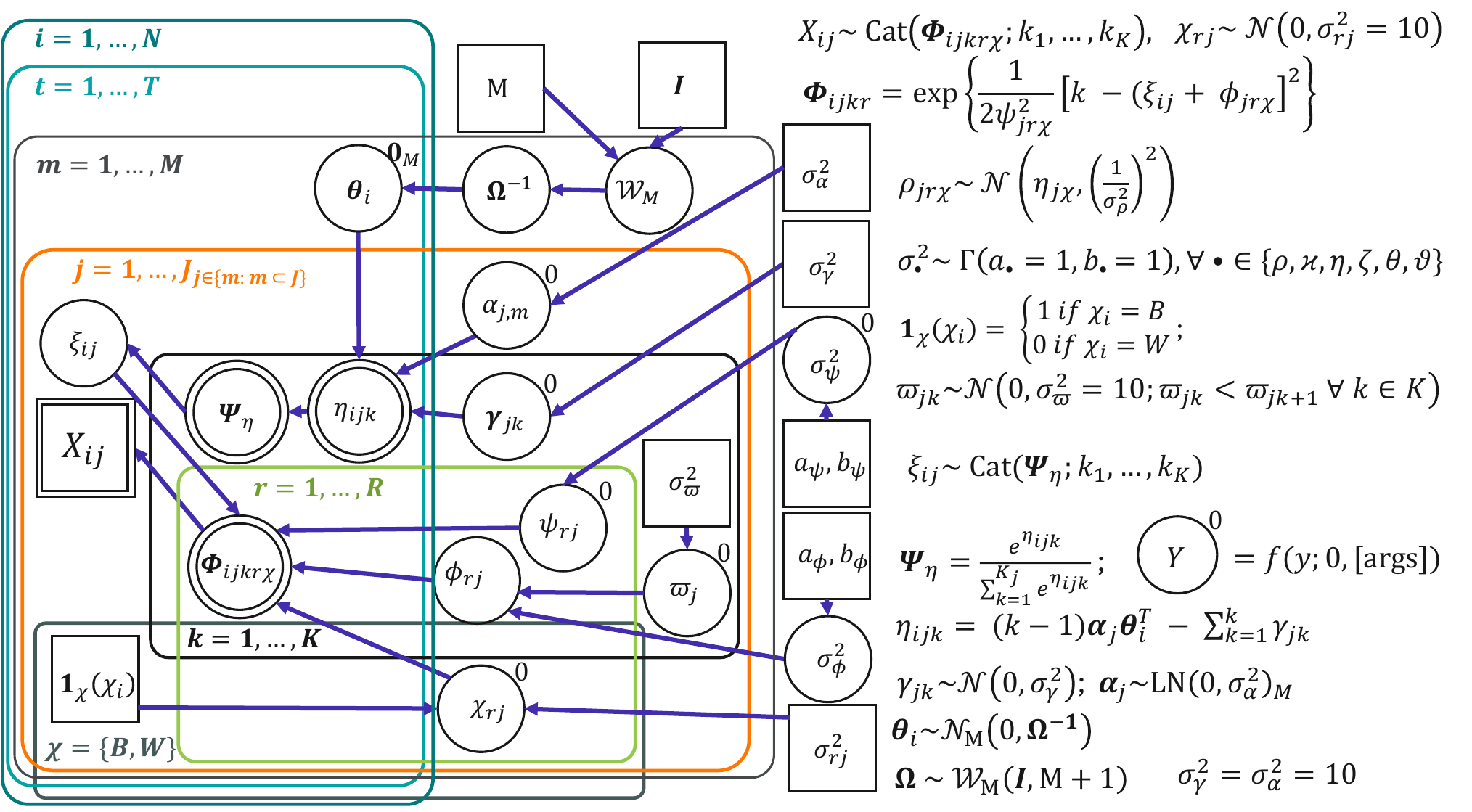}
        \caption{Structural plate diagram for model described in Section \ref{sec:fairness}.
}
        \label{fig:plate}
\end{figure}
\end{landscape}

\begin{landscape}
    
\begin{figure}[h]
    \centering
        \includegraphics[width=0.8\paperheight]{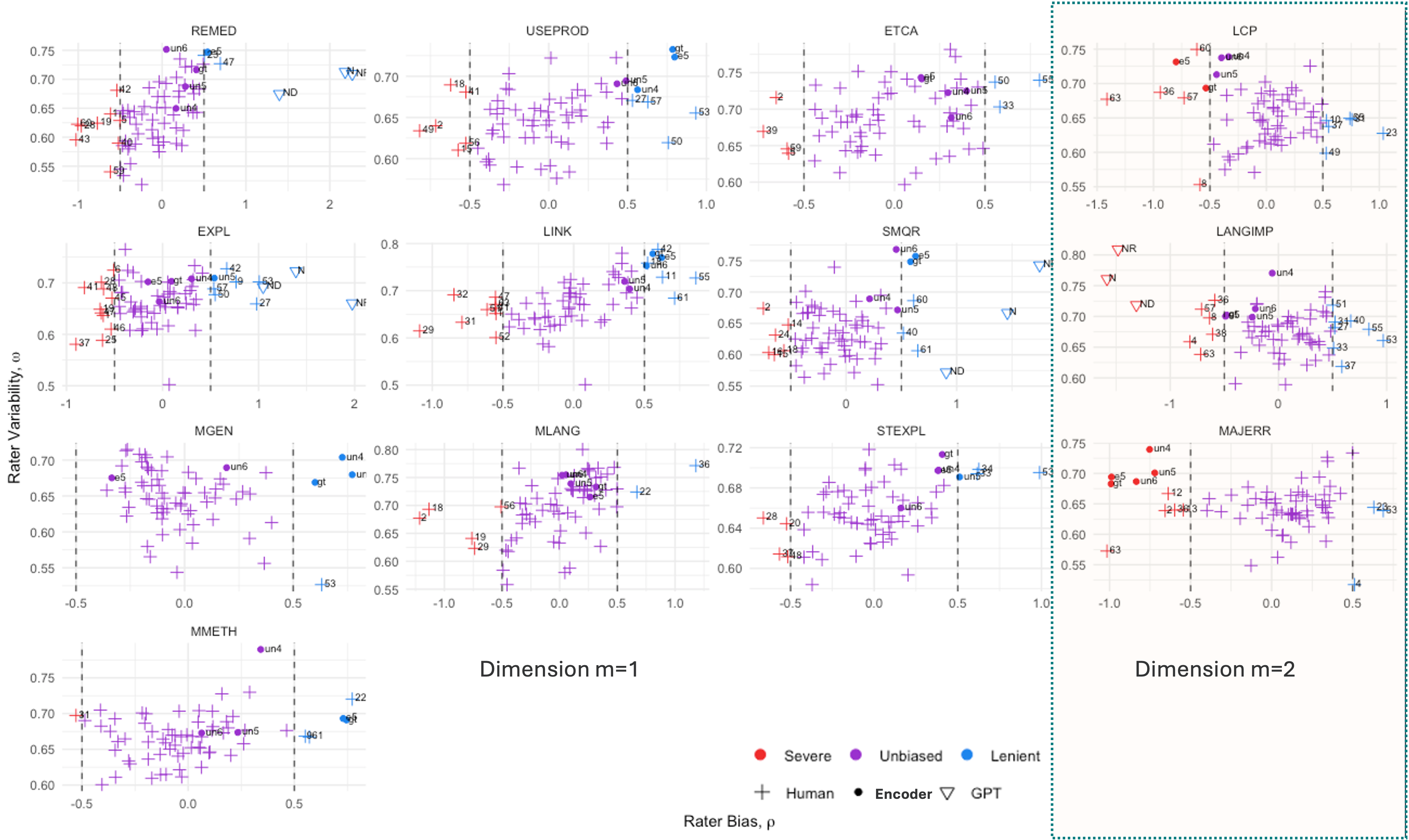}
        \caption{Rater biases, $\rho_{jr}$ , for each item $j \in \text{MQI}$ centered at an item-level detection effect$\eta_j$, and variabilities, $\omega^2$, by MQI Item and visually grouped by dimension, $m$ and marked by severity/leniency. Each point is an individual rater: a “+” marker is a single human rater; “$\bullet$”  and “$\bigtriangledown$” are specific encoder and GPT models, respectively.  X-axis is rater bias. Right is more lenient, left more severe. Color (via x-axis) are bias categories. Y-axis is rater variability (lower is more consistent. Horizontal lines 95\% CI for bias via MCMC Bayes Estimation
}
        \label{fig:item_rater_bias}
\end{figure}
\end{landscape}

\begin{landscape}
    
\begin{figure}[h]
    \centering
        \includegraphics[width=0.8\paperheight]{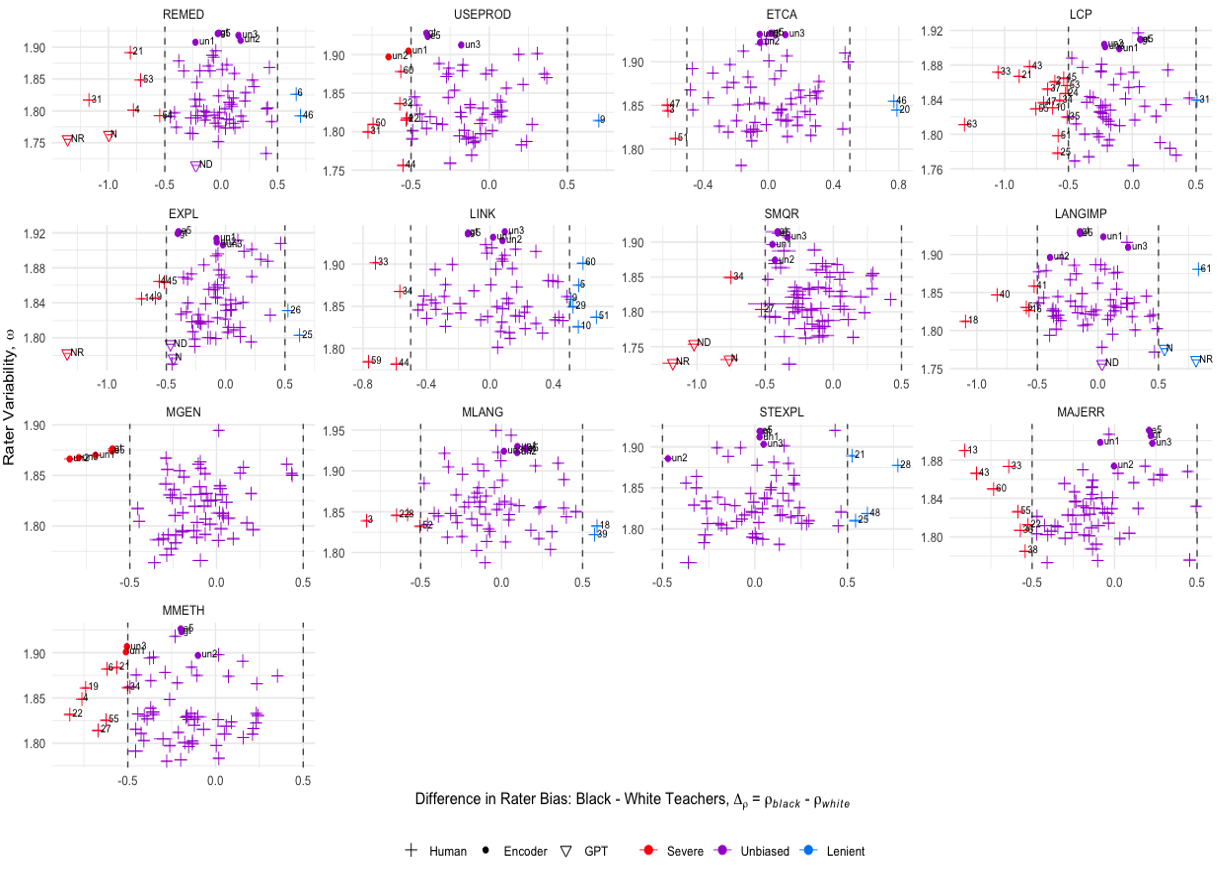}
        \caption{\textbf{Fairness across Racial Lines}. Section \ref{sec:fairness}: Standardized difference in rater bias $\phi_r$ (x axis) and rater combined variability/consistency, $\psi_r$, (y axis) across Black teachers and White teachers. Leftward values are more severe towards Black teachers, rightward are more lenient. Any horizontal bar present with a marker represents 95\% CI for bias. Differences in rater biases, $\Delta\rho_{jr}$ , for each item $j \in \text{MQI}$ centered at an item-level detection effect$\eta_j$, and variabilities, $\omega^2$ Each point is an individual rater: a “+” marker is a single human rater; “$\bullet$”  and “$\bigtriangledown$” are specific encoder and GPT models, respectively.  X-axis is rater bias. Right is more lenient, left more severe. Color categories along x-axis are bias categories. Y-axis is rater variability (lower is more consistent. Horizontal lines 95\% CI for bias via MCMC Bayes Estimation
}
        \label{fig:race_mqi_fairness}
\end{figure}
\end{landscape}

Conducting a full fairness analysis across both CLASS and MQI items and raters is considerably more complicated when accounting for all four construct dimensions in \cite{blazar_attending_2017}. If only MQI items are modeled, as was the case in the plots of Figure \ref{fig:panels}, the model can be simplified two dimensions. Full item-level MQI results for those models for disentangling biases from Section \ref{sec:biases} are in Figure \ref{fig:item_rater_bias}. The item-level results for corresponding racial bias difference models from Section \ref{sec:fairness} are in Figure \ref{fig:race_mqi_fairness}. \texttt{JAGS} code for MCMC in R is available online.\footnote{\url{https://github.com/hardy-education/LLM-Psychometrics}} A structural plate diagram for the model in Section \ref{sec:fairness} is in Figure \ref{fig:plate}. 

\texttt{JAGS} code of a full model representing Section \ref{sec:fairness}, including code for the additional estimation of CLASS items and simultaneous estimation of human and model parameters, as seen in Figure \ref{fig:plate}. To reduce the total length of code, Code Listing \ref{list:jags:dcMHRM} encapsulates all code for the various MCMC estimations used in this paper. For the creation of Panels (d) and (e) of Figure \ref{fig:panels} and Figures \ref{fig:item_rater_bias} and \ref{fig:race_mqi_fairness}, model parameters were estimated after human raters and teacher parameters were estimated and only using MQI items (i.e., \texttt{xi[i,j]} is held as fixed when estimating parameters for Encoders and GPTs). It also includes an additional hierarchical structure in latent abilities to allow for estimation of ideal scores at the lesson observation-level $\xi_{oij}$ so teacher latent abilities, $\theta_{oi}$, can vary across lessons during the year and jointly be informed by the teacher's true year-level latent abilities $\Theta_i$. This would update the top latent ability estimation Equation \ref{eq:baseHRM} to the following.

\begin{gather}\label{eq:updatedtheta}
\text{HRM}    
\begin{cases}
    \boldsymbol{\theta}_{oi} \sim \text{MVN}(\boldsymbol{\Theta}_{M \times 1},\textbf{I}_{M \times M})\text{; } \Theta_{im} \sim \mathcal{N}(0,1)
  \text{,}\\
  \xi_{oij} \sim \text{\textbf{IRT model}} \\
  X_{soijr} \sim \text{\textbf{SDT model}}    
\end{cases}
\end{gather}

\begin{longlisting}
\begin{minted}{R}
model
{
## Signal detection theory model with rater covariates
    for (i in 1:NN) {
        x[i] ~ dcat(prob.sdt[i, ])
        for (k in 1:K) {
            d[i, k] <- k - xi[subject[i], item[i]] - rhocov.r[rater[i], 
                item[i], race[i]]
            z[i, k] <- exp(-d[i, k] * d[i, k]/2 * exp(zeta.r[rater[i], 
                item[i], race[i]]))
            prob.sdt[i, k] <- ifelse((K - maxscore.by.item[item[i]]), 
                ifelse(k < (maxscore.by.item[item[i]] + 1), z[i, 
                  k]/sum(z[i, ]), 0.00000E+00), z[i, k]/sum(z[i, 
                  ]))
        }
    }

## Multidimensional Generalized Partial Credit Model
    for (i in 1:N) {
        for (j in 1:J) {
            xi[i, j] ~ dcat(prob.irt[i, j, ])
            for (m in 1:M) {
                kern[i, j, m] <- alpha[j, m] * (theta[i, m])
            }
            for (k in 1:K) {
                dotprod[i, j, k] <- (k - 1) * sum(kern[i, j, 
                  ])
                eta[i, j, k] <- dotprod[i, j, k] - sum(gamma[j, 
                  1:k])
                exp.eta[i, j, k] <- exp(eta[i, j, k])
                prob.irt[i, j, k] <- ifelse(K - maxscore.by.item[j], 
                  ifelse(k <= (maxscore.by.item[j]), exp.eta[i, 
                    j, k]/sum(exp.eta[i, j, 1:maxscore.by.item[j]]), 
                    0), exp.eta[i, j, k]/sum(exp.eta[i, 
                    j, 1:maxscore.by.item[j]]))
            }
        }
    }
## Rater Parameters
    for (nu in r1.raters) {
        for (s in r.1.in) {
            for (ra in 1:RA) {
                rhocov.r[nu, s, ra] ~ dnorm(eta.rt[s, ra], prec.rhocov)
                zeta.r[nu, s, ra] ~ dnorm(kappa.rt[s, ra], prec.zeta)
                omega.r[nu, s, ra] <- sqrt(1/exp(zeta.r[nu, s, 
                  ra]))
            }
        }
        for (s in r.1.out) {
            for (ra in 1:RA) {
                rhocov.r[nu, s, ra] <- 0
                zeta.r[nu, s, ra] <- 0
                omega.r[nu, s, ra] <- 1
            }
        }
    }
    for (nu in r2.raters) {
        for (s in r.2.in) {
            for (ra in 1:RA) {
                rhocov.r[nu, s, ra] ~ dnorm(eta.rt[s, ra], prec.rhocov)
                zeta.r[nu, s, ra] ~ dnorm(kappa.rt[s, ra], prec.zeta)
                omega.r[nu, s, ra] <- sqrt(1/exp(zeta.r[nu, s, 
                  ra]))
            }
        }
        for (s in r.2.out) {
            for (ra in 1:RA) {
                rhocov.r[nu, s, ra] <- 0
                zeta.r[nu, s, ra] <- 0
                omega.r[nu, s, ra] <- 1
            }
        }
    }
    for (nu in r3.raters) {
        for (s in r.3.in) {
            for (ra in 1:RA) {
                rhocov.r[nu, s, ra] ~ dnorm(eta.rt[s, ra], prec.rhocov)
                zeta.r[nu, s, ra] ~ dnorm(kappa.rt[s, ra], prec.zeta)
                omega.r[nu, s, ra] <- sqrt(1/exp(zeta.r[nu, s, 
                  ra]))
            }
        }
    }
    for (nu in r4.raters) {
        for (s in r.4.in) {
            for (ra in 1:RA) {
                rhocov.r[nu, s, ra] ~ dnorm(eta.rt[s, ra], prec.rhocov)
                zeta.r[nu, s, ra] ~ dnorm(kappa.rt[s, ra], prec.zeta)
                omega.r[nu, s, ra] <- sqrt(1/exp(zeta.r[nu, s, 
                  ra]))
            }
        }
        for (s in r.4.out) {
            for (ra in 1:RA) {
                rhocov.r[nu, s, ra] <- 0
                zeta.r[nu, s, ra] <- 0
                omega.r[nu, s, ra] <- 1
            }
        }
    }

## Multidimension parameters
    for (m in 1:M) {
        pi.rt[m] <- 0
        delta.rt[m] <- 0
        sigma.rt[m] <- 1
    }

## Item Parameters
    for (s in 1:S) {
        for (ra in 1:RA) {
            eta.rt[s, ra] ~ dnorm(pi.rt[factors.by.item[s]], 
                prec.eta)
            kappa.rt[s, ra] ~ dnorm(delta.rt[factors.by.item[s]], 
                prec.kappa)
            tau.rt[s, ra] <- sqrt(1/exp(kappa.rt[s, ra]))
        }
    }

## Initializations for rater and item parameters
    prec.pi ~ dgamma(a.precpi, b.precpi)
    prec.delta ~ dgamma(a.precdelta, b.precdelta)
    prec.eta ~ dgamma(a.preceta, b.preceta)
    prec.kappa ~ dgamma(a.preckappa, b.preckappa)
    prec.rhocov ~ dgamma(a.precrhocov, b.precrhocov)
    prec.zeta ~ dgamma(a.preczeta, b.preczeta)
    sd.rhocov <- sqrt(1/prec.rhocov)
    sd.zeta <- sqrt(1/prec.zeta)
    sd.pi <- sqrt(1/prec.pi)
    sd.delta <- sqrt(1/prec.delta)
    sd.eta <- sqrt(1/prec.eta)
    sd.kappa <- sqrt(1/prec.kappa)
    for (m in 1:M) {
        alpha[d2[1], m] <- ifelse(m == 2, 1, 0)
        alpha[d1[1], m] <- ifelse(m == 1, 1, 0)
        alpha[d3[1], m] <- ifelse(m == 3, 1, 0)
        alpha[d4[1], m] <- ifelse(m == 4, 1, 0)
    }
    for (j in d1[2:D1]) {
        alpha[j, 1] ~ dlnorm(0, prec.alpha)
        alpha[j, 2] <- 0
        alpha[j, 3] <- 0
        alpha[j, 4] <- 0
    }
    for (j in d2[2:D2]) {
        alpha[j, 2] ~ dlnorm(0, prec.alpha)
        alpha[j, 1] <- 0
        alpha[j, 3] <- 0
        alpha[j, 4] <- 0
    }
    for (j in d3[2:D3]) {
        alpha[j, 3] ~ dlnorm(0, prec.alpha)
        alpha[j, 2] <- 0
        alpha[j, 1] <- 0
        alpha[j, 4] <- 0
    }
    for (j in d4[2:D4]) {
        alpha[j, 4] ~ dlnorm(0, prec.alpha)
        alpha[j, 1] <- 0
        alpha[j, 2] <- 0
        alpha[j, 3] <- 0
    }
    for (j in 1:J) {
        gamma[j, 1] <- 0
        for (k in 2:maxscore.by.item[j]) {
            gamma[j, k] ~ dnorm(0, prec.gamma)
        }
        for (k in (maxscore.by.item[j] + 1):(K + 1)) {
            gamma[j, k] <- 0
        }
    }
## Theta estimations
    for (i in 1:TY) {
        for (m in 1:M) {
            ty[i, m] ~ dnorm(0, prec.ty)
        }
    }
    for (i in 1:N) {
        theta[i, 1:M] ~ dmnorm(ty[tyr.by.obs[i], ], Tau[, ])
    }
    Tau[1:M, 1:M] ~ dwish(W[, ], DF)
    Sigma <- inverse(Tau[, ])
    sd.th1 <- sqrt(Sigma[1, 1])
    sd.th2 <- sqrt(Sigma[2, 2])
    rho12 <- Sigma[1, 2]/sqrt(Sigma[1, 1] * Sigma[2, 2])
    prec.ty ~ dgamma(a.precty, b.precty)
    sd.ty <- 1/sqrt(prec.ty)
    prec.b <- pow(var.b, -1)
    prec.g <- pow(var.g, -1)
    prec.alpha <- pow(var.alpha, -1)
    prec.gamma <- pow(var.gamma, -1)
    prec.phi <- pow(var.phi, -1)
}
## initial values
inits <- function() list(
    alpha = item.dims  * runif(J*M,0.1,1.5),
    gamma = item.cats.by.score * rnorm(J*(K+1),0,0.5), # 
    ty= matrix(rep(rnorm(TY, 0, 1),M),nrow=TY,ncol=M),
    theta = matrix(rnorm(N*M,0,1),ncol=M),
    phi = rnorm(R, 0, 1),
    tau = runif(R, 0.1, 8),
    rhocov = array(rnorm(R*S*RA),dim = c(R,S,RA)) * rnorm(R*S*RA,0,.5),
    zeta =  array(rnorm(R*S*RA),dim = c(R,S,RA)), 
    pi = rnorm(R,0,.5),
    delta = rnorm(R,0,.5),
    kappa = rnorm(R,0,.5),
    theta.prec = rgamma(1,100,100))
\end{minted}
\caption{\texttt{JAGS} code of a full model representing Section \ref{sec:fairness}, including code for the additional estimation of CLASS items and simultaneous estimation of human and model parameters, as seen in Figure \ref{fig:plate}. For brevity, this includes all code which can be reduced for the various methods herein. For the creation of Panels (d) and (e) of Figure \ref{fig:panels} and Figures \ref{fig:item_rater_bias} and \ref{fig:race_mqi_fairness}, model parameters were estimated after human raters and teacher parameters were estimated and only using MQI items (i.e., \texttt{xi[i,j]} is held as fixed when estimating parameters for Encoders and GPTs).}\label{list:jags:dcMHRM}
\end{longlisting}

\newgeometry{margin=2 cm}
\restoregeometry

\newgeometry{margin=2 cm}
\restoregeometry

\newgeometry{margin=2 cm}
\restoregeometry

\section{Generalizability and Decision Studies}\label{apx:gstudies}
\subsection{Generalizability Study Human Results (for NCTE Main Study)}\label{apdx:humgstudy}
The results of the item-level G-study for human expert ratings, consisting of only the estimates for individual items using the NCTE Main Study data \cite{kane_national_2015} to replicate Section 2.d from the Appendix. All calculations and representations are according to the design details listed therein. 

In the Appendix of the NCTE study, the authors submitted a G-study on the MQI instrument, but not for data of the study: they provide a separate G-study of only eight (8) different middle school teachers teaching three (3) lessons each with only nine (9) raters, instead of the corresponding 317 NCTE Study teachers with an average 5.34 lessons each and 63 raters. For completeness, this paper conducts the G-study for the NCTE main study Appendix, Section 3, using the NCTE dataset. The full results of the human label G-study are in Table \ref{tab:g_study_res}.

\begin{figure}
    \centering
    \includegraphics[width=1\linewidth]{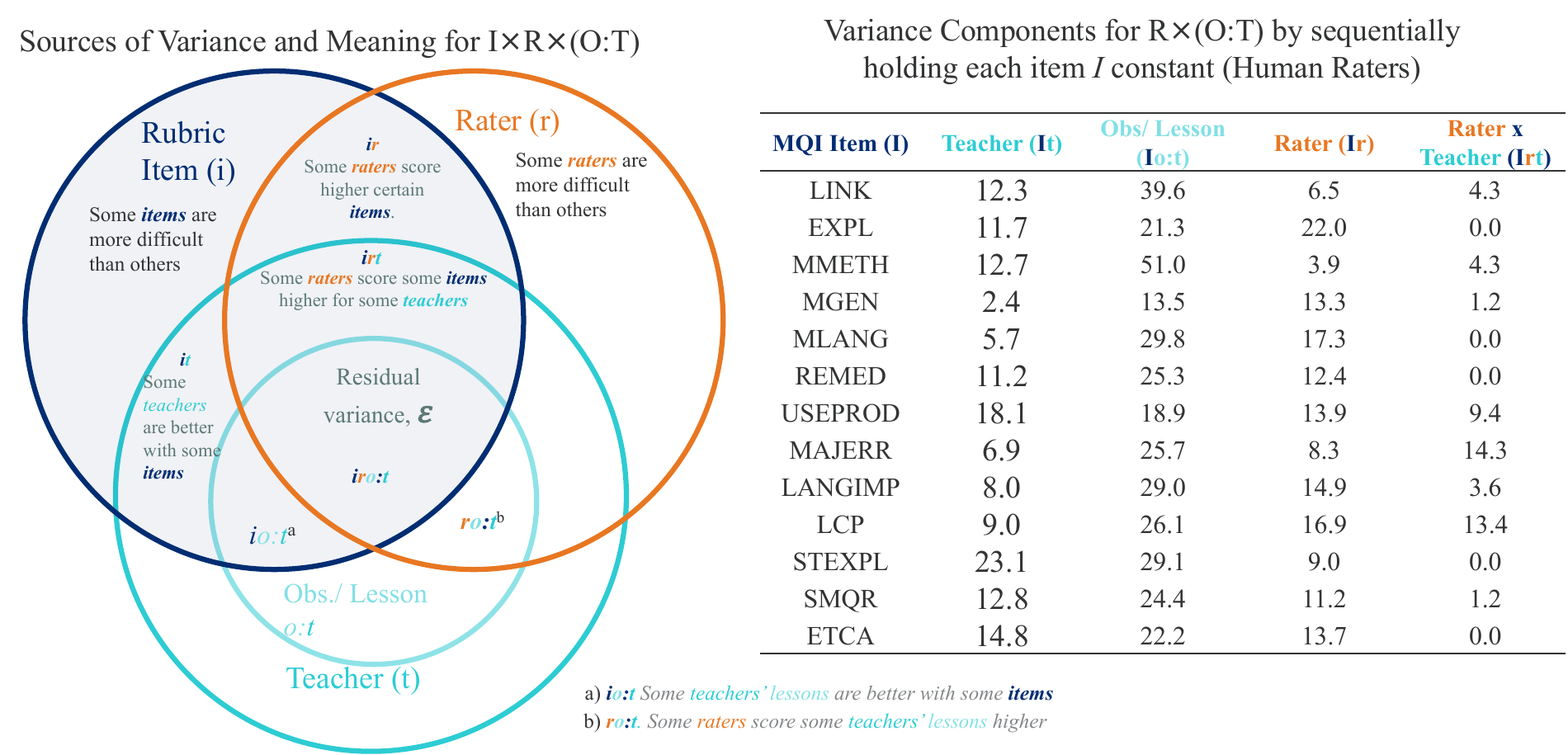}
    \caption{Variance components for Generalizability Calculations}
    \label{fig:enter-label}
    \begin{table}[H]
\caption{By item, the percentage contribution, excluding the residual (which accounts for the remainder of the variance), of each variance component in the given MQI Item's  R x (O:T) Generalizability Study}
\label{tab:g_study_res}
\end{table}
\end{figure}

\begin{table}[ht]
\caption{By item, the percentage contribution, excluding the residual (which accounts for the remainder of the variance), of each variance component in the given MQI Item's  R x (O:T) Generalizability Study}
\label{tab:g_study_res}
\end{table}

\subsection{Item Generalizability and Item-score Reliability}\label{apx:ep2_g6}
\begin{figure}
    \centering
    \includegraphics[width=1\linewidth]{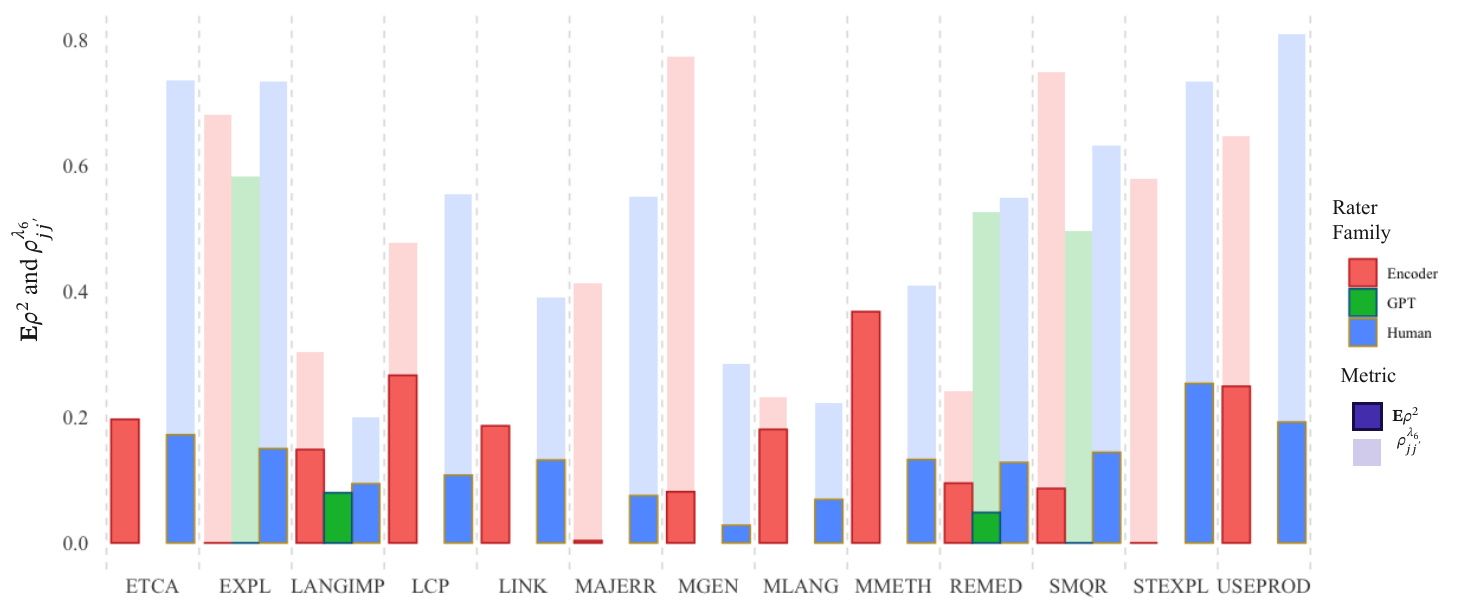}
    \caption{Estimates for Family-wise Item-level Generalizability,  $\mathbf{E}\hat{\rho}^2_j$ , and Reliability $\hat{\rho}^{\lambda_6}_{jj\prime}$}.
    \label{fig:ep2_pjj}
\end{figure}

As a complement and context stemming from Sections \ref{sec:reliabilities} and \ref{sec:gtheorystudy}, $\mathbf{E}\hat{\rho}^2_j$ item values to item-level reliability estimates related to Guttman's $\lambda_6$ \citep{guttman_basis_1945}, $\hat{\rho}^{\lambda_6}_{jj\prime}$ \citep{zijlmans_item-score_2018,zijlmans_methods_2018}. $\hat{\rho}^{\lambda_6}_{jj\prime}$ represents the proportion of an item's variance shared by the to variance captured by other items. This estimate from Classical Test Theory (naïvely, in this case) assumes that all items measure the same latent construct, i.e., the Mathematical Quality of Instruction \citep{hill_mathematical_2008}. $\hat{\rho}^{\lambda_6}_{jj\prime}$ removes the variance in the residual error, $\sigma^2_{\varepsilon_j}$, from a multiple regression of item $j$ on the scores from the remaining $J - 1$ items to estimate the proportion of total item variance $\sigma^2_{X_j}$ consistent with the unidimensional construct shared with the other items. Figure \ref{fig:ep2_pjj} highlights the large difference in the measurement used in Section \ref{sec:gtheorystudy} and item reliabilities from classical test theory. The latter of which describes the item reliability based on \textit{all} \textit{scores}, while the former is used in this study because it is more related to the reliability of \textit{individual} \textit{scores} for a given item. 

\subsection{Generalizability Theory Parameters and Code}\label{apdx:gstudy}
A helpful heuristic for understanding the mathematics of G-theory might be they are very computationally similar to hierarchical mixed effect models, where estimates of interest are found in variation of the random effects. The two code blocks represent by item $(O:I) \times R$ and $(S:O:I) \times R$ parameterizations, respectively, using variable names from the original dataset. The former replicates the methods used in \cite{hill_when_2012} and the Appendix Section 2.d of \cite{kane_national_2015} to create Table \ref{tab:g_study_res} in Appendix section \ref{apdx:humgstudy}, and was used in this study to calculate the family generalizability metrics in Section \ref{sec:gtheorystudy}, including those used in Section \ref{sec:spurious}. The latter is used for the decision studies described in Section \ref{sec:dstudy}. Studies were conducted using \texttt{lme4} \citep{bates_fitting_2015} in \texttt{R} \citep{r_core_team_r_nodate}

\begin{landscape}
    \begin{figure}[h]
        \centering
        \includegraphics[width=0.7\paperheight]{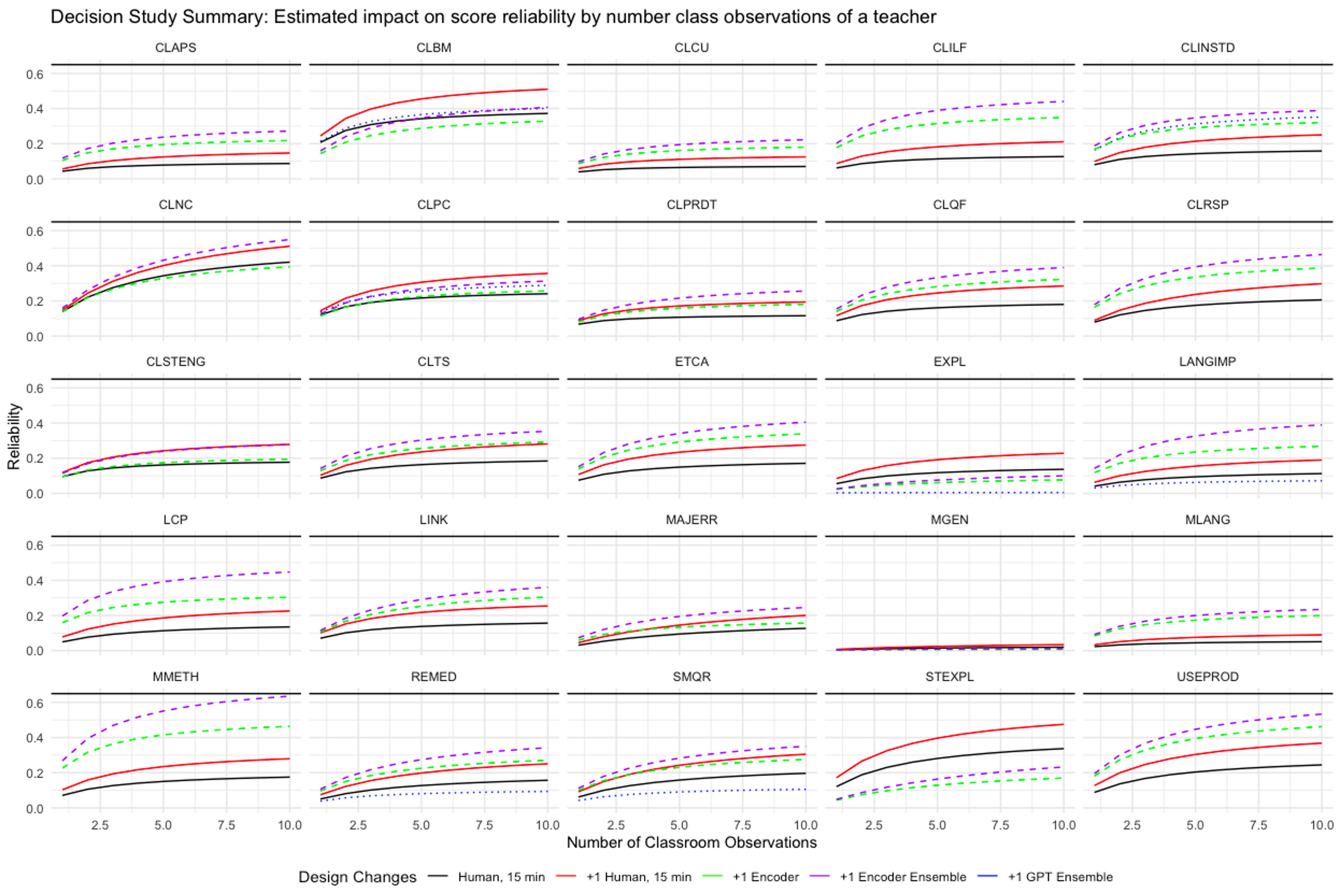}
        \caption{Expected changes to rating reliability are estimated improvements to quality (via reliability) of classroom ratings for various contexts. The single individual human baseline (black) estimates reliability improvements by visiting the same class the x axis represents the number of different 15 min. classroom observations of the same teacher. The red line is estimate of having a \textit{different} human observer conduct observations as described. By contrast, for the model raters--single Encoder (green), Encoder ensemble (average of 3 encoders) (Red), and GPT ensemble (average of 3 GPT prompt engineered models)--the x-axis for models is the number of full classroom observations conducted where the human (black) observes at least 15 minutes (in-the-loop) of the same classroom (models observe the entire class period).}
        \label{fig:dstudyfullcombined}
    \end{figure}
\end{landscape}

Full results for item-level d-studies as defined in Section \ref{sec:dstudy} are in Figure \ref{fig:dstudyfullcombined}. 

\begin{longlisting}
    \begin{minted}{R} 
for (item in ITEMS){
    m[[item]] <- lmer(data = df|>
                filter(R_TYPE == rater.type),
        formula = SCORE ~ 
                       (1|RATERID) + 
                       (1|NCTETID/OBSID) +
                       (1|ITEM) + 
                       (1|RATERID:NCTETID) +
                       (1|RATERID:OBSID) + 
                       (1|ITEM:NCTETID) + 
                       (1|ITEM:OBSID) + 
                       (1|RATERID:ITEM) + 
                       (1|ITEM:RATERID:NCTETID) 
}
\end{minted}
\label{code:gtheory_full}
\caption{\texttt{lme4} code for Family-wise all item estimations in Table \ref{tab:gen_and_dep}}
\end{longlisting}

\begin{longlisting}
    \begin{minted}{R} 
for (item in ITEMS){
    m[[item]] <- lmer(data = df|>
                filter(ITEM == item)|>
                filter(R_TYPE == rater.type),
        formula = SCORE ~ (1|NCTETID/OBSID) +
                (1|RATERID) + 
                (1|RATERID:NCTETID)
}
\end{minted}
\label{code:gtheory}
\caption{\texttt{lme4} code for item-level estimations of $\mathbf{E}\hat{\rho}^2_j$ in Equation \ref{eq:Erho}}
\end{longlisting}

\begin{longlisting}
    \begin{minted}{R} 
for (item in ITEMS){
    m[[item]] <- lmer(data = df|>
                filter(ITEM == item)|>
                filter(R_TYPE == rater.type),
        formula = SCORE ~ (1|NCTETID_SCHOOLYEAR_SP/OBSID/CHAPNUM) +
                (1|RATERID) + 
                (1|RATERID:NCTETID_SCHOOLYEAR_SP)
}
\end{minted}
\label{code:dstudy}
\caption{\texttt{lme4} code for item-level estimations used in Equation \ref{eq:dstudyrho}}
\end{longlisting}


\restoregeometry

\section{Interpretability of Encoder Labels}
\subsection{Feature Attribution Models and Tools}\label{apx:feature_attribution}
The Explainable Artificial Intelligence (XAI) community has proposed various cutting-edge methodologies to enhance the explainability of deep learning models. A popular strategy is feature attribution, wherein for a given neural network model f, an attribution method E delineates the significance of each input feature of x to the prediction y = f(x). Various strategies to ascertain feature importance have been introduced, encompassing gradient-based methods, surrogate methods, and perturbation-based methods. Our study employs Integrated Gradients, a gradient-based approach developed by \citet{sundararajan_axiomatic_2017}, to identify pivotal sentences for classroom quality assessment. Integrated Gradients is engineered to comply with two essential axioms—Sensitivity and Implementation Invariance—that attribution methods ought to adhere to, as defined below:

\begin{center}
\begin{align*}\label{eq:ig}
&\text{IntegratedGrads}^{approx}_{i}(x):: =  \\
&(x_{i}-x'_{i})\times\sum_{k=1}^{m}\frac{\partial F(x' + \frac{k}{m}\times(x - x'))}{\partial x_{i}} \times \frac{1}{m}.
\end{align*}
\end{center}

In the above, $(x_{i}-x'_{i})$ is the difference between the inputs, $x_i$ and the baseline, and $m$ is the number of loops used for each step in a Riemann approximation of the exact integral, as presented by \citet{sundararajan_axiomatic_2017}. Integrated Gradients compute the average gradient by interpolating between a chosen baseline and the input. The resulting attributions are subsequently obtained as the element-wise product of this path-averaged gradient vector and the difference vector between the input and the baseline.

\end{document}